%% file: main.tex
\documentclass[10pt,twocolumn,letterpaper]{article}

\usepackage[pagenumbers]{iccv} %

\input{preamble}

\definecolor{iccvblue}{rgb}{0.21,0.49,0.74}
\usepackage[pagebackref,breaklinks,colorlinks,allcolors=iccvblue]{hyperref}
\usepackage{multirow}

\def\paperID{3444} %
\def\confName{ICCV}
\def\confYear{2025}

\newcommand{\xpar}[1]{\noindent\textbf{#1}\ \ }

\newcommand\blfootnote[1]{
  \begingroup
  \renewcommand\thefootnote{}\footnote{#1}
  \addtocounter{footnote}{-1}
  \endgroup
    \vspace{-\baselineskip}
}

\newcommand{\norm}[1]{\left\lVert#1\right\rVert}
\DeclareMathOperator*{\minimize}{minimize}

\title{Princeton365: A Diverse Dataset with Accurate Camera Pose}

\author{Karhan Kayan$^*$, Stamatis Alexandropoulos$^*$, Rishabh Jain, Yiming Zuo, Erich Liang, Jia Deng\\
Princeton University\\
{\tt\small \{karhan,sa6924,jainr,zuoym,el1685,jiadeng\}@princeton.edu}}

\begin{document}

\maketitle

\begin{abstract}
We introduce \projectname, a large-scale diverse dataset of 365 videos with accurate camera pose. Our dataset bridges the gap between accuracy and data diversity in current SLAM benchmarks by introducing a novel ground truth collection framework that leverages calibration boards and a $360^{\circ}$ camera. We collect indoor, outdoor, and object scanning videos with synchronized monocular and stereo RGB video outputs as well as IMU. We further propose a new scene scale-aware evaluation metric for SLAM based on the optical flow induced by the camera pose estimation error. In contrast to the current metrics, our new metric allows for comparison between the performance of SLAM methods across scenes as opposed to existing metrics such as Average Trajectory Error (ATE), allowing researchers to analyze the failure modes of their methods. We also propose a challenging Novel View Synthesis benchmark that covers cases not covered by current NVS benchmarks, such as fully non-Lambertian scenes with $360^{\circ}$ camera trajectories. Please visit \href{https://princeton365.cs.princeton.edu/}{princeton365.cs.princeton.edu} for the dataset, code, videos, and submission. 
\end{abstract}

\blfootnote{*These authors contributed equally (random order).}

\begin{figure*}[t]
  \centering
  \setlength{\tabcolsep}{0.5pt} %
  \renewcommand{\arraystretch}{0.0}
    \begin{tabular}{c|ccccc}
     \raisebox{0.25\height}{\rotatebox{90}{Scanning}} &
    \includegraphics[width=0.19\textwidth]{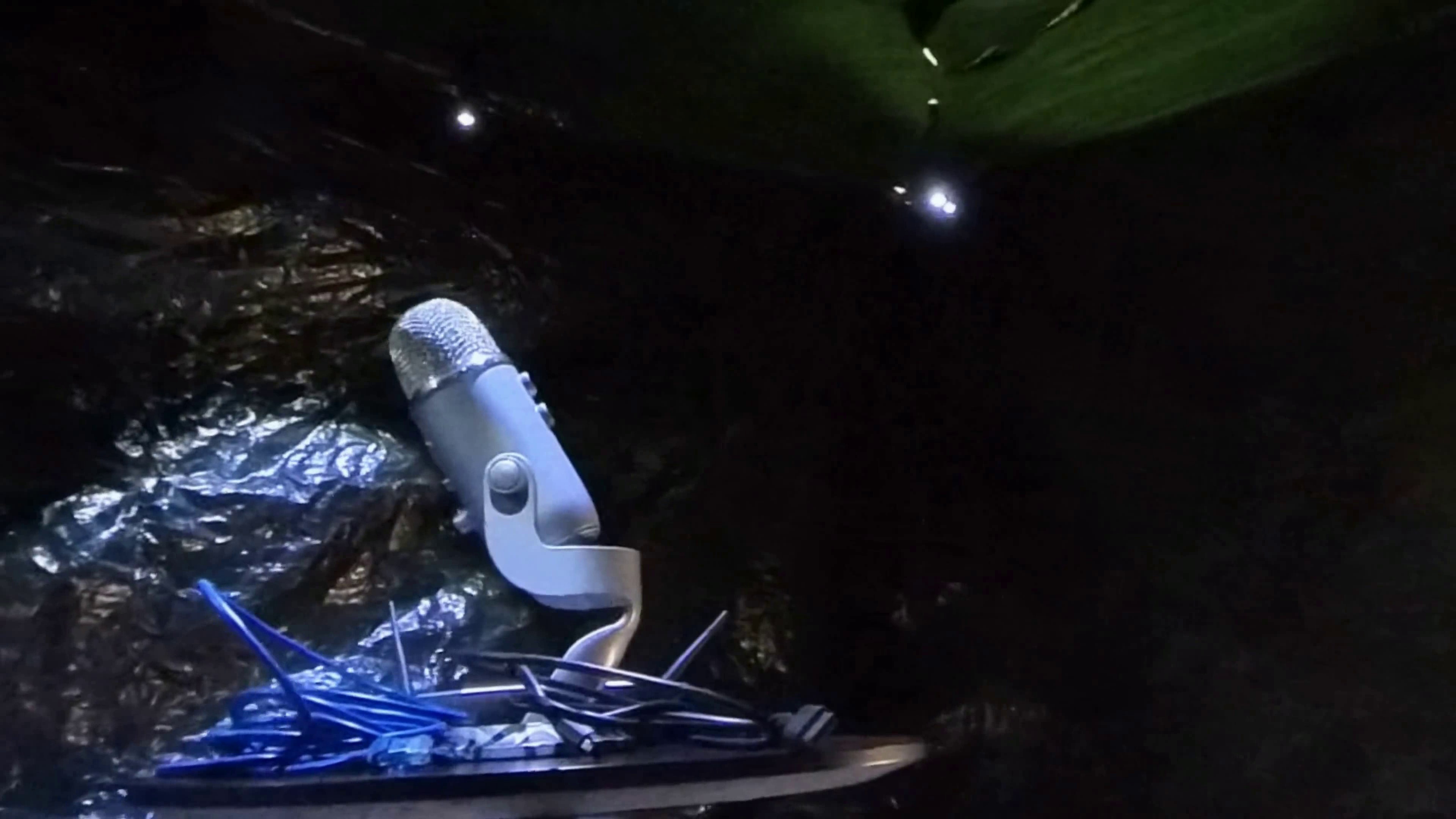} &
    \includegraphics[width=0.19\textwidth]{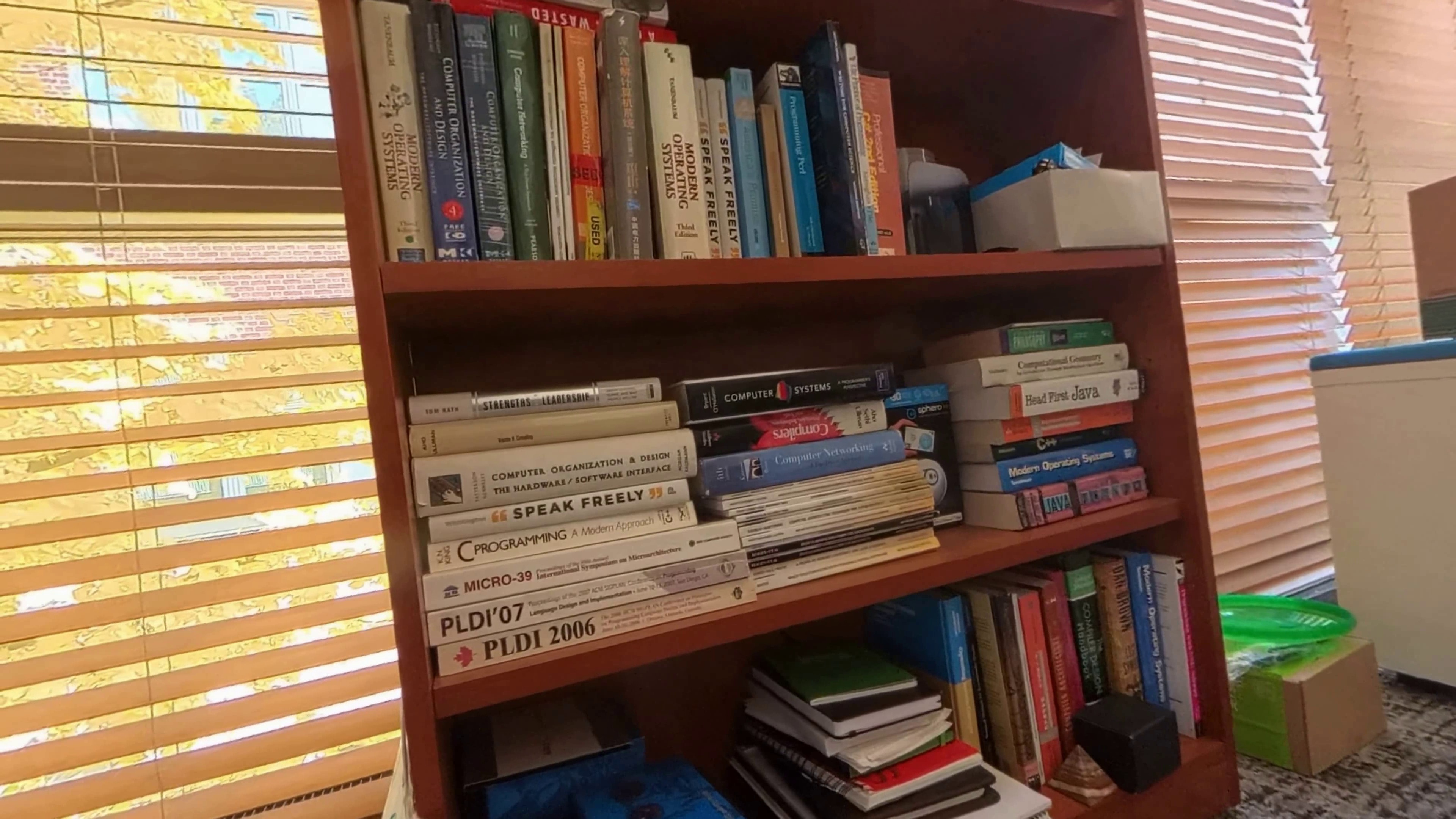} &
    \includegraphics[width=0.19\textwidth]{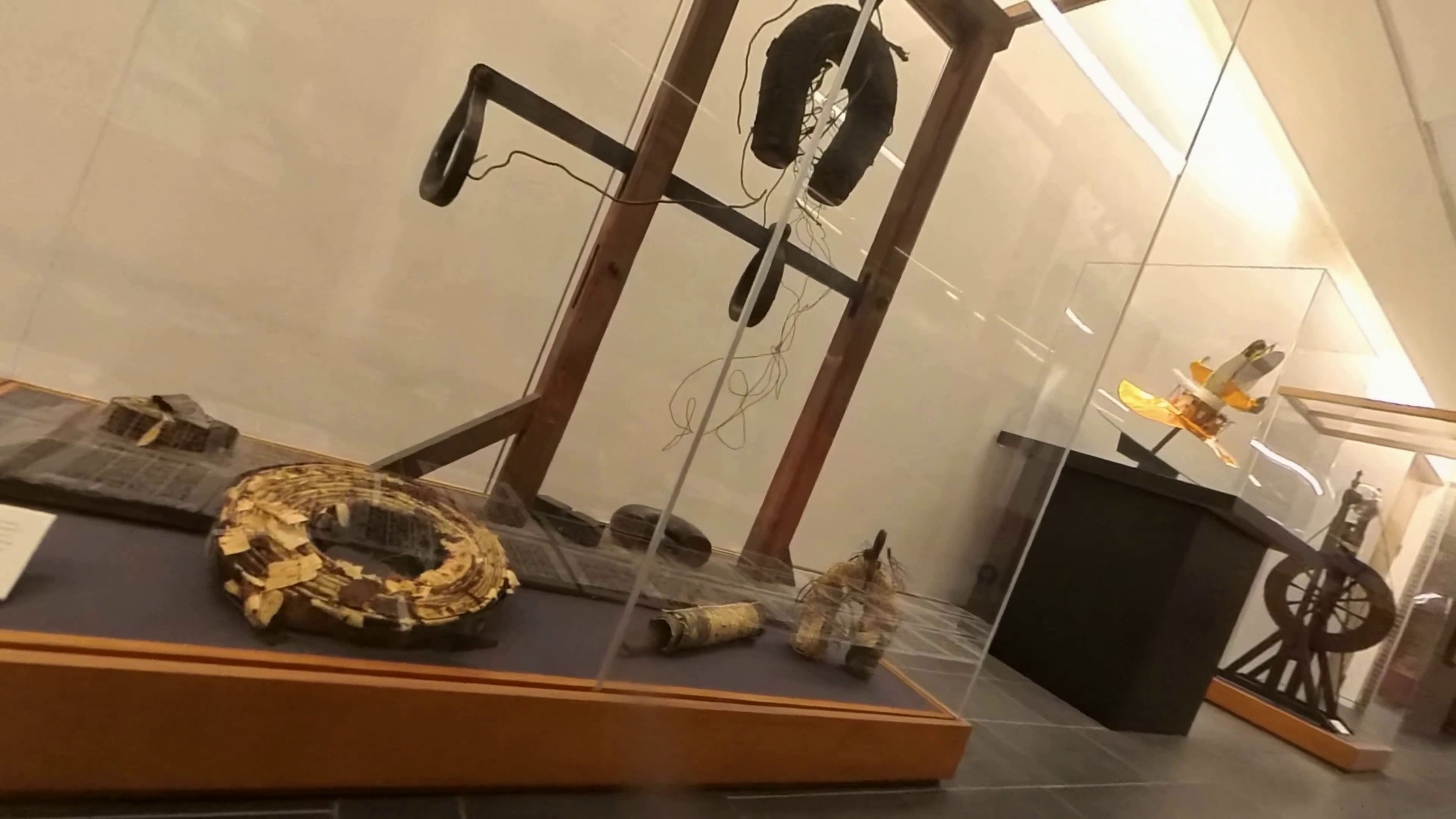} &
    \includegraphics[width=0.19\textwidth]{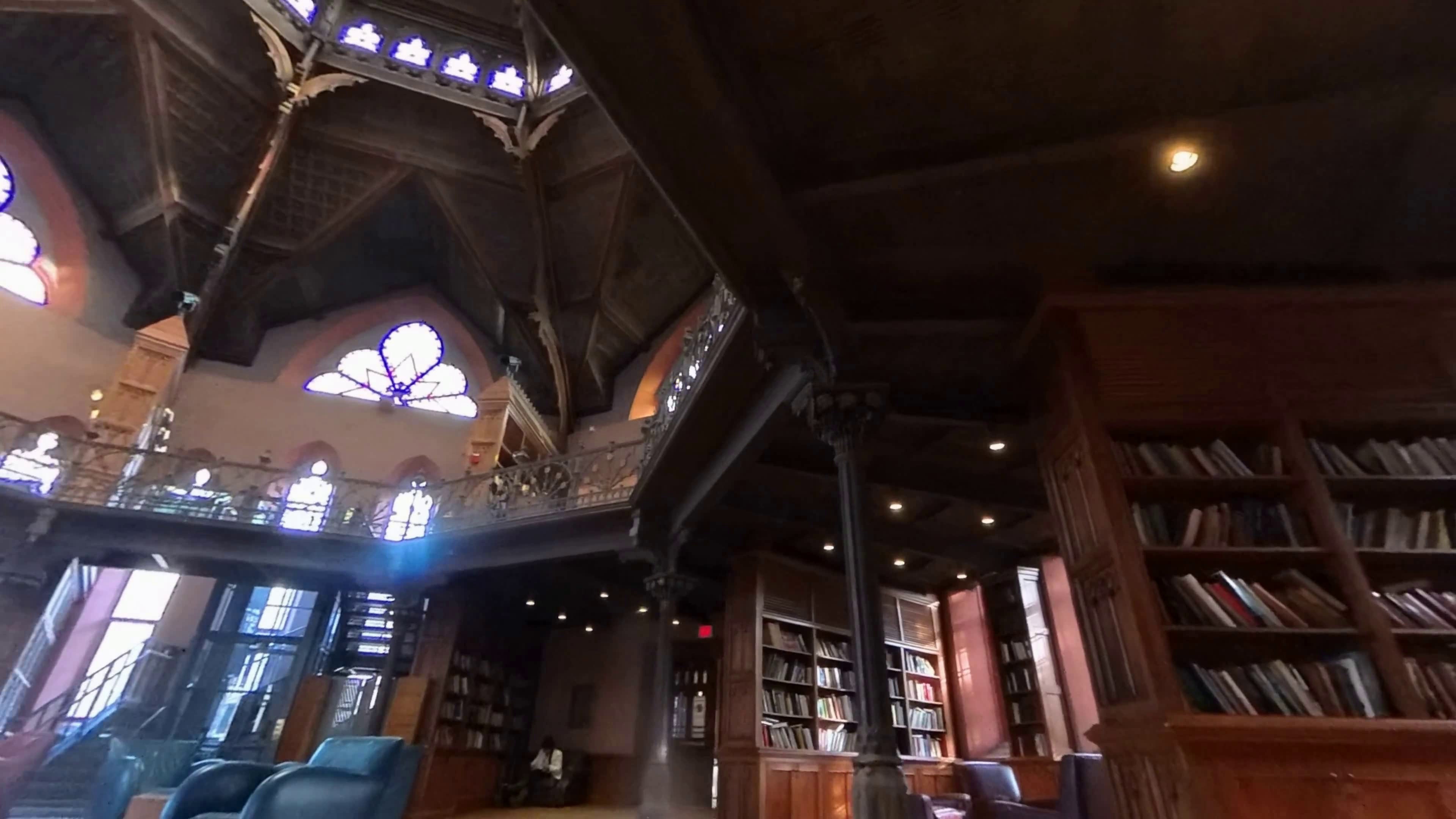} &
    \includegraphics[width=0.19\textwidth]{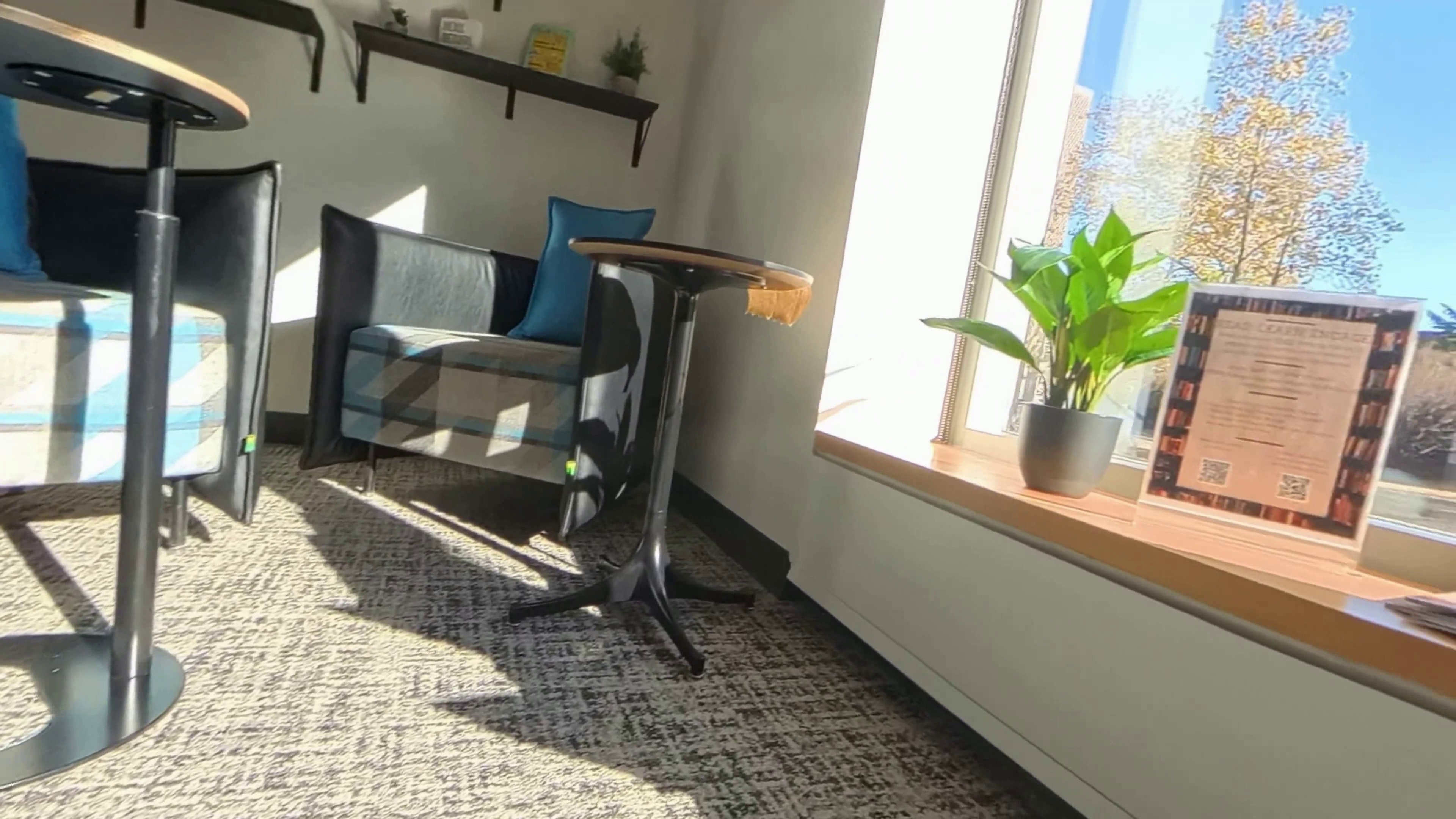} \\
    
    \raisebox{0.25\height}{\rotatebox{90}{Indoors}}
 &
    \includegraphics[width=0.19\textwidth]{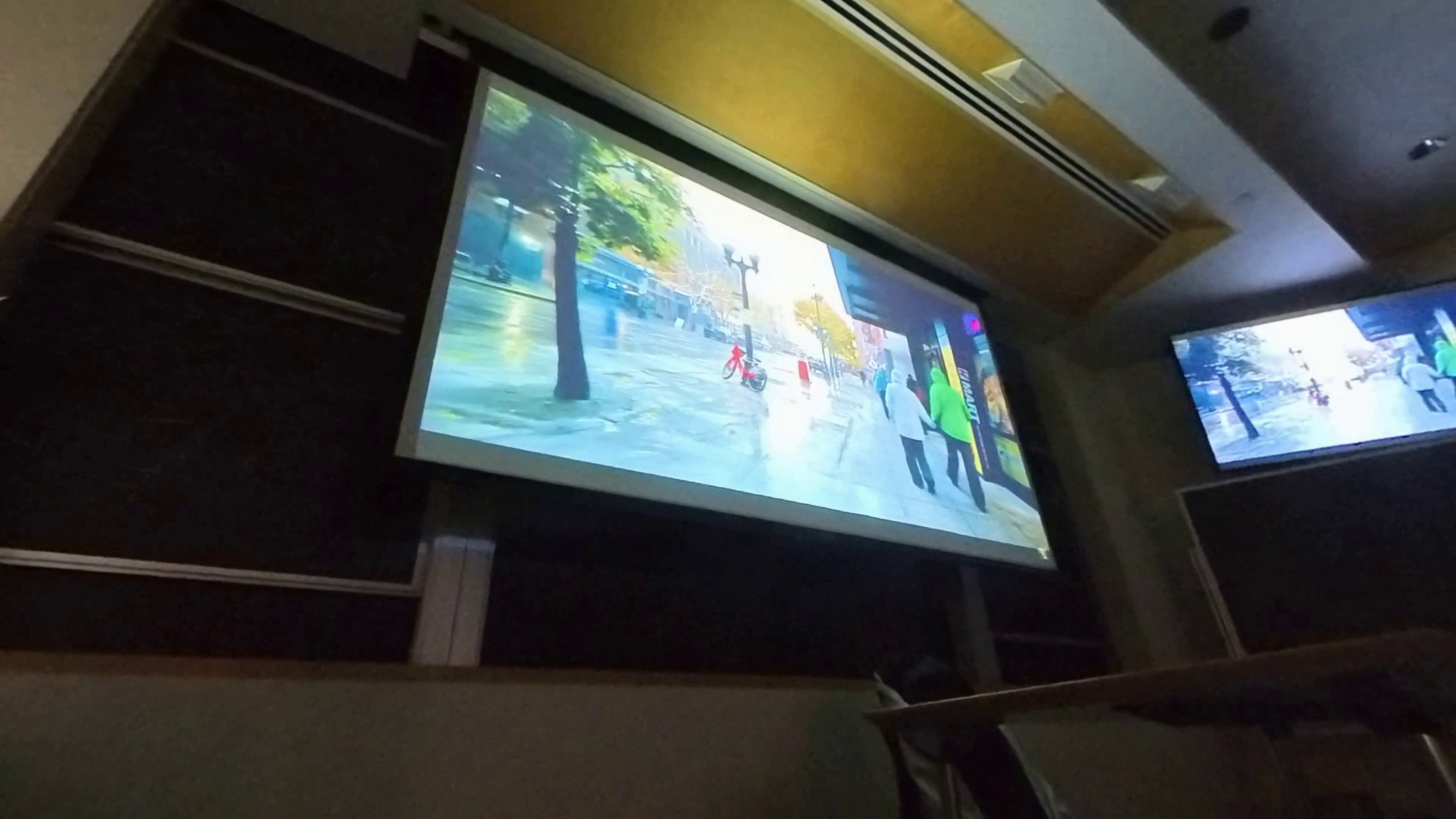} &
    \includegraphics[width=0.19\textwidth]{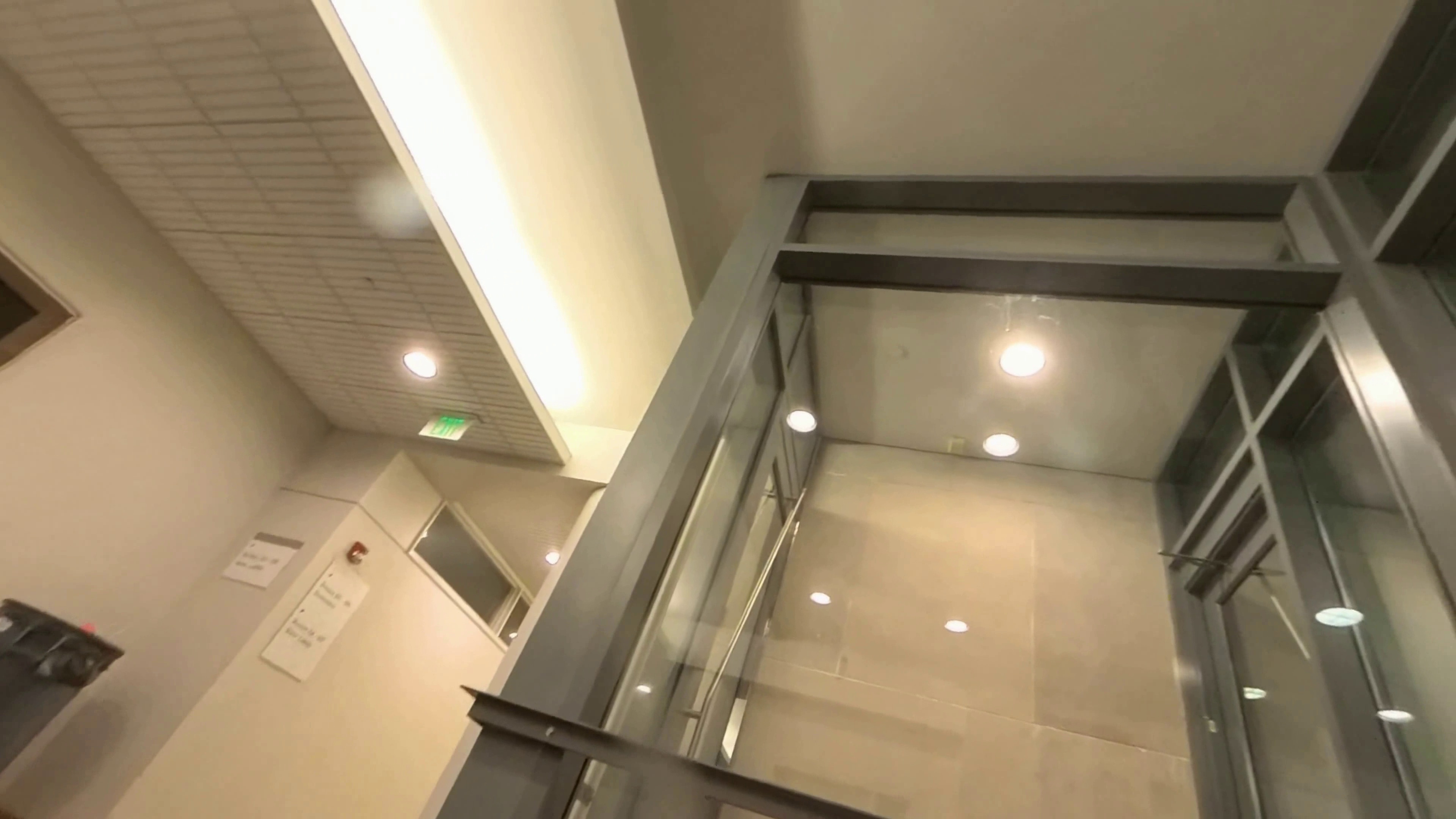} &
    \includegraphics[width=0.19\textwidth]{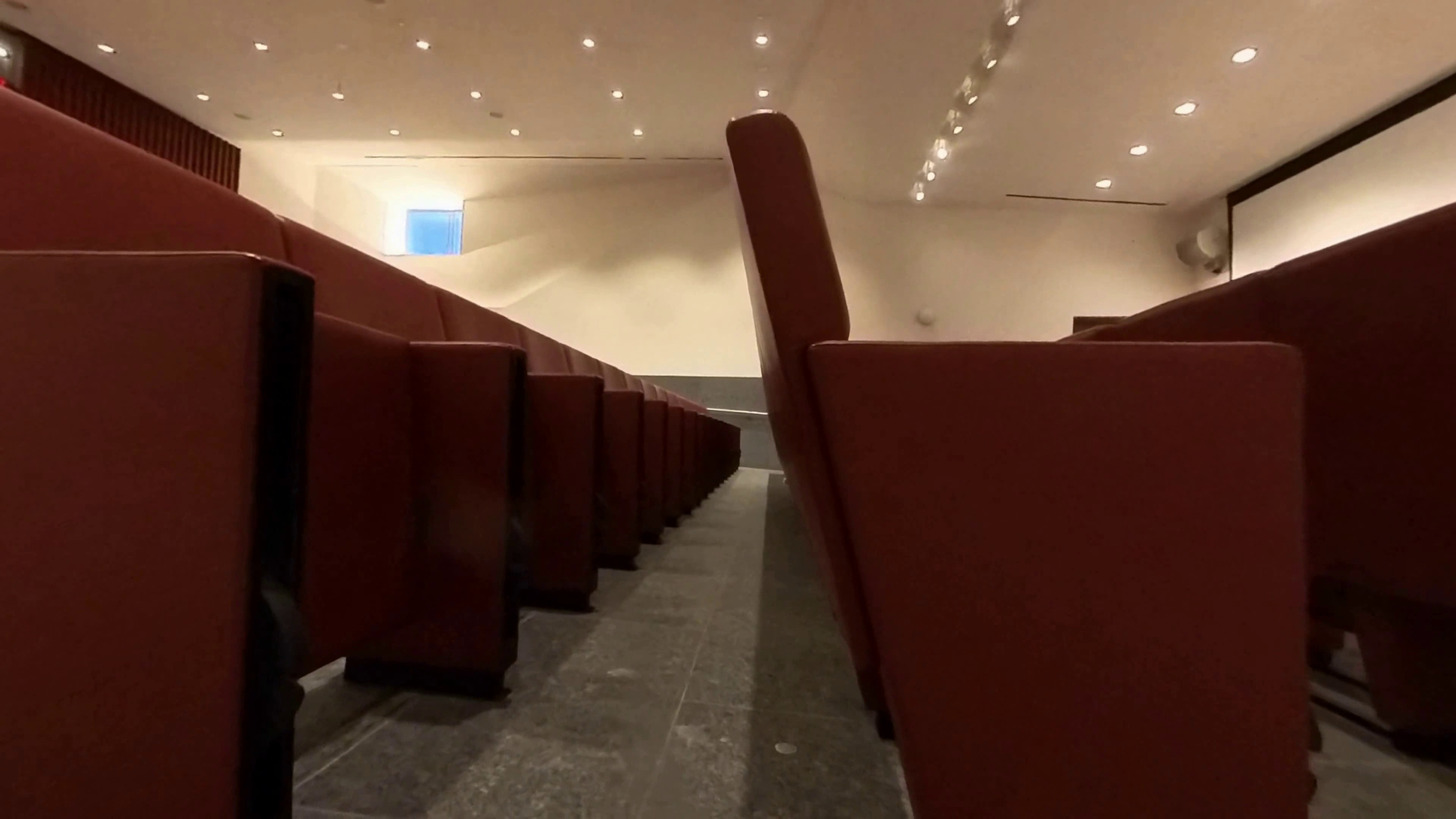} &
    \includegraphics[width=0.19\textwidth]{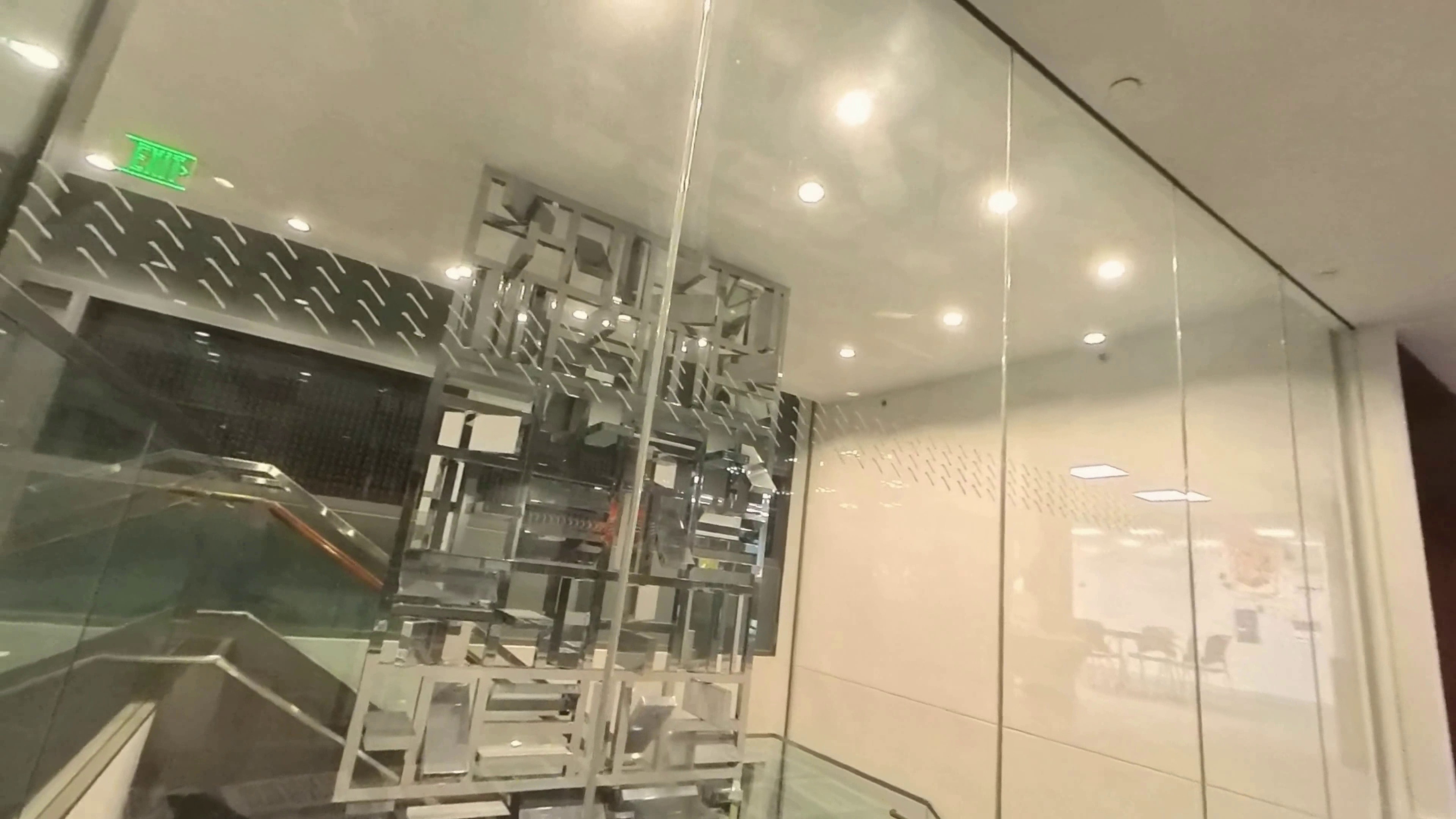} &
    \includegraphics[width=0.19\textwidth]{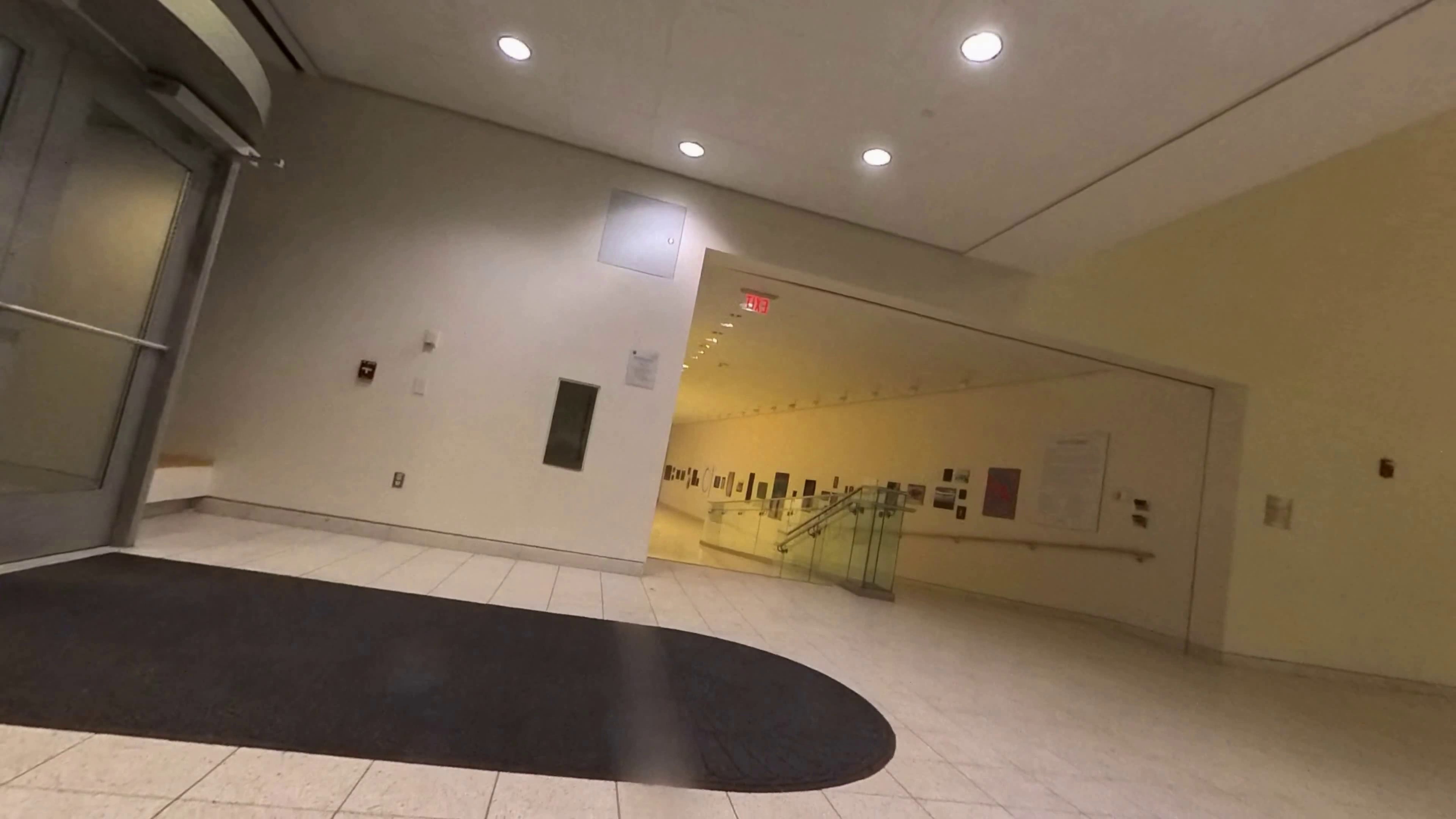} \\
    
     \raisebox{0.25\height}{\rotatebox{90}{Outdoors}} &
    \includegraphics[width=0.19\textwidth]{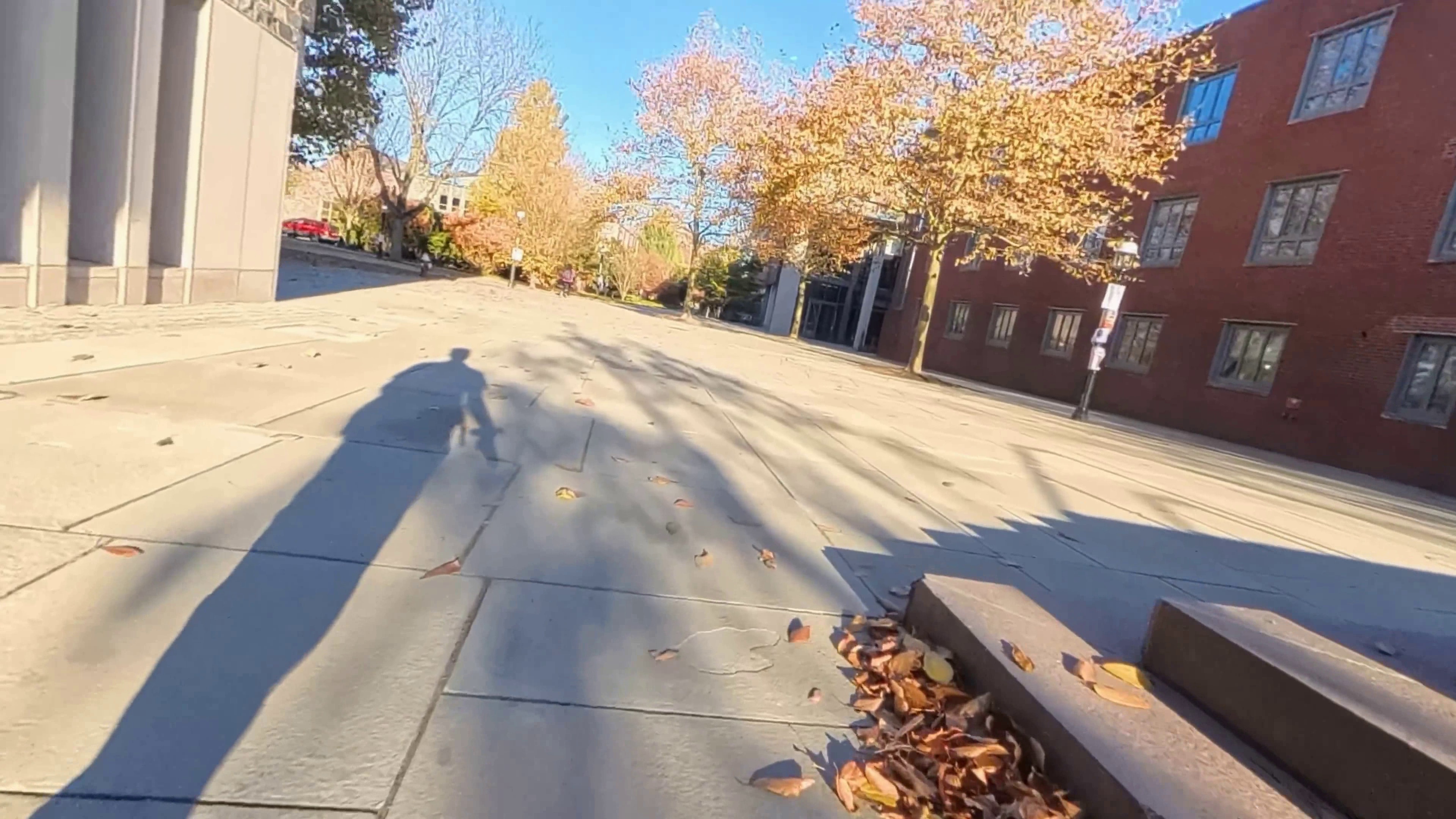} &
    \includegraphics[width=0.19\textwidth]{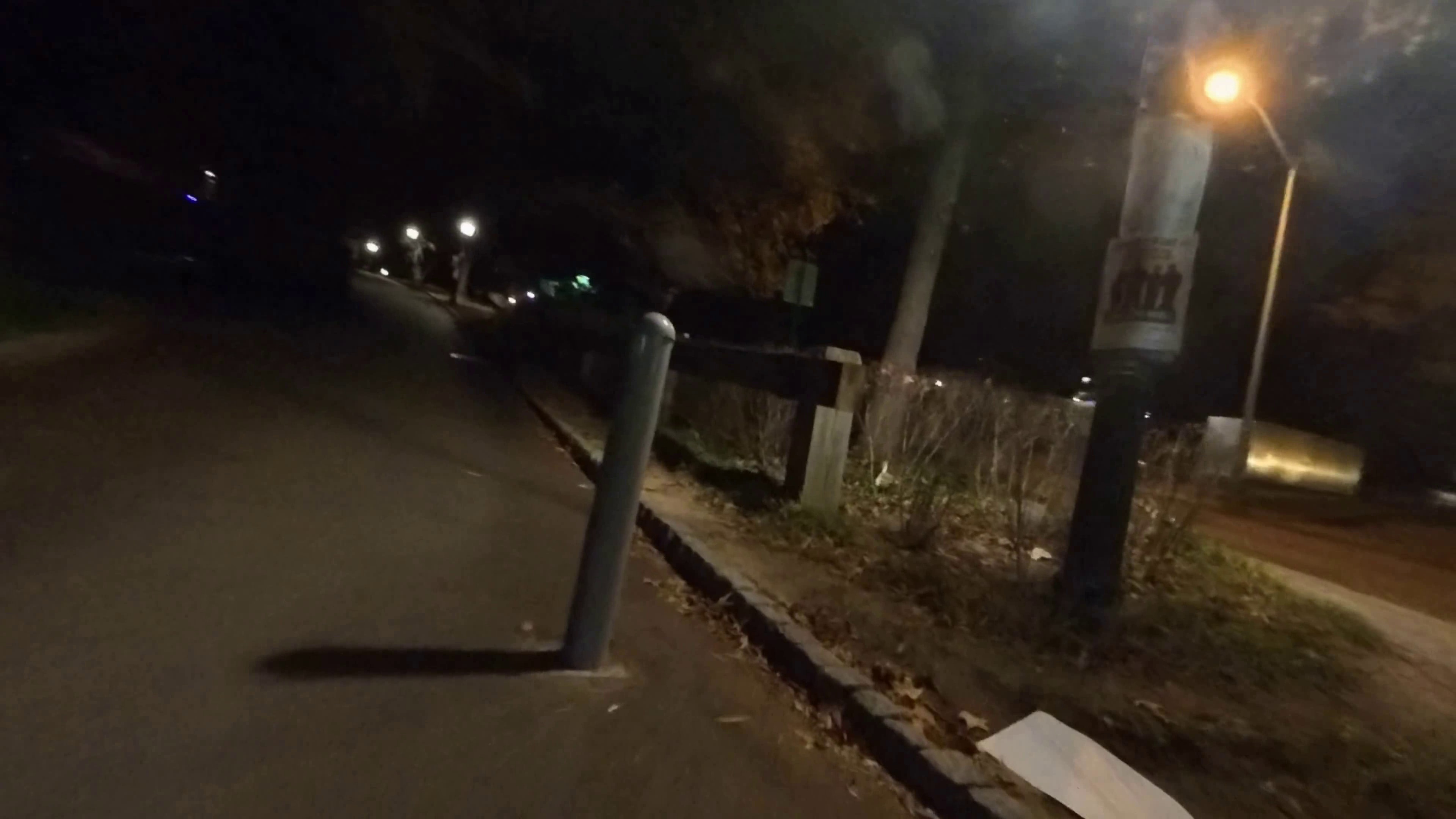} &
    \includegraphics[width=0.19\textwidth]{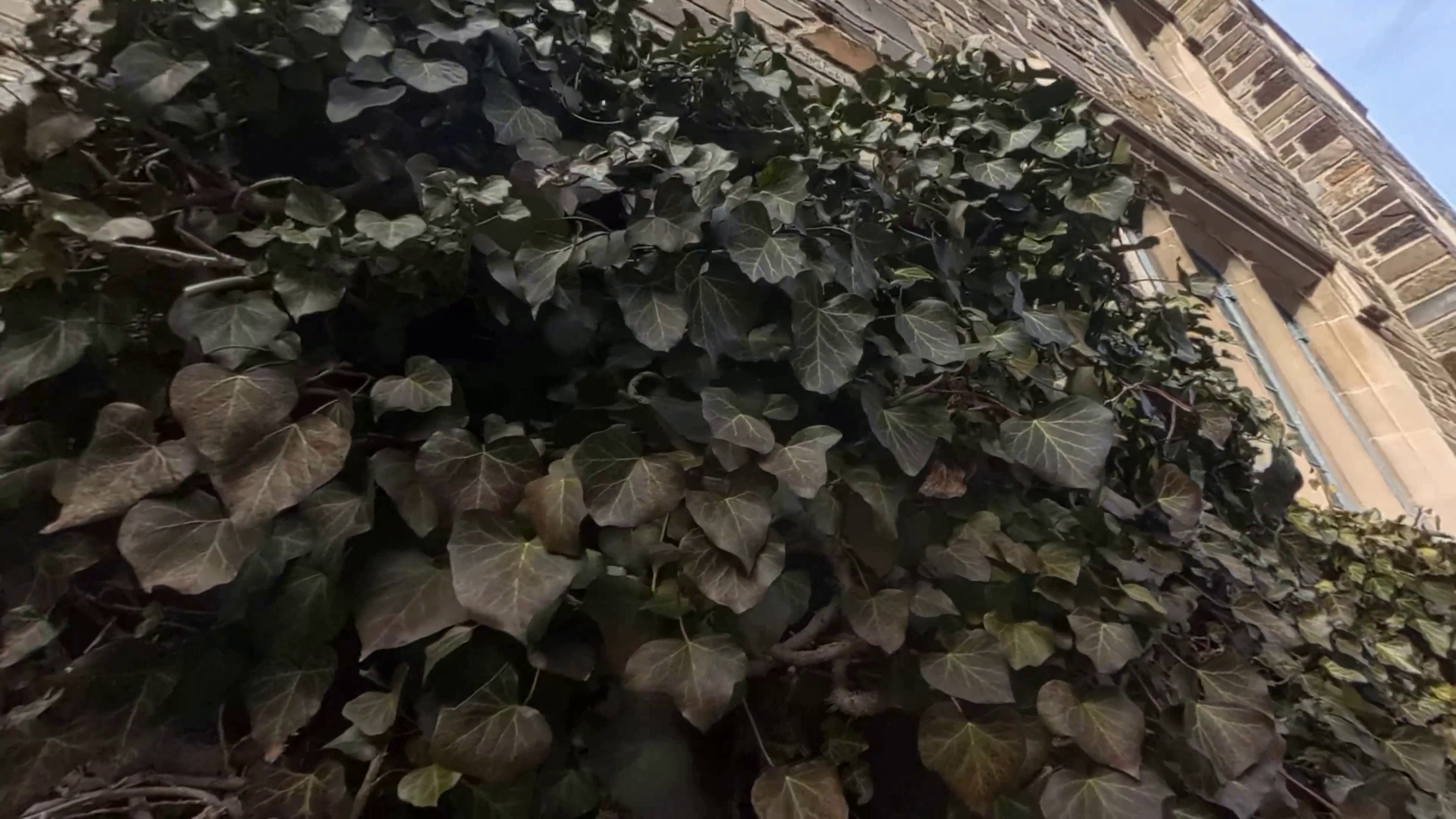} &
    \includegraphics[width=0.19\textwidth]{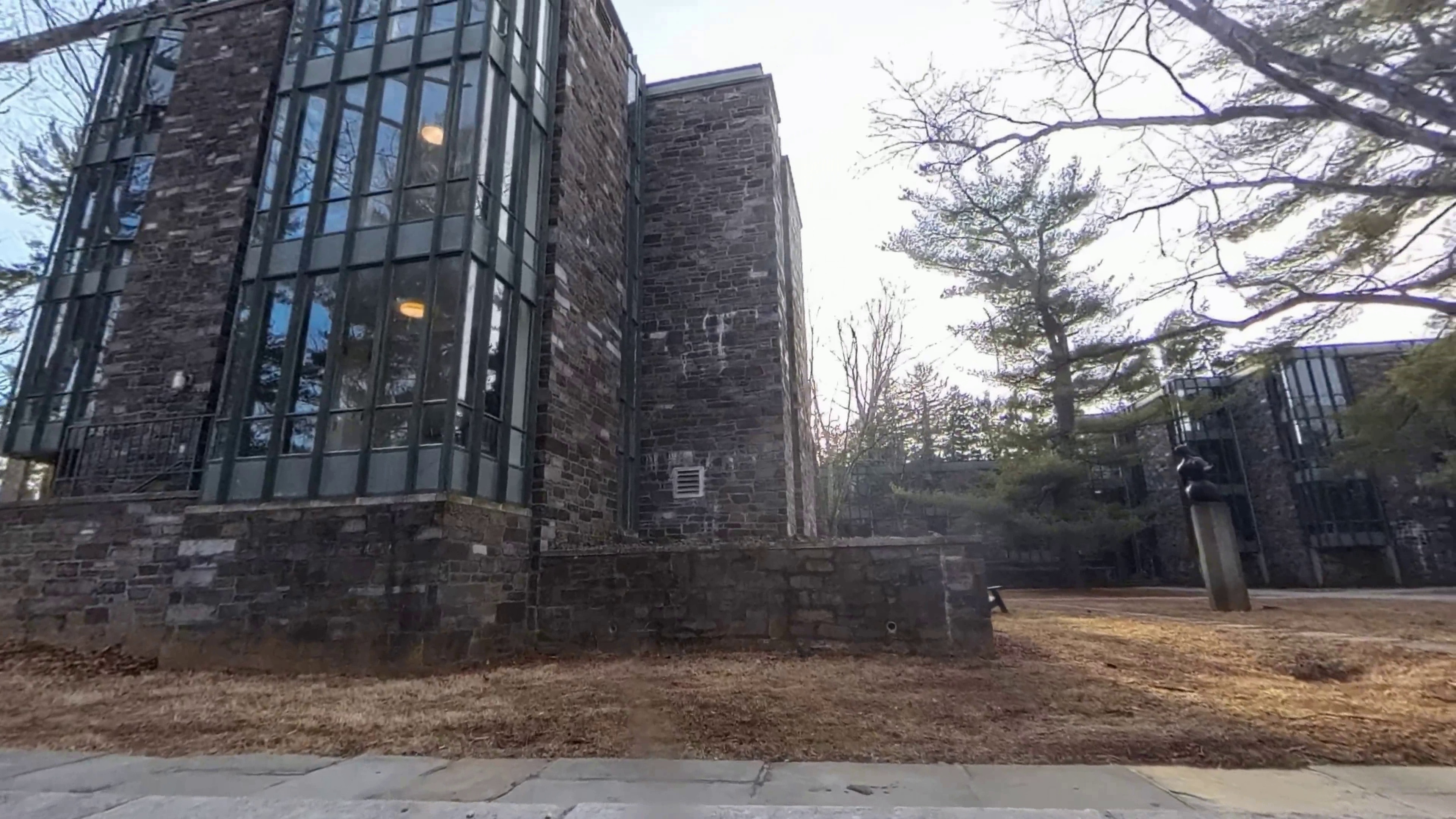} &
    \includegraphics[width=0.19\textwidth]{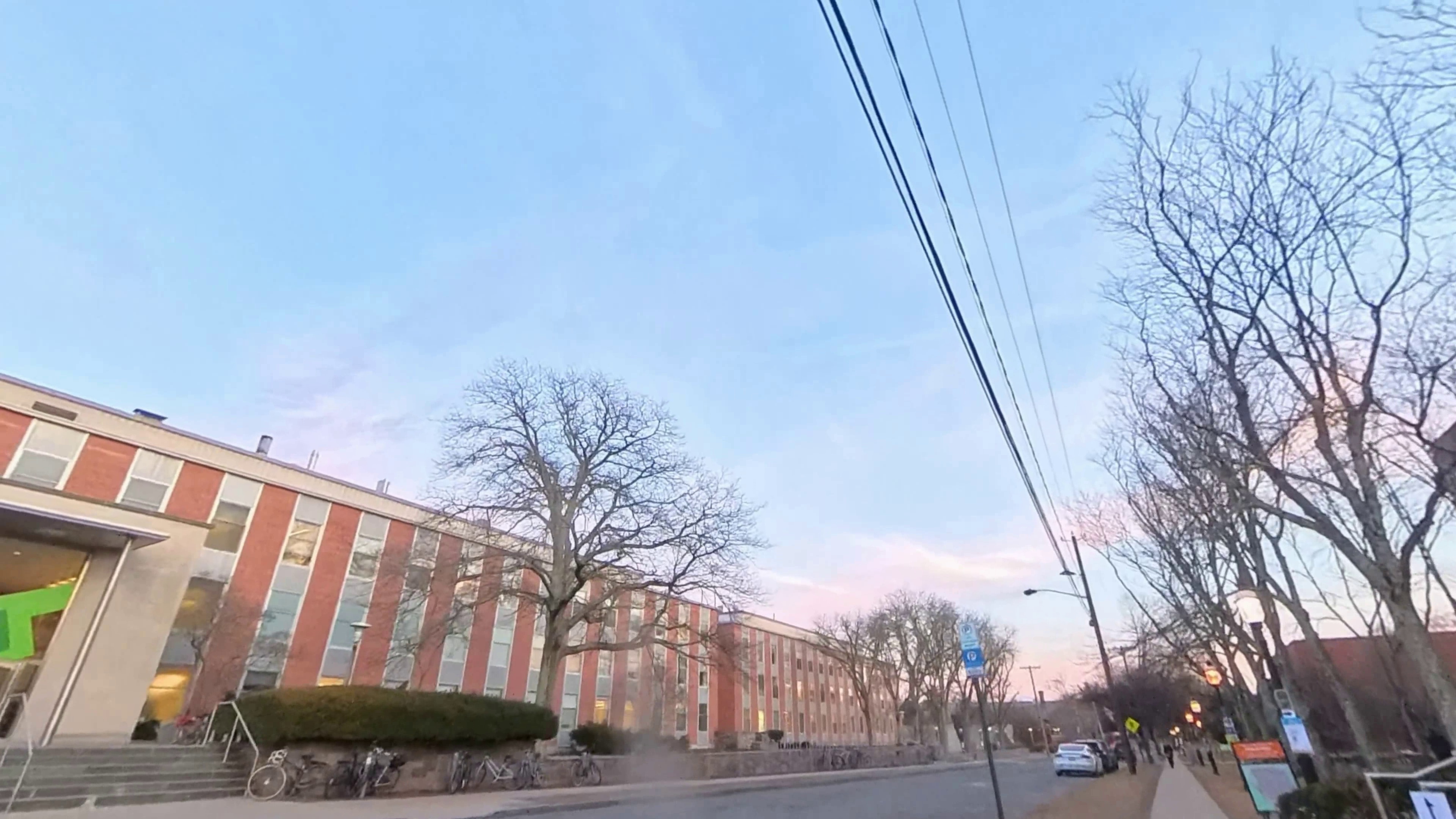} \\
  \end{tabular}
  \caption{Randomly picked frames from \projectname, consisting of three main categories: object scanning, long indoor videos and long outdoor videos. These include challenging cases such as reflective surfaces, dynamic objects, lighting changes, and fast camera motion.}
  \label{fig:example_frames}
\end{figure*}

\section{Introduction}
\label{sec:intro}

Simultaneous Localization and Mapping (SLAM) is the task of estimating a camera's 6-DoF trajectory and creating a 3D reconstruction of the scene using RGB video input. The 3D reconstruction of the scene can be in different formats such as point clouds \cite{lipson_deep_2024, teed_droid-slam_2022, mur-artal_orb-slam_2015}, Neural Radiance Fields \cite{zhu_nice-slam_2022, zhang_go-slam_2023, zhu_nicer-slam_2023}, or meshes \cite{ruan_slamesh_2023}. In contrast, the camera trajectory output is typically a sequence of SE(3) poses $\{T_i\}_{i=1}^T$. As a result, SLAM algorithms are generally evaluated on how well they reconstruct the camera trajectory compared to a ground-truth camera trajectory. 

The issue with current SLAM benchmarks is that their ground-truth camera trajectories are either inaccurate or not diverse. For instance, the TUM RGB-D \cite{tumrgbd} benchmark has millimeter accuracy ground-truth collected with a Motion Capture (MoCap) system, but it is restricted to only two indoor environments. On the other hand, the Michigan NCLT dataset \cite{carlevaris-bianco_university_2016} covers 147.4 km of both indoor and outdoor scenes across different seasons and times of day. However, it has inaccurate ground truth trajectories with a 10 cm error margin. 

The reason why current SLAM benchmarks have to make an accuracy-diversity tradeoff is that accurate ground truth collection methods either do not scale or are restricted to particular scenes. For instance, MoCap is a common approach that provides mm camera pose accuracy, but requires a cumbersome setup every time the environment is changed, preventing data collection from scaling. Furthermore, it generally does not work outdoors \cite{ramezani_newer_2022}, preventing data collection in outdoor environments. While RTK-corrected GPS provides centimeter level positional accuracy outdoors, it does not work indoors or near buildings \cite{carlevaris-bianco_university_2016, ramezani_newer_2022}. Structure-from-Motion (SfM) or IMU based ground truth collection methods are scalable because they do not require a special setup. However, they are inaccurate and can bias the evaluation of SLAM methods \cite{brachmann_limits_2021}. 

Additionally, large-scale benchmarks have restricted camera motion. For instance, KITTI \cite{kitti}, Malaga Urban \cite{malaga}, and Michigan NCLT \cite{michigannc} cover large distances with long recordings. However, the rotation of their camera is restricted to one axis since the camera is mounted on cars or wheeled robots. This means that SLAM methods are evaluated on 4-DoF camera trajectories in such benchmarks instead of a full 6-DoF movement. Handheld datasets such as Newer College Dataset \cite{ramezani_newer_2022}, TUM RGB-D \cite{tumrgbd}, and TUM VI \cite{tumvi} do not suffer from this problem. However, their ground-truth collection methods are not scalable and restricted to small environments.  

We propose a novel ground-truth collection method that allows us to collect a large-scale dataset with accurate camera pose and full 6-DoF camera motion. Our method requires about five minutes to set up data collection in a completely new environment. We simply place calibration boards with uniquely identifiable markers and capture a video with a $360^{\circ}$ camera where we render two views: the \emph{user-view}, which records the trajectory, and the \emph{ground-truth view}, which sees the calibration boards. The user-view is provided to the SLAM/NVS methods and the ground-truth view is used to obtain the ground-truth camera poses. The user-view does not see any calibration boards, which prevents the SLAM methods from cheating and allows us to record normal scenes without introducing bias. See \cref{fig:hardware} for our data collection setup. 

We obtain the ground-truth camera poses by a novel optimization-based method using the 2D-3D correspondences between calibration markers and their 2D detections. We first construct a pose graph of calibration boards, and then initialize the camera poses by leveraging the Perspective-n-Point (PnP) algorithm on multiple boards. We improve these initializations by using median-based edge selection, pose graph optimization, and encoding a planarity prior to the board poses. We then process the initialized camera poses by an algorithm that we call \emph{Bundle PnP} that is analogous to bundle adjustment, but utilizes the physical constraints on the local 3D marker coordinates. See \cref{fig:gt_framework} for an overview of our method.

A separate issue with current SLAM benchmarks is that they evaluate based on metrics that do not factor in the scene scale. The most popular metric is called Average Trajectory Error (ATE) and it calculates the average positional error after aligning the estimated trajectory to the ground-truth trajectory. However, nearby objects should make it easier for the camera to be tracked than further away objects due to parallax. Furthermore, an ATE of 10 cm has a greater impact on applications like augmented reality (AR) when filming nearby objects compared to filming a large scene outdoors. As a result, it is hard for SLAM researchers to compare the performance of their method across trajectories with different scene scales. We propose a new evaluation metric based on the magnitude of the optical flow induced by the error in camera pose estimation. Our metric factors in the scene scale and allows for universal comparison between trajectories with different scene scales.

Our ground-truth collection method also allows us to collect a unique novel view synthesis (NVS) benchmark and address a long-standing issue with the current NVS benchmarks. Existing benchmarks tend not to cover scenes with a large number of non-Lambertian objects, even though such objects are quite common in real life and interesting from an NVS perspective. This is because most NVS benchmarks run COLMAP \cite{colmap} to get the ground-truth camera poses, but COLMAP does not work well with non-Lambertian scenes \cite{ge_ref-neus_2023}. In contrast, our ground-truth method is not affected by non-Lambertian objects being present in the user view since it obtains the ground-truth from the ground-truth view, which only sees calibration boards.  

We thus propose a challenging NVS benchmark consisting of six sequences from \projectname shown in \cref{fig:nvs_benchmark}. We include scenes with $360^{\circ}$ scans around Non-Lambertian setups due to the interest in the NVS community for such scenes. To our knowledge, our benchmark is the first NVS benchmark that includes scenes filmed with $360^{\circ}$ camera trajectories and include a large density of non-Lambertian objects. Note that all 365 sequences in \projectname can be used for NVS evaluation. We provide an option for both COLMAP and our ground-truth in terms of camera pose.  

\begin{table*}[ht]
\centering
\resizebox{\textwidth}{!}{
    \begin{tabular}{ccccclccccc}
    \toprule
    Benchmark & \# Indoor  & \# Outdoor & \# Scanning & Total & Ground Truth Method & \# DOF & Accuracy & Frames & Dist. Covered & Duration \\ 
    \midrule
    ETH-3D \cite{eth3d} & 2 & 3 & 91 & 96 & MoCap, SfM & 6 & -- & 104K & -- & 1h 5m \\
    KITTI \cite{kitti} & 0 & 22 & 0 & 22 & GPS, IMU, RTK & 4 & 10 cm & 41K & 39.2 km & 1h 8m \\
    EuRoC MAV \cite{euroc} & 11 & 0 & 0 & 11 & MoCap, Laser tracker, 3D scan & 6 & 1 mm & 27K & 0.9 km & 23m \\
    Malaga Urban \cite{malaga} & 0 & 15 & 0 & 15 & Stereo, Laser scan, IMU, GPS & 4 & -- & 113K & 36.8 km & 1h 33m \\
    TUM RGB-D \cite{tumrgbd} & 39 & 0 & 0 & 39 & MoCap, IMU & 6 & 0.86 mm & 87K & 543 m & 49 m \\
    Rawseeds \cite{rawseeds} & 5 & 6 & 0 & 11 & GPS, Laser scanner & 4 & few cm/m & -- & -- & -- \\
    PennCOSYVIO \cite{penncosyvio} & 2 & 2 & 0 & 4 & SfM, IMU & 6 & 15 cm & 16K & 600 m & 9m \\
    TUM VI Benchmark \cite{tumvi} & 17 & 11 & 0 & 28 & IMU, Motion capture & 6 & mm & -/290K & -/20 km & -/4h 1m \\
    Zurich Urban MAV \cite{zurichurban} & 0 & 1 & 0 & 1 & SfM, IMU, SLAM & 6 & -- & 81K & 2 km & 1h 7m \\
    Ford Multi-AV \cite{fordav} & 0 & 18 & 0 & 18 & GPS, IMU, SLAM, LiDAR & 4 & cm & -- & -- & -- \\
    Michigan North Campus \cite{michigannc} & 13 & 14 & 0 & 27 & IMU, RTK, LiDAR, GPS & 4 & 10 cm & 628K & \textbf{147.4 km} & \textbf{34.9h} \\
    The Newer College Dataset \cite{newcollege} & 0 & 6 & 0 & 6 & IMU, LiDAR, Laser scanner & 6 & 3 cm & 69K & 2.2 km & 38m \\
    \midrule
    \textbf{\projectname} & \textbf{40} & \textbf{40} & \textbf{285} & \textbf{365} & $360^\circ$ camera, Calibration Boards & \textbf{6} & \textbf{mm} & 1.1 M/\textbf{2.1 M} & 4.3km/26 km & 5h 32m/9h 51m \\
    \bottomrule
    \end{tabular}
}
\caption{Comparisons to existing benchmarks in terms of the number of sequences, ground-truth method, how many degrees of freedom the camera movement has, ground-truth accuracy, number of frames, distance covered, and duration. x/y denotes numbers with pose and total respectively. Dashes represent numbers we could not acquire or estimate. We were not able to obtain posed numbers for TUM-VI, so we only report total numbers. Many existing benchmarks are either inaccurate or not diverse enough since they are restricted to the same filming environment. \projectname is diverse, accurate, and has 6-DoF camera pose.}
\label{tab:related_work}
\end{table*}

Our contributions can be summarized as follows:

\begin{itemize}
    \item We collect a large-scale, diverse dataset of videos with accurate ground-truth camera pose and 6 DoF camera motion involving indoor, outdoor, and object scanning scenes. Our benchmark consists of 365 unique sequences in total with monocular, stereo, and IMU output.
    \item We propose a novel data collection pipeline as well as an optimization method for ground-truth camera pose that allows our dataset to be millimeter accurate, easily scalable, and diverse. 
    \item We propose a new optical-flow based evaluation metric that allows universal comparison between trajectories of different scene scales.
    \item We leverage our pipeline to curate a challenging NVS benchmark that involves important cases that are not covered by the current benchmarks such as $360^{\circ}$ scans of non-Lambertian objects. 
\end{itemize}

\section{Related Work}
\label{sec:related}

Modern SLAM methods consist of classical methods such as ORB-SLAM \cite{mur-artal_orb-slam_2015} and BAD-SLAM \cite{Schps2019BADSB} along with neural network based methods such as DROID-SLAM \cite{teed_droid-slam_2022}, DPV-SLAM \cite{lipson_deep_2024}, NICE-SLAM \cite{zhu_nice-slam_2022}, and LEAP-VO \cite{chen2024leap}. Our goal is to construct a benchmark that will be appropriately challenging to these methods. We further aim to provide insights to researchers who develop these methods by allowing for comparisons between different camera trajectories with our new Induced Optical Flow (IOF) error metric. Note that our dataset can also be used to evaluate SfM methods such as COLMAP \cite{colmap}, and DUSt3R \cite{wang_dust3r_2024}.  

\cref{tab:related_work} shows a comparison of our dataset to the current benchmarks with ground-truth camera pose. Existing SLAM benchmarks include large-scale datasets such as KITTI \cite{kitti}, Michigan NCLT \cite{michigannc}, Malaga Urban \cite{malaga}, and Rawseeds \cite{rawseeds}. These datasets cover large distances with long videos since they mount cameras on cars and wheeled robots. However, they do not have full 6-DoF camera movement since the camera can only rotate in one axis. Furthermore, their ground-truth poses have cm accuracy, which is not as accurate as the smaller-scale counterparts. \projectname is three times larger than the largest of these datasets in terms of the number of frames, is more accurate with mm accuracy, and has full 6-DoF camera movement. 

The smaller scale benchmarks include TUM RGB-D \cite{tumrgbd}, ETH-3D \cite{eth3d}, EuRoC MAV \cite{euroc}, and the Newer College Dataset \cite{newcollege}. They usually have full 6-DoF hand-held camera movement and have around mm accuracy since they use methods like MoCap to obtain their ground-truth. However, they suffer from the drawbacks of MoCap as the sequences are constrained to one or two rooms where the MoCap rig is located and do not include outdoor scenes since MoCap does not work well outdoors. TUM-VI \cite{tumvi} captures longer videos by leaving the MoCap zone for intermediate frames and sacrificing ground-truth for those frames. However, the frames that have ground-truth are still constrained to one room since the sequences have to begin and end in the MoCap room. Therefore, the intermediate frames are also constrained to a radius around the MoCap room. \projectname is roughly 7 times larger than TUM-VI \cite{tumvi} in terms of the number of frames, has mm accuracy, and covers a more diverse set of scenes since the ground-truth zone is not constrained. 
\begin{table}[t]
\centering
\resizebox{\linewidth}{!}{
    \begin{tabular}{lcccc}
    \toprule
    Benchmark & \begin{tabular}{@{}c@{}}{Ground-truth} \\ {Pose Source} \end{tabular} & Scene Coverage & \begin{tabular}{@{}c@{}}{Non-Lambertian} \\ {Objects Density} \end{tabular} & \begin{tabular}{@{}c@{}}{Scene} \\ {Diversity} \end{tabular} \\ 
    \midrule

    Blender~\cite{nerf} & Synthetic & $360^\circ$ & * & Low \\

    DTU~\cite{DTU} & Robotic Arm & Forward-Facing & * & Low \\

    LLFF~\cite{LLFF} & COLMAP & Forward-Facing & * & High \\

    NeRFstudio~\cite{nerfstudio} & COLMAP & $360^\circ$ & * & High \\

    MipNeRF-360~\cite{mipnerf-360} & COLMAP & $360^\circ$ & * & Mid \\

    NeX~\cite{nerf_nex} & COLMAP & Forward-Facing & ** & High \\
    
    \midrule
    \textbf{\projectname} & Calibration Boards & \textbf{360°+ Forward Facing} & \textbf{***} & \textbf{High} \\
    \bottomrule
    \end{tabular}
}
\caption{Comparison of \projectname NVS to the existing NVS datasets. We include fully non-Lambertian scenes and 360° camera movement. The stars qualitatively indicate how much of the view is covered by non-Lambertian objects. }
\label{tab:related_nvs}
\end{table}

As seen in \cref{tab:related_nvs}, most existing NVS benchmarks such as MipNeRF-360 \cite{mipnerf-360}, NeRFstudio \cite{nerfstudio}, and LLFF \cite{LLFF} do not include a large density of non-Lambertian objects since they use COLMAP \cite{colmap} to obtain the camera pose ground-truth, and COLMAP struggles with non-Lambertian scenes. The NeX benchmark \cite{nerf_nex} includes non-Lambertian objects, but does not include scenes where the entire view is non-Lambertian like the top left image in \cref{fig:rebuttal-nonlambertian dynamic}. Furthermore, the camera motion in NeX is limited to forward-facing. In contrast, our \projectname NVS benchmark includes full 360 camera motion where the entire view is non-Lambertian.

\begin{table}[ht]
\centering
\resizebox{\columnwidth}{!}{%
\begin{tabular}{lcccc}
\toprule
Category & Scanning & Indoors & Outdoors & Total \\
\midrule
Total frames & 840,639 & 539,847 & 749,109 & 2,129,595 \\
Total duration & 3h 53m & 2h 29m & 3h 28m & 9h 51m \\
Posed frames (\%) & 98.5\% & 42.0\% & 18.8\% & 56.1\% \\
Posed dist. covered & 2,585 & 922 & 843 & 4,352 \\
Est. Total dist. (m) & 2,887 & 8,233 & 15,032 & 26,153 \\
Avg frames & 2,939 & 13,167 & 18,728 & 5,803 \\
Avg duration & 49.0s & 3m 39.4s & 5m 12.1s & 1m 36.7s \\
Avg dist. (m) & 10.10 & 200.81 & 375.81 & 71.26 \\
\bottomrule
\end{tabular}%
}
\caption{A breakdown of dataset statistics per scene category. \projectname has more than 2 million frames, nearly 10 hours of recording, and an estimated total distance coverage of 26 km. Total frames only count the $360^\circ$ camera frames and not the frames from the stereo camera. Posed frames show the percentage of frames that have ground-truth camera pose. The average numbers are averaged over the number of sequences.}
\label{tab:dataset_stats}

\end{table}

\section{Dataset Description}
\label{sec:benchmark}

As shown in \cref{tab:dataset_stats}, our dataset includes 2,129,595 frames, 9 hours 51 minutes of recording, 4352 meters of distance covered with ground-truth pose, and an estimated 26 km of distance covered in total (See \cref{sec:dist_est} for how this is calculated). $56.1\%$ of the frames have ground-truth pose.

\begin{figure}[t]
\centering
\setlength{\tabcolsep}{0pt} %
\renewcommand{\arraystretch}{0} %
\begin{tabular}{ccc}
\includegraphics[width=0.33\linewidth]{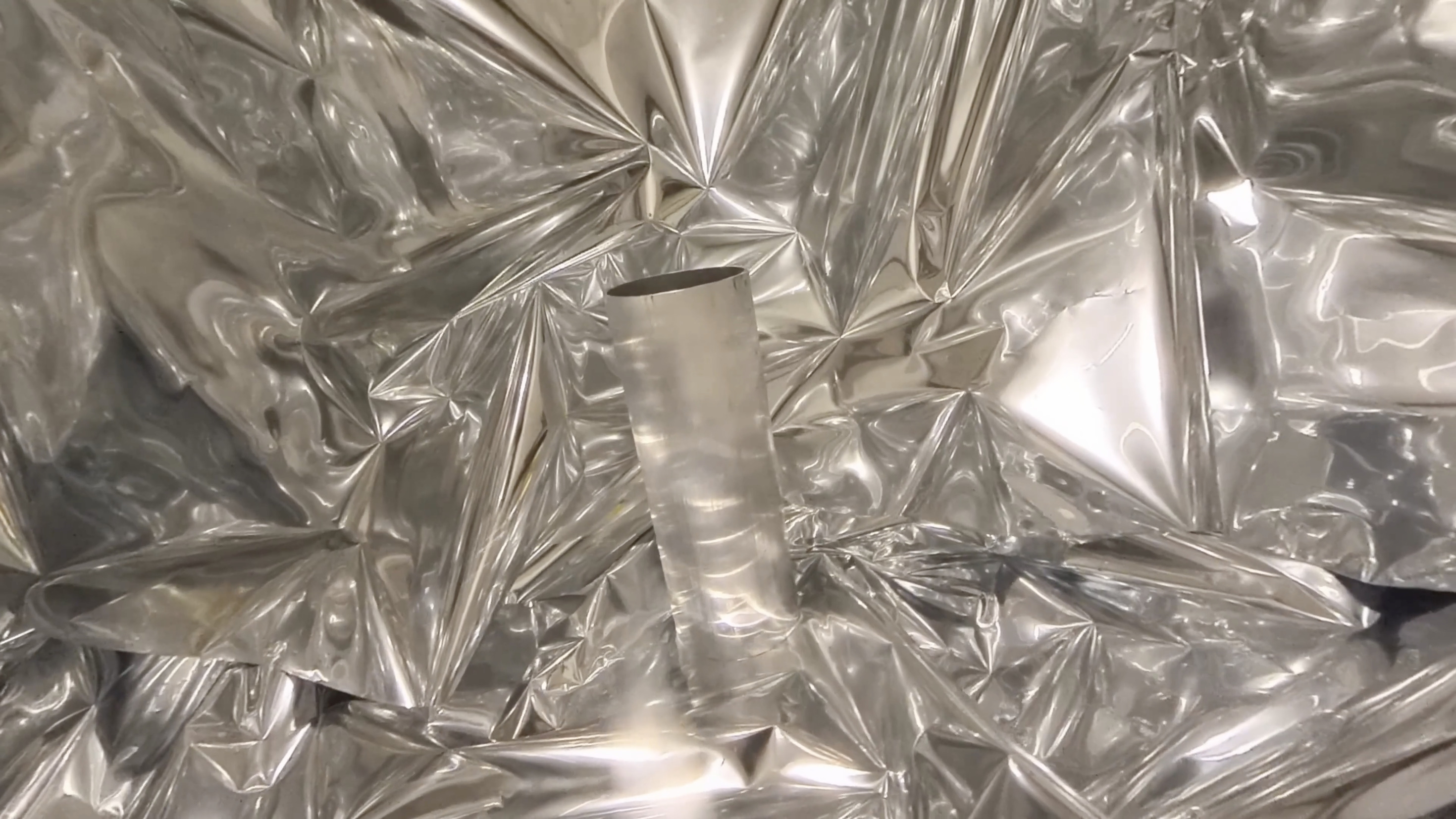}&
\includegraphics[width=0.33\linewidth]{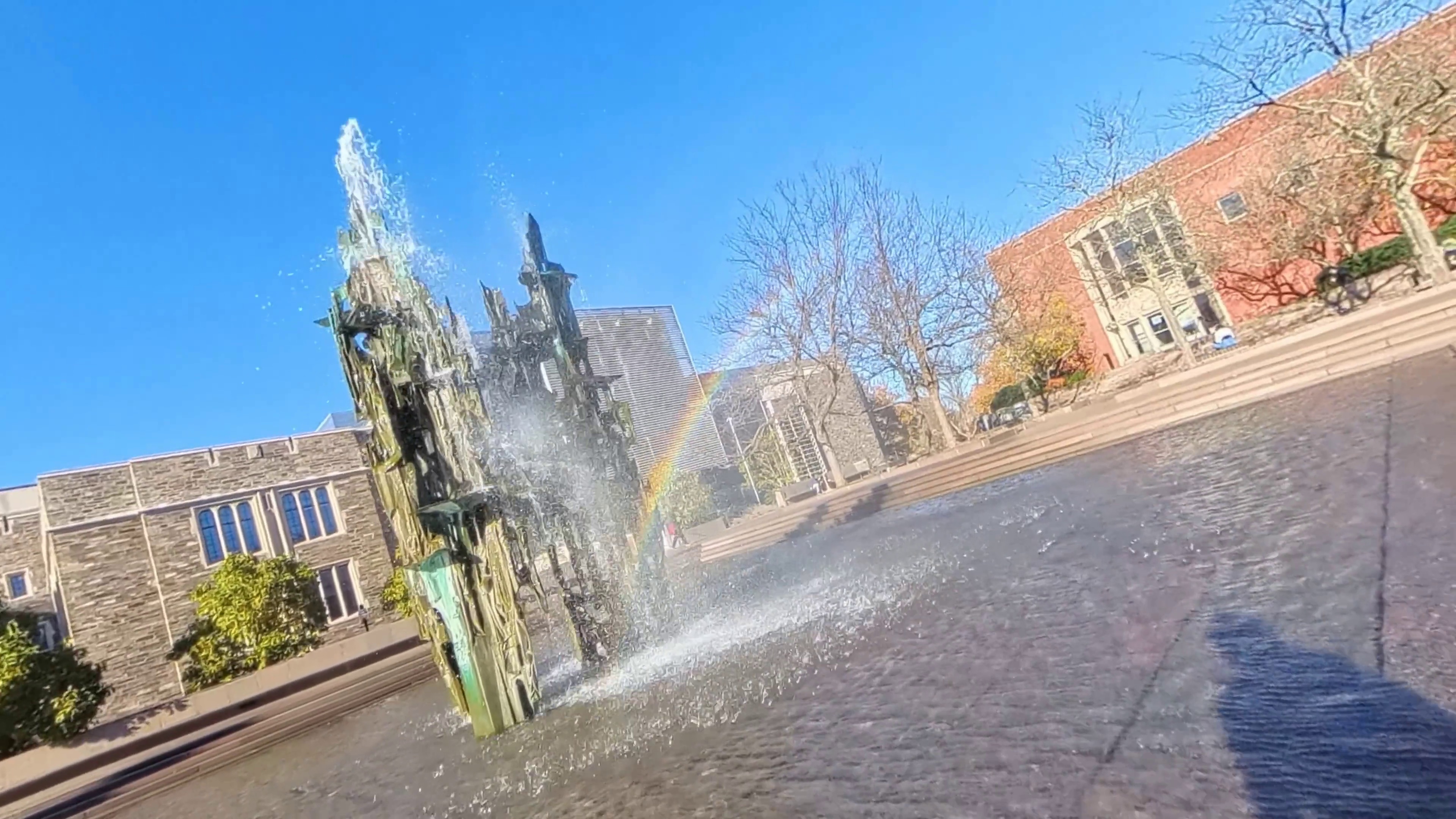} &
\includegraphics[width=0.33\linewidth]{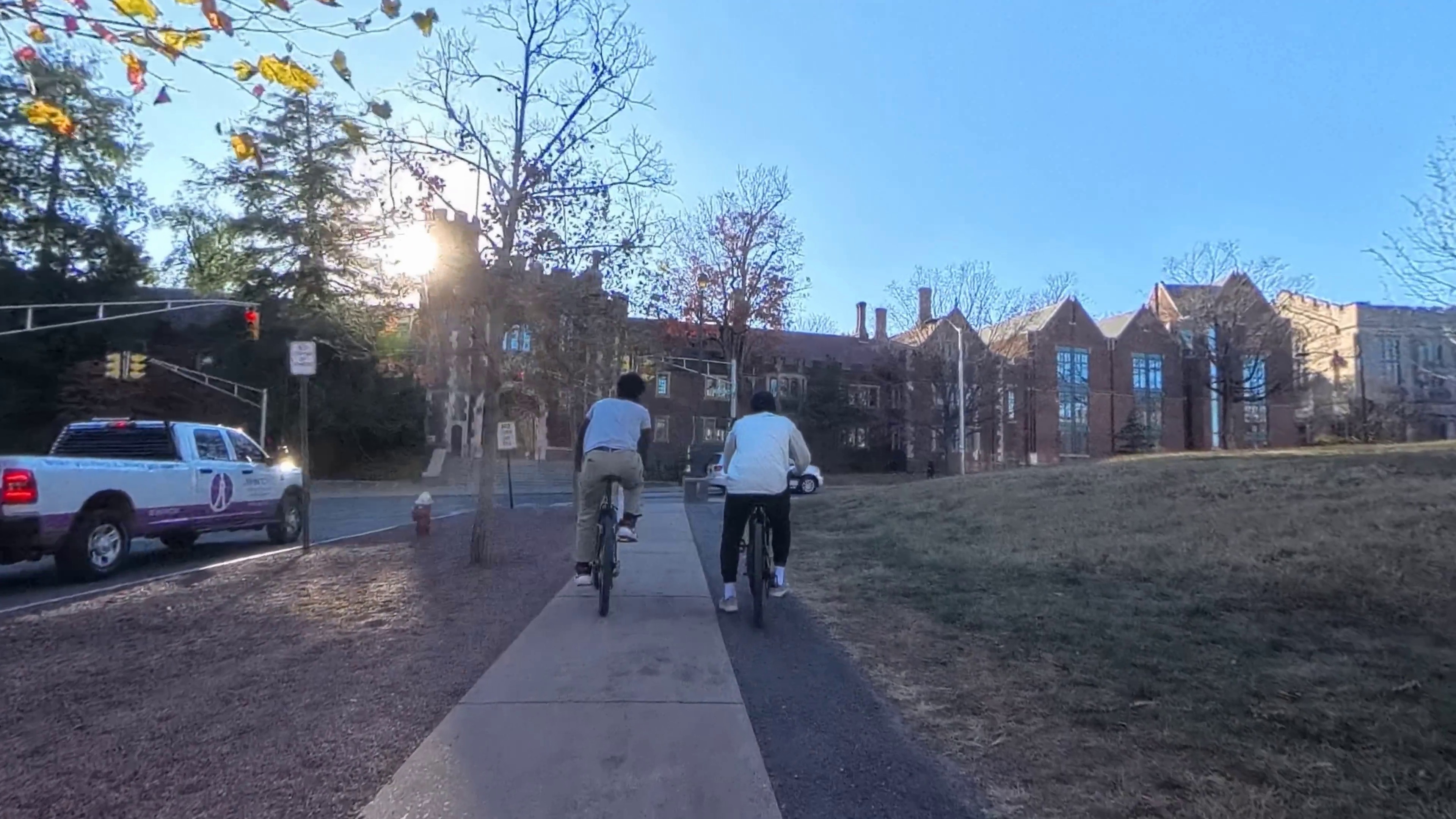} \\
\includegraphics[width=0.33\linewidth]{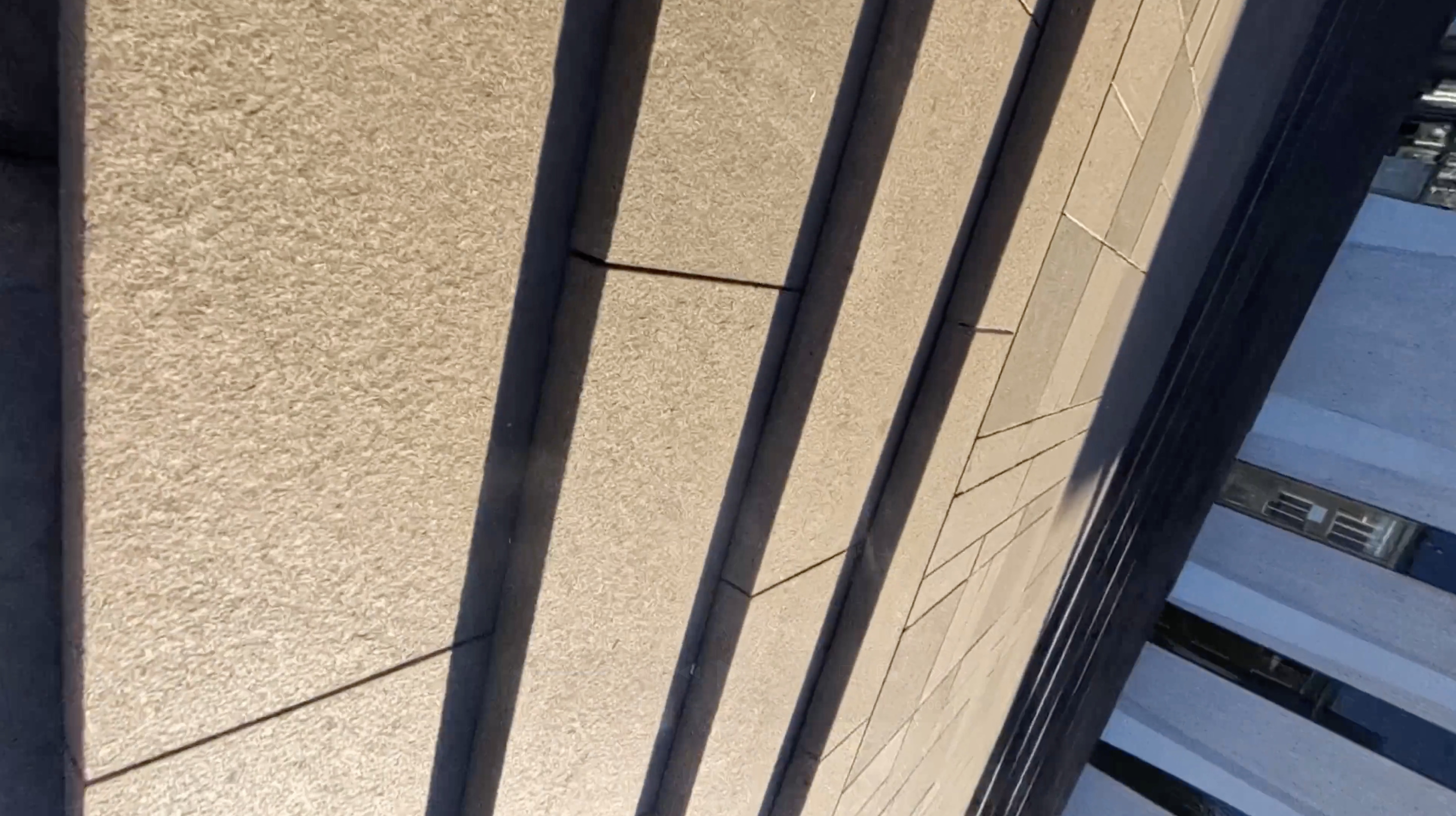} &
\includegraphics[width=0.33\linewidth]{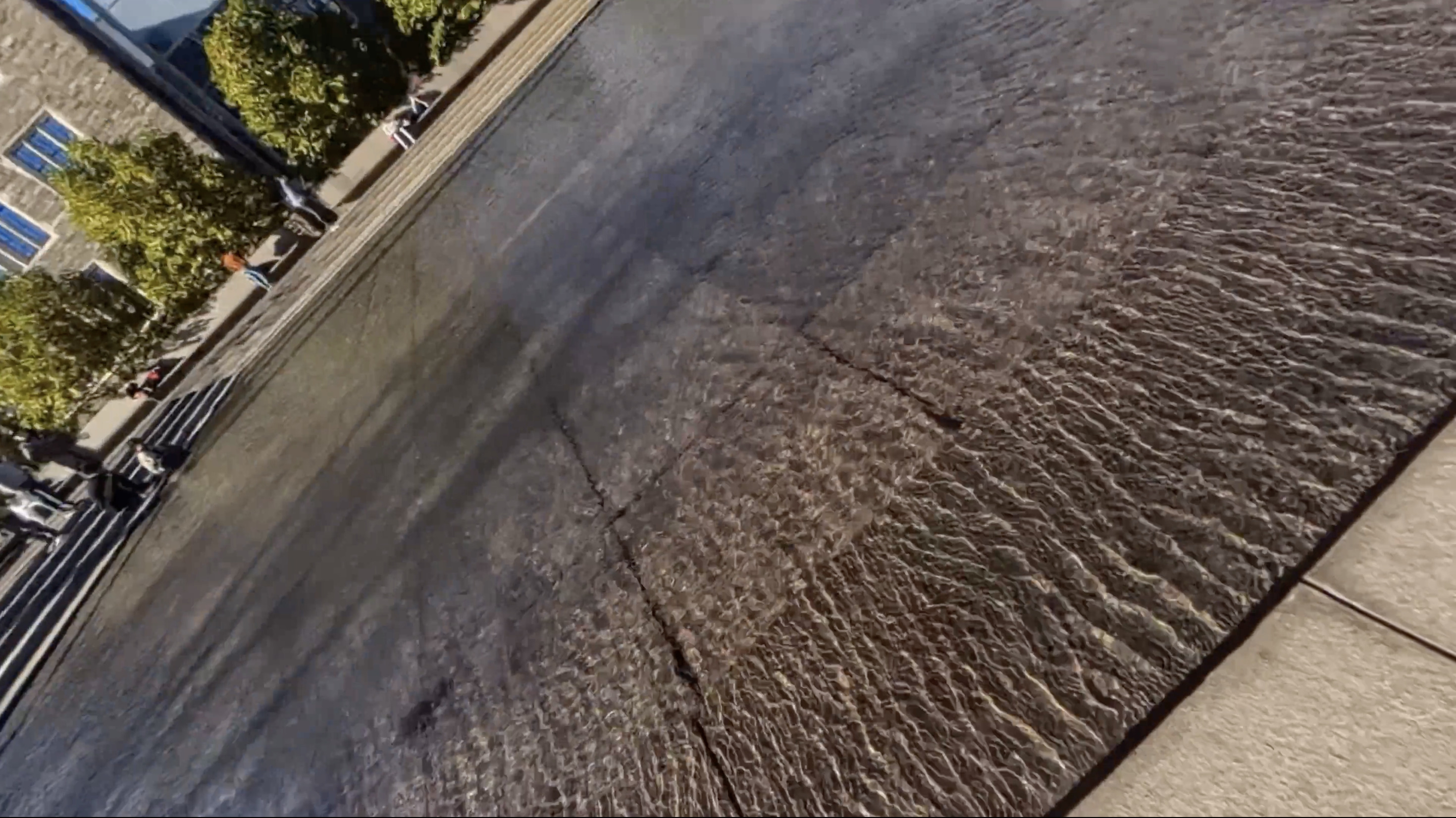} &
\includegraphics[width=0.33\linewidth]{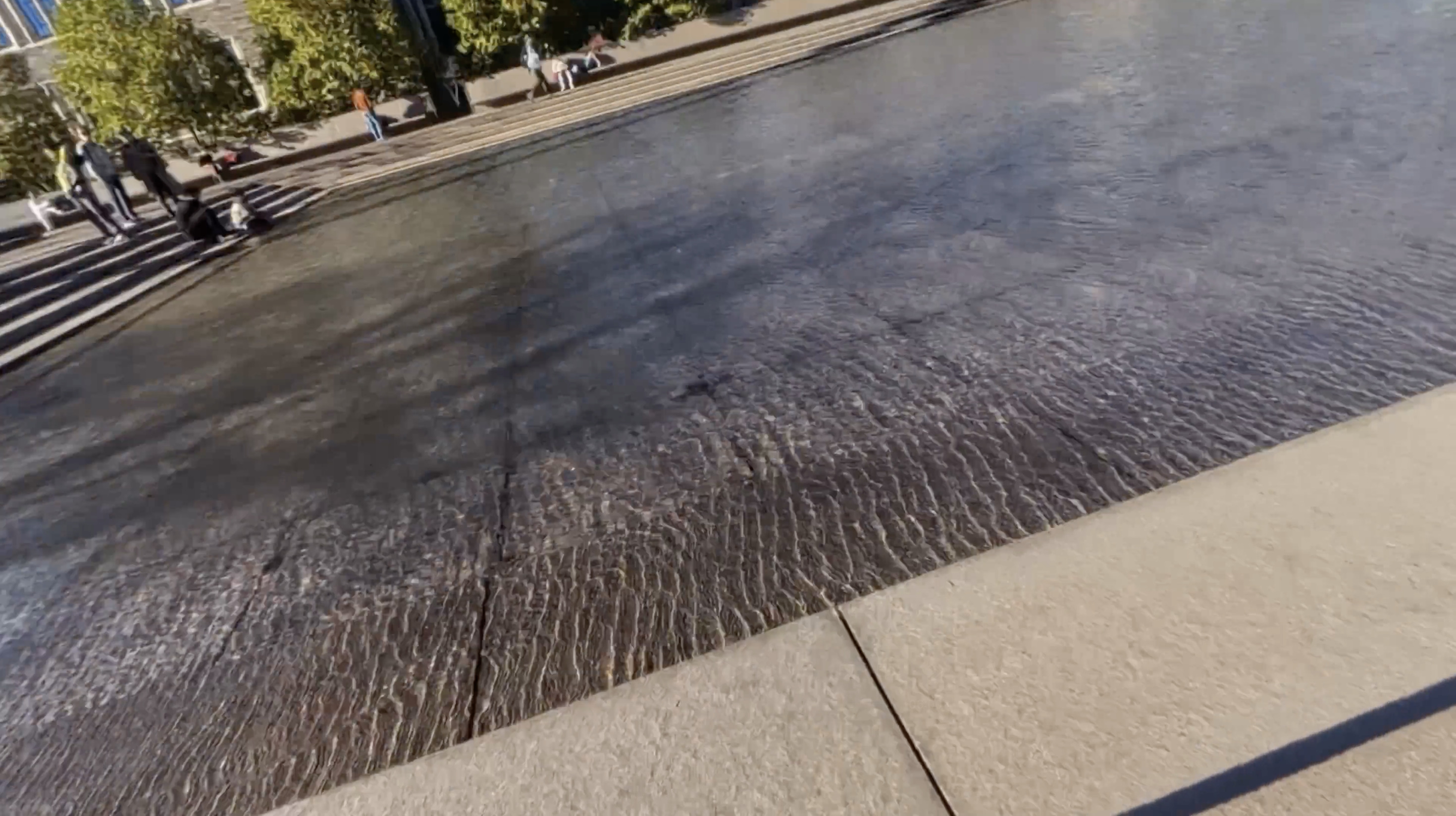} \\
\end{tabular}
\caption{Top row: Non-Lambertian and dynamic scenes. Top left: a camera view completely covered by non-Lambertian objects. Top right: moving car and bikers. Bottom row: large camera motion showing successive frames 1 second apart. In indoor and outdoor sequences, we specifically move the camera rig in interesting ways to make videos more challenging for SLAM methods. }
\label{fig:rebuttal-nonlambertian dynamic}
\end{figure}

\xpar{Camera Motion} We use 50 calibration boards, which have a coverage radius of around 20 m. While scanning videos stay within this zone, we adopt a similar approach to TUM-VI \cite{tumvi} for longer indoor and outdoor videos. We start within the ground truth zone, then venture beyond it before returning. This approach allows us to film longer trajectories while providing ground-truth only for the beginning and end frames. We evaluate methods on the posed frames and film such that beginning and end frames do not share views to prevent loop closure at the end. We try to prevent fast camera motion within the ground-truth zone in order to preserve the accuracy of the ground-truth. However, we specifically try to move the camera in interesting ways while walking at a fast pace outside of the ground-truth zone in order to provide a challenging task for SLAM methods (see \cref{fig:rebuttal-nonlambertian dynamic} bottom row). We rotate the camera in all three axes both within the ground-truth and outside it.  

\begin{figure}[t]
\centering
\setlength{\tabcolsep}{0pt} %
\renewcommand{\arraystretch}{0} %
\begin{tabular}{ccc}
\includegraphics[width=0.33\linewidth]{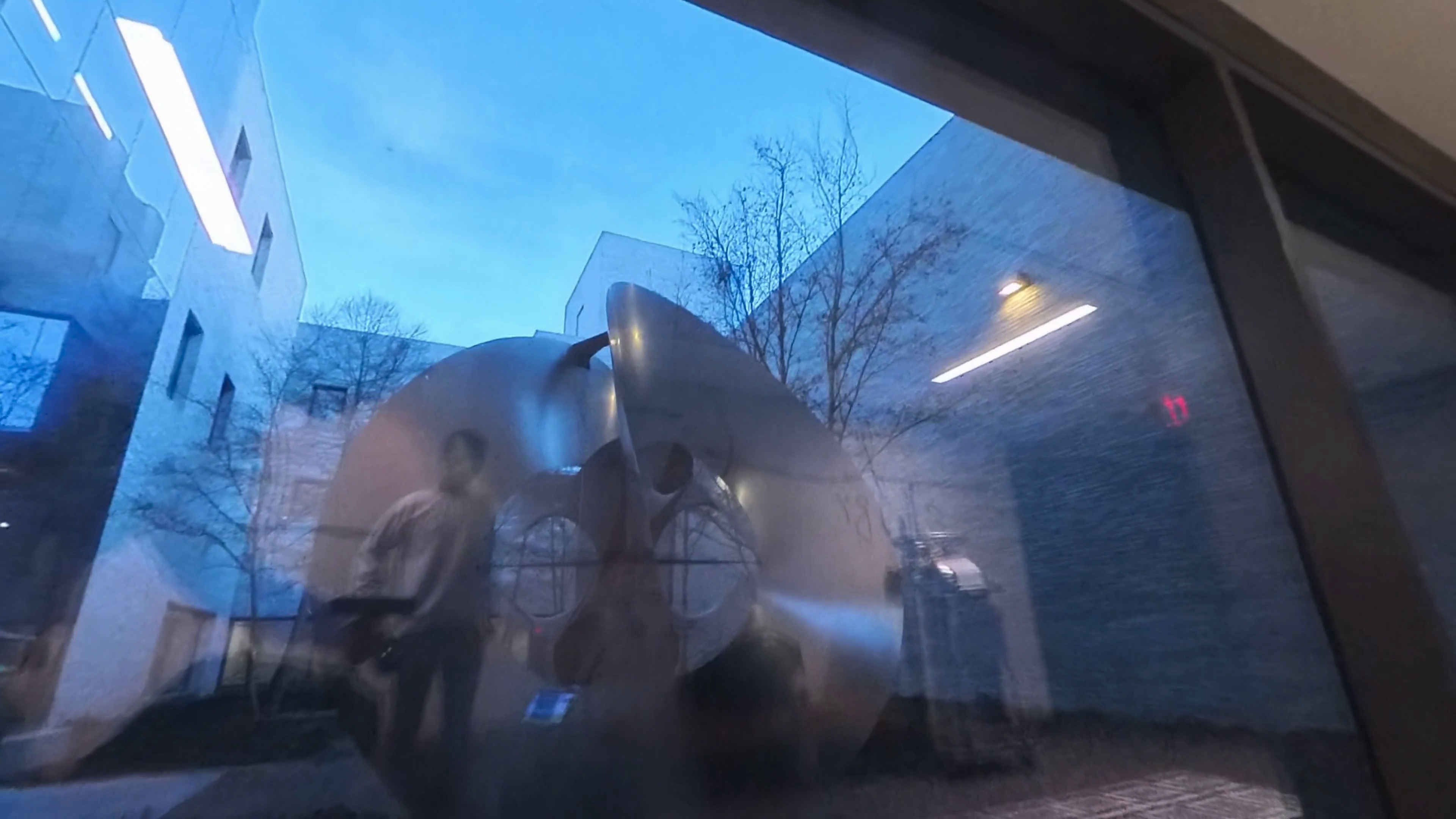}&
\includegraphics[width=0.33\linewidth]{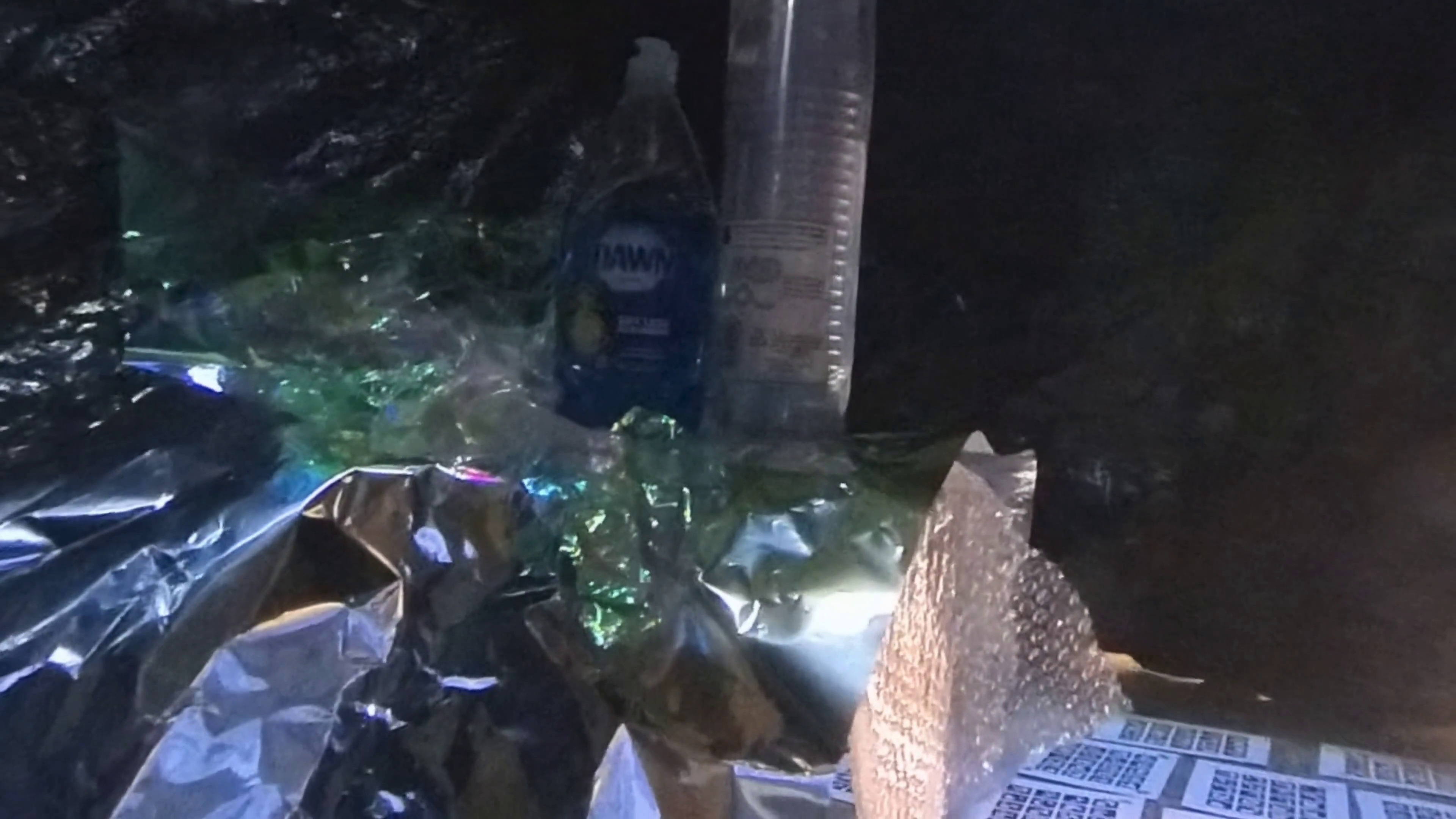} &
\includegraphics[width=0.33\linewidth]{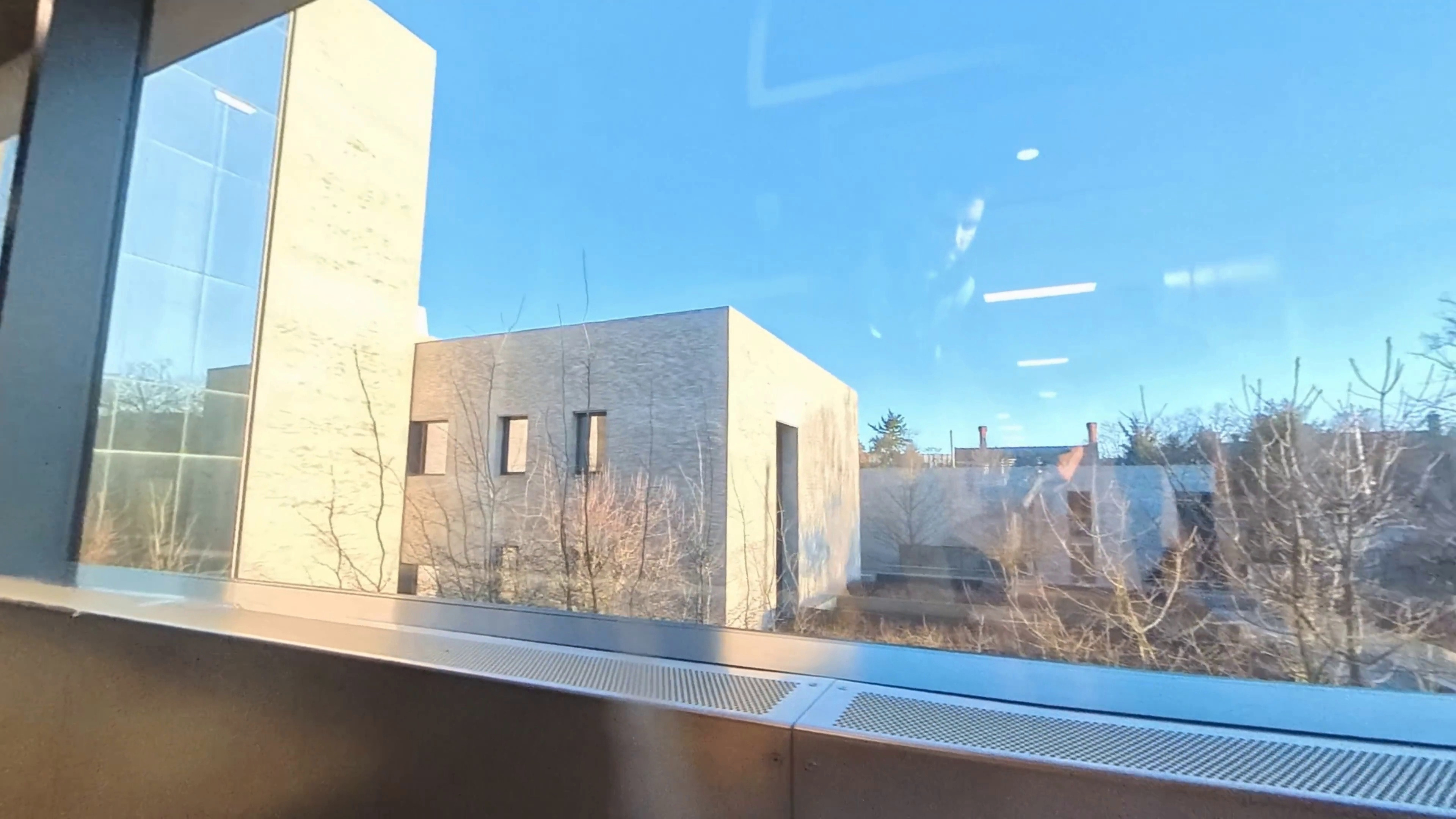} \\
\includegraphics[width=0.33\linewidth]{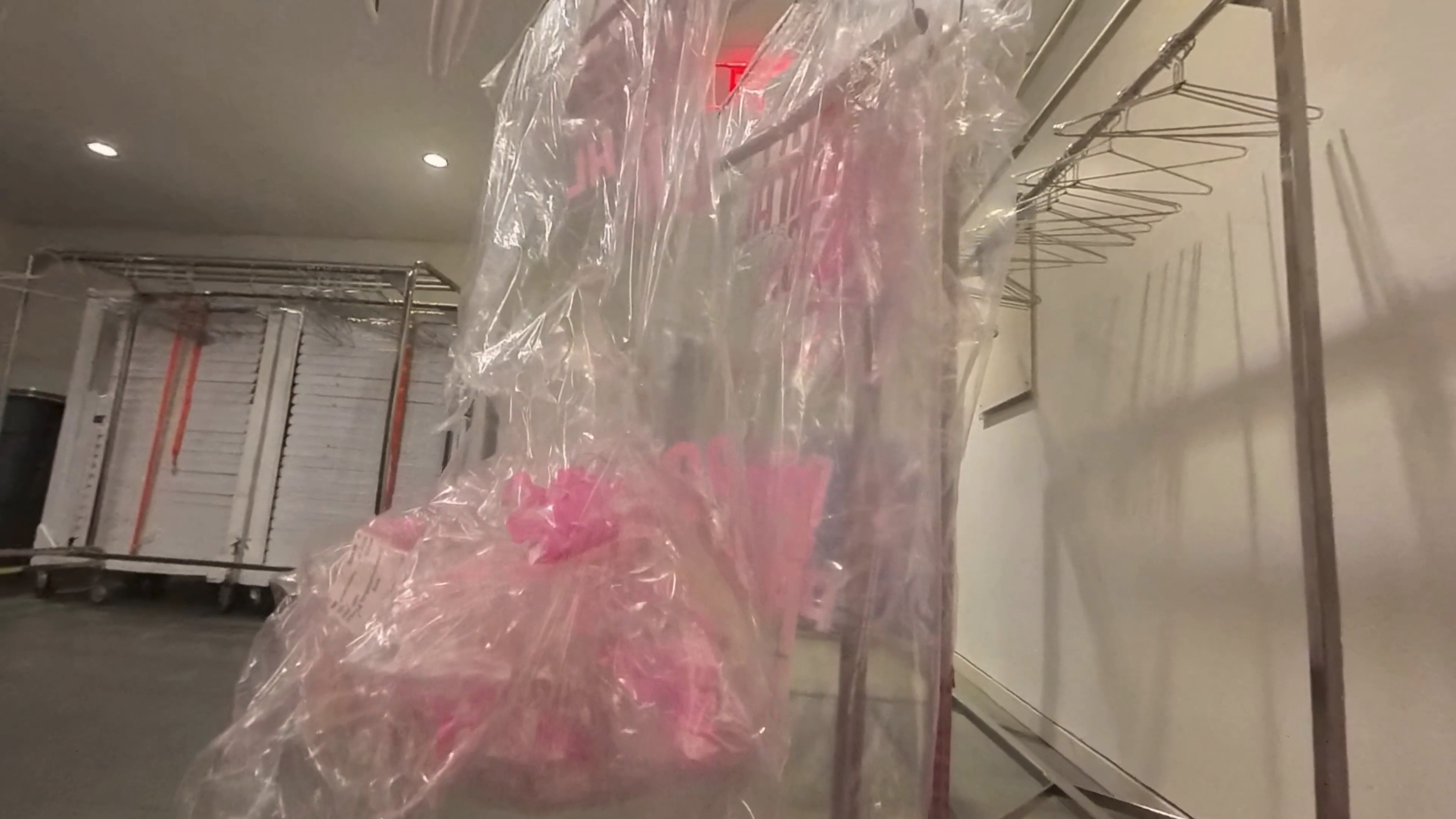} &
\includegraphics[width=0.33\linewidth]{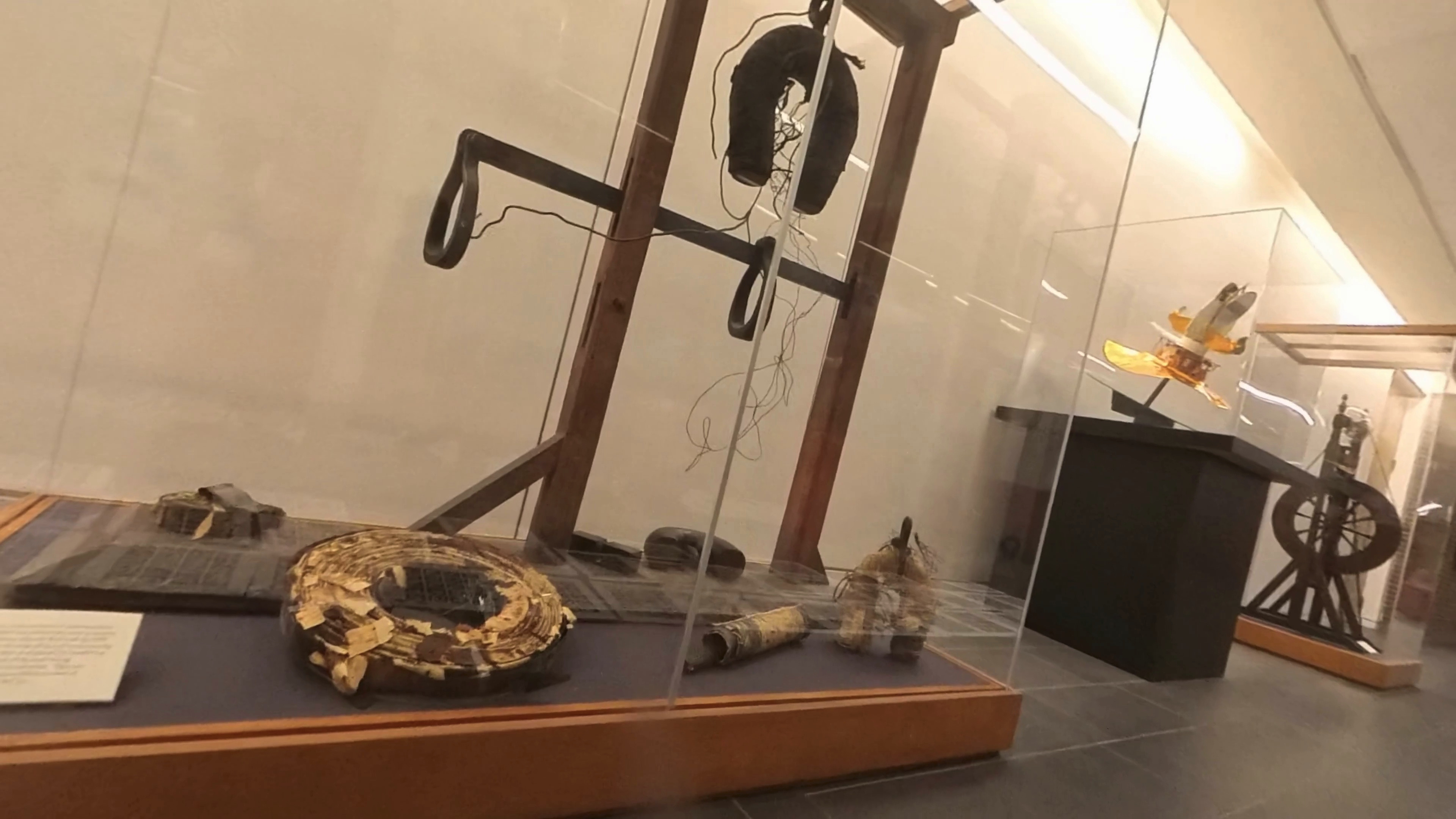} &
\includegraphics[width=0.33\linewidth]{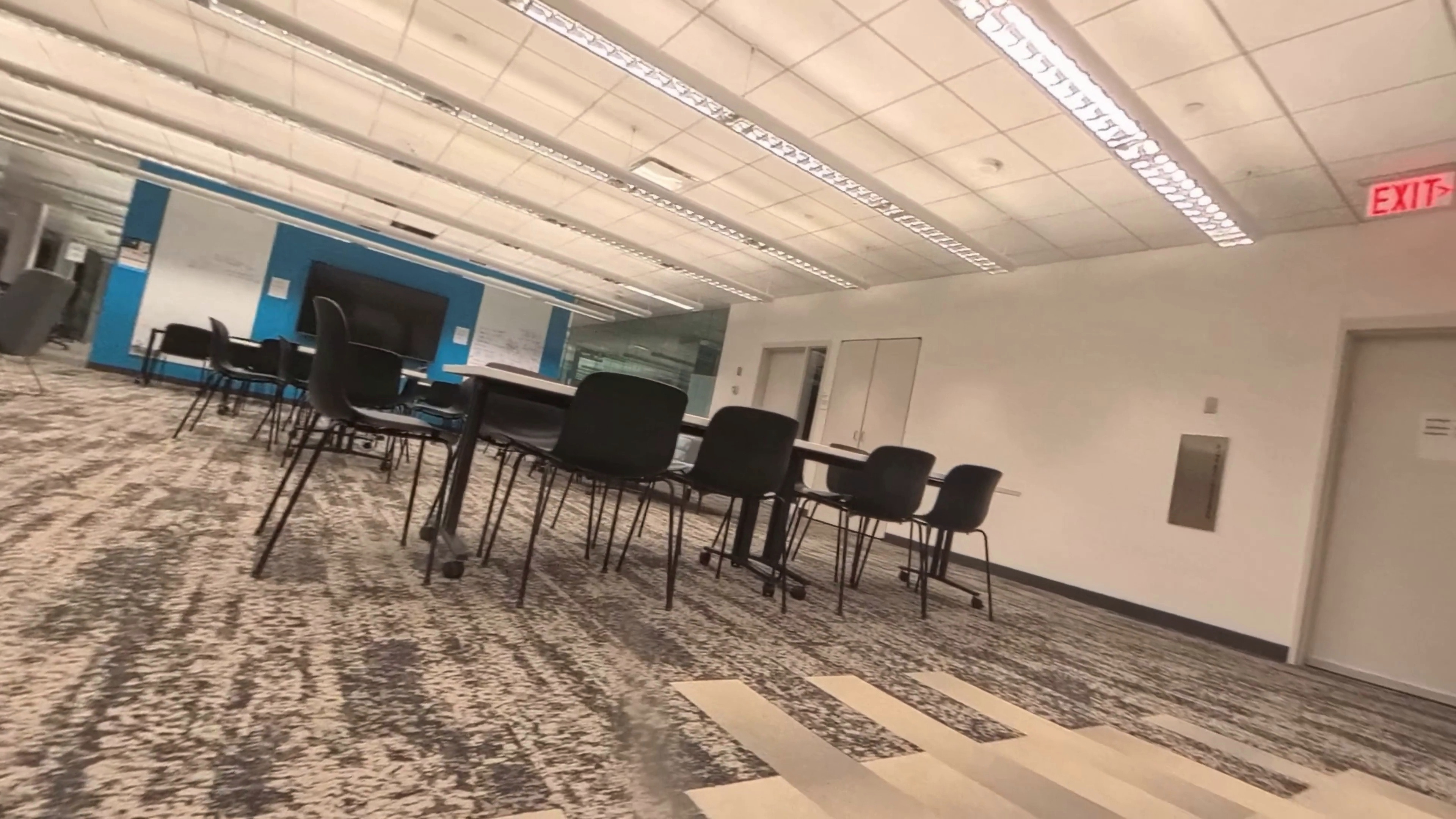} \\
\end{tabular}
\caption{6 scenes from our NVS benchmark. Our benchmark contains cases where the entire camera view is covered by non-Lambertian objects such as glass, iridescent sheet, and mylar.}
\label{fig:nvs_benchmark}
\end{figure}

Our dataset consists of 285 object scanning sequences, 40 long indoor sequences, and 40 long outdoor sequences. 

\xpar{Object scanning sequences} generally involve a particular object that is the main focus of the video. The sequences are filmed in different environments and with different camera trajectories in order to increase diversity. Object scanning videos include static scenes, lighting changes, non-Lambertian surfaces, dynamic objects, texture-less scenes. As shown in \cref{tab:dataset_stats}, an object scanning sequence has roughly 3K frames and is 49 seconds on average. 

\xpar{Indoor sequences} are longer videos filmed in indoor environments. We also include cases going from indoor to outdoor. These trajectories are challenging for current SLAM methods as they include frames completely covered with textureless surfaces, non-Lambertian objects, dynamic objects, and challenging camera motion. When filming indoor videos, we leave the ground-truth zone for the intermediate frames in order to cover longer distances. As shown in \cref{tab:dataset_stats}, an indoor sequence has roughly 13K frames and is 3 minutes 39 seconds on average.

\xpar{Outdoor sequences} are longer videos mostly filmed in outdoor environments. These sequences capture diverse lighting conditions, including sunny as well as nighttime shots, dynamic objects, and interesting camera motion. Similar to indoor videos, we leave the ground-truth zone for the intermediate frames in order to cover longer distances. As shown in \cref{tab:dataset_stats}, an outdoor sequence has roughly 18K frames and is 5 minutes 12 seconds on average. 

 An illustration of some randomly selected scenes from our benchmark can be seen in \cref{fig:example_frames}.

\section{Method}
\label{sec:method}

\subsection{Sensor Setup}

We used an Insta360 x4 - Ultimate 8K 360 Action Camera~\cite{insta360} for data capture. This camera can record ultra-high quality 360° videos at 60 FPS.

We used a ZED X stereo camera to acquire depth maps as well as to provide stereo RGB output. The ZED X is capable of capturing high resolution 1920$\times$1200 images at 60 FPS with global shutter. The ZED X camera's API also generates per-pixel depth estimates for every video frame taken, along with confidence measures for each pixel's depth. We use the depth values from the ZED camera to calculate our IOF error metric. The ZED X camera also includes an IMU that is synced and calibrated with respect to the stereo rig. We provide the IMU data to the user as well.  

\begin{figure}
    \centering
    \includegraphics[width=\linewidth]{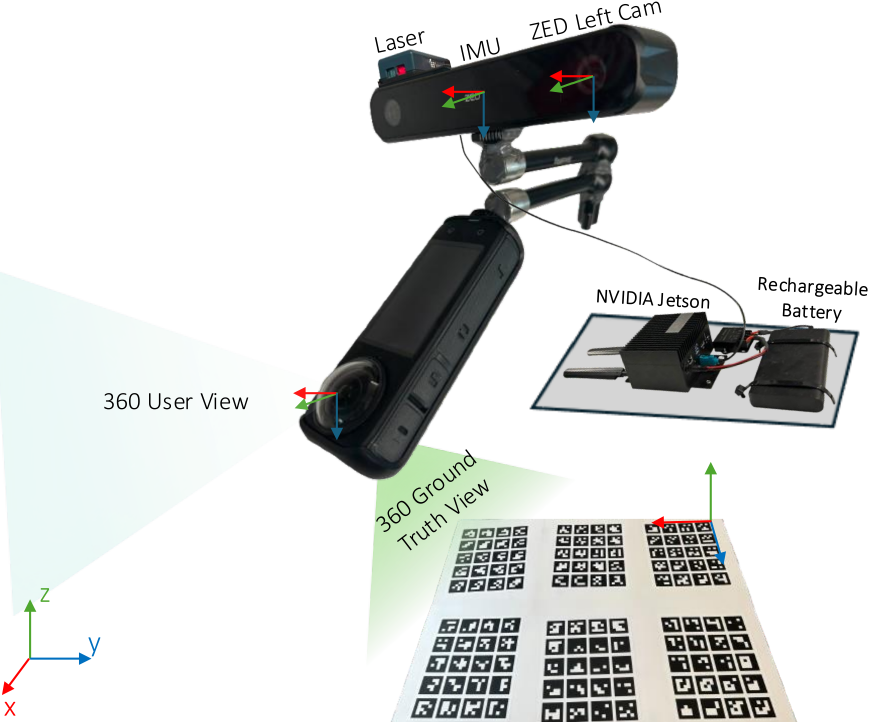}
    \caption{An illustration of our camera rig. We render a user-view and a ground-truth view from the Insta360 camera. The user-view is shown to the user, i.e. SLAM/NVS method. The GT view sees calibration boards on the ground and is used to estimate the ground-truth pose of the camera. The ZED X stereo camera is rigidly attached to Insta360. The ZED X is powered by an NVIDIA Jetson. We use a laser pointer to sync the stereo and 360 frames. We also record the IMU output, which is synchronized with the ZED X.}
    \label{fig:hardware}
\end{figure}

The ZED X camera requires hardware that can support high data throughput. We utilize the NVIDIA Jetson Orin NX along with a GMSL2 Fakra cable to serve this purpose. The NVIDIA Jetson requires a dedicated power source to read data from the ZED X camera. However, the camera rig also needs to be portable and functional in a variety of indoor and outdoor environments, making it infeasible to keep the NVIDIA
Jetson plugged into a power outlet at all times during recording. As a result, we used an external, rechargeable battery pack and soldered together a new power adapter to connect it to the NVIDIA Jetson. We then secured this setup, along with an external SSD, to a laser cut acrylic board to prevent components from becoming loose while recording. We placed the acrylic board into a  backpack with a small opening at the top for the GMSL2 Fakra cable to connect to the camera outside the backpack. See \cref{fig:hardware} for an illustration of our setup.

\subsection{Data Collection}
\label{sec:gt}

To capture data, we first lay out AprilTag calibration boards around the scene of interest. We then start recording with the Insta360 along with the stereo camera and IMU using the NVIDIA Jetson. Note that setting this up in a completely novel scene takes about 5 minutes.

In order to time-sync the 360 camera and the stereo camera, we flash a laser in view of both at the start and end of the recording. We then manually align the frames using this laser flash, giving us time-aligned streams for both. Note that the IMU stream is provided by the ZED X and is already synchronized with the RGB-D stereo stream.

For each sequence, we record at least two videos with the 360-camera. The first video is the actual trajectory for which we provide the camera pose ground truth. The second video is a close-up of the calibration boards, which provides additional constraints to our optimization method. This is then used to construct a pose graph that shows the relative and global poses of the calibration boards.

We postprocess the .insv videos that we obtain from the 360 camera by rendering two views. We choose the rotation for the user view such that it sees the scene of interest and does not see the calibration boards. We choose the rotation for the ground truth view such that it clearly sees that calibration boards and can be used to obtain camera poses. We calibrate each rendered view separately, meaning that we map a set of rotation parameters from the 360 to both intrinsics and extrinsics relative to a default view. 

\begin{figure*}[ht]
  \centering
    \includegraphics[width=\linewidth]{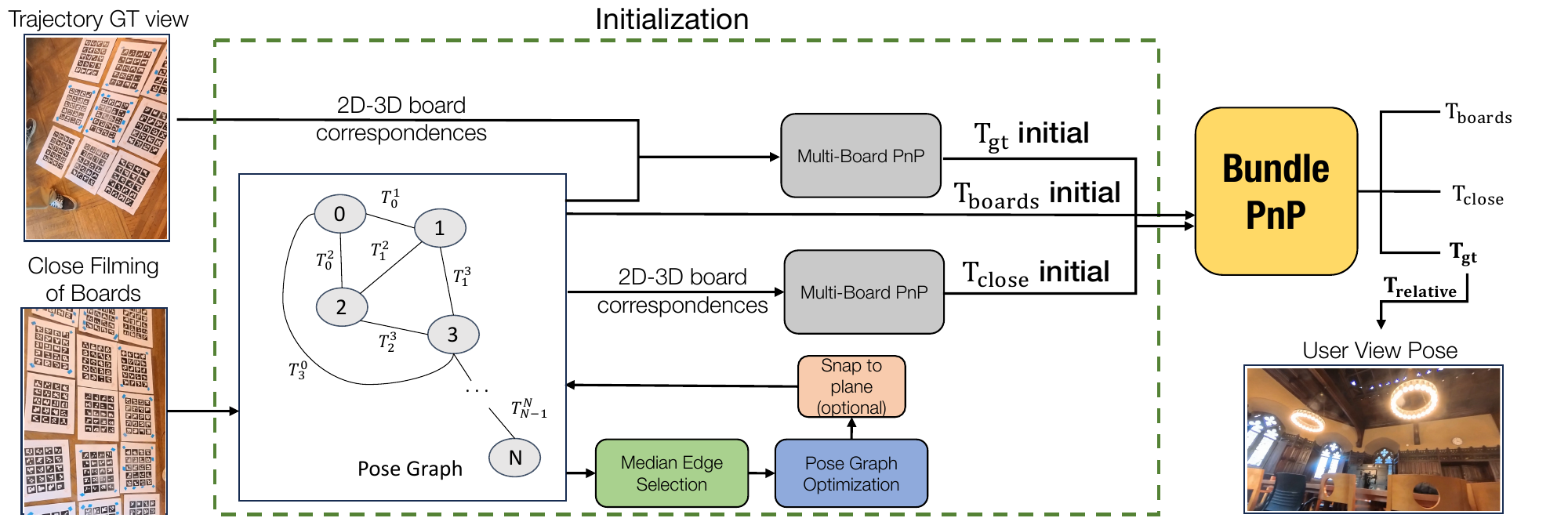} 
    \label{fig:short-a}
  \hfill
  \caption{Our ground-truth method. We place calibration boards in a new environment and film using our 360-degree camera and stereo camera setup. The Insta360 recording is rendered as both a user-view, which does not see the boards and is what the methods are evaluated on, along with a ground-truth view from which we obtain the camera poses. For every frame, we estimate the pose between the Insta360 and the boards using PnP. This relative pose is chained to the pose graph between the boards to get the global pose of the camera. Constructing the pose graph consists of filming the boards separately. We estimate the relative poses between the boards by using the camera as an anchor and utilize median-based outlier rejection, pose graph optimization, and the optional assumption that the boards are coplanar to get the ground-truth pose graph (snapping). After these initialization stages, we perform Bundle PnP to get the final poses.}
  \label{fig:gt_framework}
\end{figure*}

\subsection{Ground-Truth Pipeline}

An overview of our ground-truth method can be seen in \cref{fig:gt_framework}. At a high-level, we solve an optimization objective that takes in 2D-3D correspondences from the ground-truth view of the trajectory and the close filming of the boards, and outputs the camera poses as well as the poses of the boards. We propose several techniques to initialize high-quality poses for both boards and the camera, and propose an optimization method that outputs the camera trajectory. 

\subsubsection{Constructing the Board Pose Graph}
\label{sec:pose_graph}

 Each calibration board has distinct marker patterns that serve as unique identifiers, recognizable through marker detection techniques. We first locate the 2D pixel coordinates of each corner. We also know the 3D coordinates of each corner in local board coordinates. We provide the corresponding 3D and 2D coordinates as input to Perspective-n-Point and obtain the relative pose between the camera frame $c_t$ and the calibration board $B_i$, denoted $\mathbf{T}_{B_i}^{c_t}$.

This procedure results in multiple board poses relative to the camera per frame. For any detected pair of boards $(i,j)$, we compute their relative pose as

\begin{equation}
    \mathbf{T}_{B_i}^{B_j} = \mathbf{T}_{c_t}^{B_j}\mathbf{T}_{B_i}^{c_t}.
\end{equation}

We compute relative poses between each visible pair and thus construct a pose graph. In this graph, nodes represent the global poses of individual boards relative to an initial reference board denoted $o$, while edges represent the relative transformation between pairs. The global pose of board $i$ can be computed by chaining together the relative transformations up to the reference board.
\begin{equation}
    \mathbf{T}_o^{B_i} = \mathbf{T}_o^{B_1} \mathbf{T}_{B_1}^{B_2} \cdots \mathbf{T}_{B_{n-1}}^{B_i}.
    \label{eq:chain}
\end{equation}

\xpar{Median Edge Selection} Since the same pair of boards are visible across multiple frames, we have many estimated relative poses for that specific pair. For each such pose, we compute the distance between the boards. We take the pose that corresponds to the median distance to rule out outliers. 

\xpar{Pose Graph Optimization} We then perform Pose Graph Optimization using g2o \cite{g2o}, which optimizes the global poses to be consistent with the measured relative poses.

\xpar{Snap to Plane} In the scenes where all calibration boards are on the same plane, we apply a snapping technique that aligns the z-axis of each board with the z-axis of the reference board. We adjust the x and y axes to preserve the orthogonality of the coordinate system and update the relative poses based on the new axis. 

See the ablation study in Section \ref{sec:Ablation_Study} for how much these techniques improve the accuracy.

\subsubsection{Initializing Camera Poses}
\label{sec:init_pose}
We estimate the camera pose of each frame by combining information from several visible boards. First, we detect markers in each visible board. We then transform each 3D marker into coordinates of a common reference board $i$ using the pose graph. Finally, we perform PnP with the 2D-3D correspondences to find the relative pose between the camera and the reference board $i$. We can then easily obtain the global pose of the camera by chaining the global pose of the reference board obtained from the pose graph and the relative pose we computed using PnP: 

\begin{equation}
    \mathbf{T}_o^{c} = \mathbf{T}_o^{i} \mathbf{T}_i^{c}
\end{equation}

\subsubsection{Bundle PnP}

We obtain the final camera poses in ground-truth view by using an optimization technique that we name \emph{Bundle PnP}. The main idea is to minimize the reprojection error of the detected markers by solving for the global poses of the calibration boards and the camera. 

Bundle PnP takes the 2D-3D correspondences of the markers as input and initializes the camera poses and the board poses as described in \cref{sec:init_pose,sec:pose_graph}. It takes the 3D points in local board coordinates, transforms them to camera coordinates, and projects them using the intrinsics.

Let $\mathbf{T}_{B_i}^{o}$ be the board $i$ to world transformation. Let $\mathbf{T}_{c_j}^{o}$ be the frame $j$ to world transformation. Let $\boldsymbol{\theta}$ denote the intrinsics including the radial and tangential distortion parameters calibrated beforehand. Let $p_{B_i}^{(d)}$ be the $d$th 3D point in local board $B_i$ coordinates. Let $u_{c_t}^{(d)}$ be the $d$th detected 2D point in frame $c_t$ in pixel coordinates. We minimize the reprojection error and solve the following optimization problem
\begin{equation}
        \minimize_{\mathbf{T}_{B_i}, \mathbf{T}_{c_t}} \quad \sum_{t = 1}^T \sum_d \norm{\pi(\mathbf{T}^{c_t}_{o} \mathbf{T}^o_{B_i} p_{B_i}^{(d)}, \boldsymbol{\theta}) - u_{c_t}^{(d)}}_2^2.
\end{equation}
We formulate Bundle PnP as a nonlinear least squares problem in the Ceres Solver \cite{Agarwal_Ceres_Solver_2022} and solve it using the Levenberg Marquardt algorithm \cite{levenberg_method_1944}.

\begin{table*}[ht]
\centering
\resizebox{\linewidth}{!}{
    \begin{tabular}{lcccccc}
    \toprule
    Scene & Dark Room & Office & Exhibit & Windows-Night & Windows-Day & Storage \\ 
     Metric & PSNR$\uparrow$ / LPIPS$\downarrow$ & PSNR$\uparrow$ / LPIPS$\downarrow$ & PSNR$\uparrow$ / LPIPS$\downarrow$ & PSNR$\uparrow$ / LPIPS$\downarrow$ & PSNR$\uparrow$ / LPIPS$\downarrow$ & PSNR$\uparrow$ / LPIPS$\downarrow$ \\
  
    \midrule

    Nerfacto, Our Pose & 24.46/0.277 & 26.14/0.220 & 24.74/0.176 & \textbf{26.88/0.186} & 24.26/0.233 & 24.68/0.228 \\
    Nerfacto, COLMAP Pose & Failed & Failed & Failed & 20.30/0.491 & Failed & Failed \\
    \midrule
    Splatfacto, Our Pose & 29.70/0.204 & 26.93/0.151 & 27.96/0.108 & \textbf{32.26/0.129} & 26.41/0.197 & 27.23/0.151 \\
    Splatfacto, COLMAP Pose & Failed & Failed & Failed & 23.28/0.379 & Failed & Failed \\
    \bottomrule
    \end{tabular}
}
\caption{NVS results on 6 scenes from our dataset. Numbers are PSNR$\uparrow$/LPIPS$\downarrow$. While COLMAP fails on 5 out of the 6 scenes, our method always succeeds. When COLMAP does not fail, our pose gives better rendering quality for both Nerfacto and Splatfacto.}
\label{tab:results_nerf}
\end{table*}

\subsection{Induced Optical Flow Metric}

We propose a new evaluation metric that calculates the expected optical flow induced by the difference between the estimated pose and the ground-truth pose. For instance, a ten centimeter error made when filming an object close-up will induce much higher optical flow than a ten centimeter error made when filming a distant landscape. Intuitively, this can also be thought of as how misaligned objects would seem on an augmented reality screen if they were placed based on a camera pose estimated by a SLAM method. Our new metric allows us to compare methods across trajectories that have different scene scales. 

At a high-level, the induced optical flow metric can be thought of as the expectation

\begin{equation}
\label{eq:IOF_expectation}
    \mathrm{IOF} = \mathop{\mathbb{E}}_{t, d, u, v}\norm{\mathbf{flow}(t,d,u,v)}_2
\end{equation}
where $t \sim \mathcal{U}(T)$ is a uniform distribution over frames, $d \sim p$ is the depth distribution of the sequence, and $ (u,v) \sim \mathcal{U}(A)$ is a uniform distribution over pixels. Here, $\mathbf{flow}(t,d,u,v)$ is the optical flow from the $(u,v)$ pixel of the ground-truth frame at time $t$ to the estimated frame.  

 The induced optical flow $\mathbf{flow}(t,d,u,v)$ re-projects pixels from the ground-truth camera pose to the estimated camera pose. Thus, we need to know both the ground-truth pose and a depth distribution. We obtain depth samples using the ZED X stereo camera. We then fit mixture of Gaussian and mixture of Gamma distributions with 1-8 components. We take the best one with respect to the Bayesian Information Criterion \cite{bic}. We denote this depth distribution as $p(\text{d})$. See \cref{sec:supp_depth_dist} for details. 

 Having a parametric depth distribution allows us to write the expectation in \cref{eq:IOF_expectation} as 

 \begin{equation}
 \label{eq:integral_IOF}
     \mathrm{IOF} = \frac{1}{AT}\sum_t \sum_{u,v} \int_{0}^{\infty} \norm{\mathbf{flow}(t,d,u,v)}_2 \, p(d) \,dd.
 \end{equation}

Thus, instead of sampling $d \sim p$, we numerically integrate the \cref{eq:integral_IOF} to get a deterministic answer. See \cref{sec:computing_IOF} for details on how we compute the IOF metric. 

\xpar{Scaling the range of IOF} The proposed IOF metric can range from 0 to infinity based on how off the estimated pose is. To ensure better interpretability, we convert IOF to an accuracy variant called \emph{Flow AUC}. We calculate this metric by computing the area under the curve (AUC) of the percentage of pixels with flow under thresholds from 0 to 100 px. This new metric ranges from $0\%$ to $100\%$. $100\%$ implies perfect pose and $0\%$ implies none of the pixels have induced optical flow under 100 pixels.

\xpar{Trajectory Alignment} 
In monocular SLAM, the scale is ambiguous. Thus, we first perform Umeyama alignment \cite{umeyama} in $\mathbf{Sim}(3)$ space to align the estimated trajectory to the ground-truth trajectory.

The rotation error has a large effect on IOF. Thus we perform an additional Kabsch alignment \cite{kabsch_solution_1976} in $\mathbf{SO}(3)$ to find a global rotation that minimizes the error between the ground truth orientations and the estimated trajectory orientations. We empirically find this to be useful on top of the initial $\mathbf{Sim}(3)$ alignment. See \cref{sec:supp_traj_alignment} for more details.

\begin{figure}[t]
  \begin{center}
  \resizebox{\linewidth}{!}{
    \includegraphics[width=\textwidth]{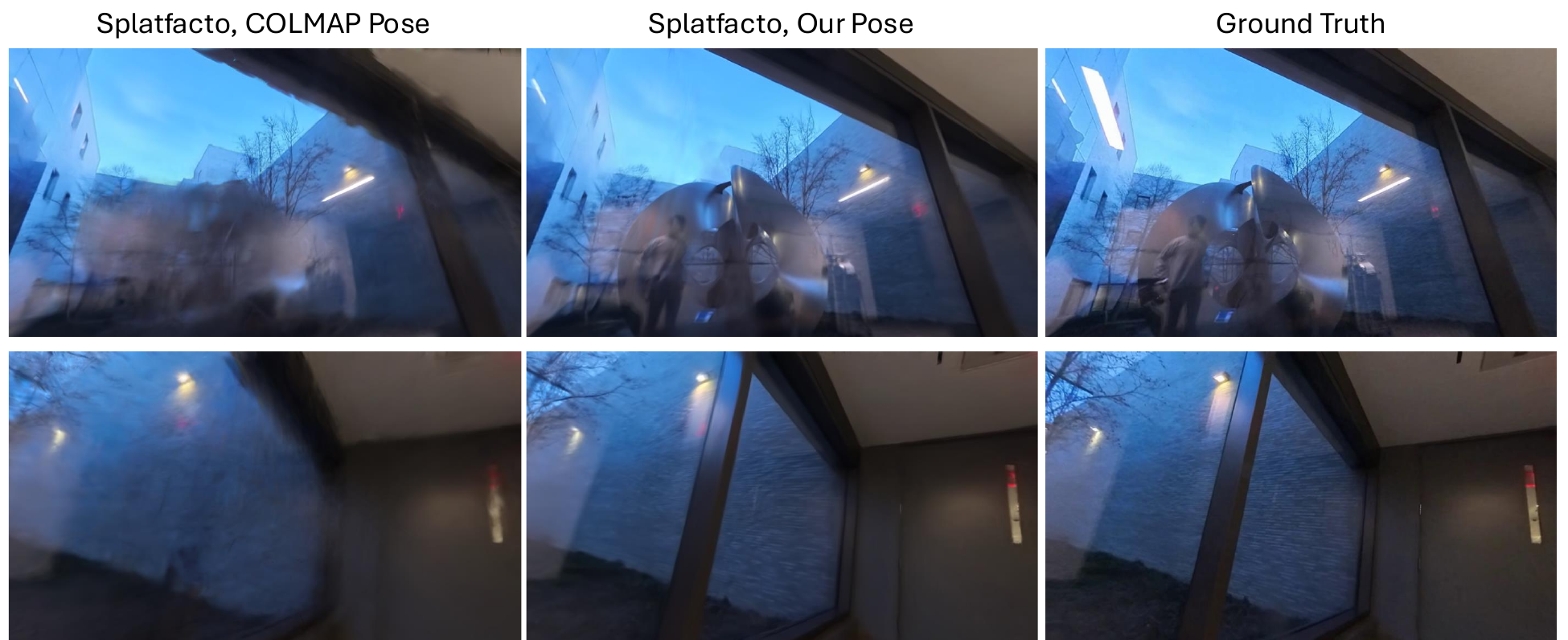}}
  \end{center}
  \vspace{-1.5em}
  \caption{On the Window-Night scene, where COLMAP successfully outputs poses, the Splatfacto~\cite{nerfstudio} model trained with our pose produces higher quality renders on validation views.}
  \label{fig:results_nerf}
  \vspace{-1.5em}
\end{figure}

\subsection{Relative Pose Calibration}
\label{sec:calibration}
Bundle-PnP returns the camera pose of the ground-truth view. However, SLAM/NVS methods only see the user-view. Thus, we need to find the relative pose between the user-view and the ground-truth view in order to evaluate these methods. Note that the relative pose is not a pure rotation due to the unknown offset between the two fisheye lenses of Insta360. Furthermore, off-the-shelf stereo calibration methods or PnP do not work in our case because the views do not overlap. 

Thus, we propose a variant of Bundle PnP called \emph{Bundle Rig PnP} that estimates the relative pose between multiple cameras that do not necessarily overlap. The idea is to assume that each camera is fixed with respect to a rig and optimize over the relative poses between the cameras and the rig jointly with the poses of the boards and the rig. Thus, the output of this optimization includes the poses between each camera and the rig. 

Formally, Bundle Rig PnP minimizes the sum of reprojection errors across all cameras and time as 
\begin{equation}
\begin{aligned}
\minimize_{\mathbf{T}^o_{B_i}, \mathbf{T}^o_{rig, t}, \mathbf{T}_{rig}^c}
\quad
\sum_{t = 1}^T \sum_c \sum_d
\norm{\mathbf{r}_{c,t,d}}^2 ,
\end{aligned}
\end{equation}

\begin{equation}
\mathbf{r}_{c,t,d}
=
\pi\!\left(
\mathbf{T}^{c}_{\mathrm{rig}}
\mathbf{T}^{\mathrm{rig}}_{o,t}
\mathbf{T}^o_{B_i}
p_{B_i}^{(d)},
\boldsymbol{\theta}_c
\right)
-
u_{c,t}^{(d)} .
\end{equation}
where $\mathbf{T}^o_{rig, t}$ is the pose of the rig at time $t$, and $\mathbf{T}_{rig}^c$ is the relative pose between the rig and camera $c$, which does not change with respect to time. 

This is a general formulation that can handle multiple cameras. In our case, we just need to estimate the pose between two different views. Therefore, we directly optimize the relative pose between the ground-truth view and the user-view, which is equivalent to taking the rig to be the ground-truth view and having only one camera, which is the user-view. See the \cref{sec:bundle_rig_twocam} for this particular optimization objective we use. 

Note that we also calibrate the relative pose between the $360^\circ$ camera and the stereo camera using Bundle Rig PnP. 

\begin{table}[t]
\centering
\resizebox{\columnwidth}{!}{
    \begin{tabular}{ccc}
        \toprule
        \textbf{Sequence} & \textbf{COLMAP - ATE (mm)} & \textbf{Ours - ATE (mm)} \\
        \midrule
        0 & 6.3  & \textbf{3.1} \\
        1 & 8.5  & \textbf{4.2} \\
        2 & 8.0  & \textbf{2.1} \\
        3 & 9.2  & \textbf{2.3} \\
        4 & 10.5 & \textbf{2.3} \\
        5 & 13.9 & \textbf{2.4} \\
        6 & 5.4  & \textbf{2.0} \\
        7 & 26.1 & \textbf{3.8} \\
        8 & 14.7 & \textbf{3.7} \\
        \midrule
        \textbf{Avg ATE} & 11.4 & \textbf{2.88} \\
        \bottomrule
       
    \end{tabular}
}
\caption{Comparison of mean Absolute Trajectory Error (ATE) between COLMAP and our method. Our ground-truth method is more accurate than COLMAP in all 9 sequences.}
\label{tab:ate_comparison_colmap}
\end{table}

\begin{table*}[ht]
\centering
\resizebox{\textwidth}{!}{%
\begin{tabular}{l l *{9}{c}}
\toprule
\multirow{2}{*}{Method} & \multirow{2}{*}{Metric} &
\multicolumn{3}{c}{Scanning Videos} &
\multicolumn{3}{c}{Indoors Videos} &
\multicolumn{3}{c}{Outdoors Videos} \\
\cmidrule(lr){3-5}\cmidrule(lr){6-8}\cmidrule(lr){9-11}
& & Easy & Medium & Hard & Easy & Medium & Hard & Easy & Medium & Hard \\
\midrule
\multirow{5}{*}{DPVO~\cite{teed_deep_2023}} 
    & ATE (m)\,$\downarrow$           & 0.01 & 0.03 & 0.26 & 0.49 & 1.01 & 0.44 & 0.66 & 0.89 & 0.54 \\
    & Rotation\,$^\circ\!\downarrow$  & 0.38 & 0.99 & 10.36 & 31.94 & 47.10 & 48.61 & 16.22 & 48.51 & 58.78 \\
    & Flow AUC\,$\uparrow$            & 84.97 & 61.96 & 19.42 & 18.11 & 0.18 & 0.01 & 14.03 & 0.30 & 0.01 \\
    & Coverage (\%)\,$\uparrow$       & 100.00 & 100.00 & 100.00 & 100.00 & 99.99 & 99.99 & 100.00 & 99.99 & 100.00 \\
    & Composite\,$\uparrow$           & 91.87 & 76.51 & 32.53 & \textbf{30.67} & 0.36 & 0.01 & 24.61 & 0.60 & 0.03 \\
\midrule
\multirow{5}{*}{DPV-SLAM \cite{lipson_deep_2024}}
    & ATE (m)\,$\downarrow$           & 0.01 & 0.04 & 0.24 & 0.54 & 1.05 & 0.45 & 0.66 & 0.89 & 0.56 \\
    & Rotation\,$^\circ\!\downarrow$  & 0.38 & 0.97 & 9.72 & 21.91 & 42.84 & 62.15 & 15.70 & 32.97 & 40.13 \\
    & Flow AUC\,$\uparrow$            & 85.44 & 63.35 & 26.76 & 17.87 & 0.46 & 0.38 & 17.32 & 0.88 & 0.89 \\
    & Coverage (\%)\,$\uparrow$       & 100.00 & 100.00 & 98.73 & 100.00 & 99.99 & 99.99 & 100.00 & 99.99 & 100.00 \\
    & Composite\,$\uparrow$           & \textbf{92.15} & \textbf{77.56} & \textbf{42.11} & 30.32 & \textbf{0.92} & \textbf{0.75} & \textbf{29.53} & 1.75 & 1.77 \\
\midrule
\multirow{5}{*}{LEAP-VO\cite{chen2024leap}}
    & ATE (m)\,$\downarrow$           & 0.33 & 0.51 & 0.54 & 0.70 & 0.95 & 0.43 & 0.68 & 0.86 & 0.59 \\
    & Rotation\,$^\circ\!\downarrow$  & 4.27 & 11.02 & 28.07 & 43.10 & 58.33 & 67.48 & 45.63 & 56.84 & 62.77 \\
    & Flow AUC\,$\uparrow$            & 29.78 & 16.12 & 3.23 & 5.89 & 0.12 & 0.17 & 1.29 & 0.74 & 0.23 \\
    & Coverage (\%)\,$\uparrow$       & 100.00 & 79.54 & 92.42 & 10.00 & 9.97 & 10.01 & 10.00 & 10.04 & 10.00 \\
    & Composite\,$\uparrow$           & 45.89 & 26.81 & 6.24 & 7.41 & 0.24 & 0.34 & 2.28 & 1.38 & 0.46 \\
\midrule
\multirow{5}{*}{ORB-SLAM3~\cite{campos2021orb}}
    & ATE (m)\,$\downarrow$           & 0.04 & 0.15 & 0.23 & 1.04 & 0.57 & 1.58 & 0.32 & 0.28 & 0.44 \\
    & Rotation\,$^\circ\!\downarrow$  & 1.07 & 14.34 & 26.94 & 16.49 & 40.86 & 46.20 & 38.73 & 56.33 & 19.45 \\
    & Flow AUC\,$\uparrow$            & 78.47 & 52.11 & 18.41 & 13.16 & 0.03 & 0.26 & 36.59 & 23.25 & 15.61 \\
    & Coverage (\%)\,$\uparrow$       & 51.88 & 33.59 & 10.80 & 8.19 & 17.79 & 6.55 & 23.54 & 16.83 & 25.89 \\
    & Composite\,$\uparrow$           & 62.46 & 40.85 & 13.61 & 10.10 & 0.06 & 0.49 & 28.65 & \textbf{19.52} & \textbf{19.48} \\
\midrule
\multirow{5}{*}{COLMAP~\cite{Schonberger_2016_CVPR}}
    & ATE (m)\,$\downarrow$           & 0.09 & 0.13 & 0.19 & ― & ― & ― & ― & ― & ― \\
    & Rotation\,$^\circ\!\downarrow$  & 1.33 & 2.33 & 8.94 & ― & ― & ― & ― & ― & ― \\
    & Flow AUC\,$\uparrow$            & 77.96 & 46.56 & 21.63 & ― & ― & ― & ― & ― & ― \\
    & Coverage (\%)\,$\uparrow$       & 61.79 & 60.43 & 44.38 & ― & ― & ― & ― & ― & ― \\
    & Composite\,$\uparrow$           & 68.94 & 52.60 & 29.08 & ― & ― & ― & ― & ― & ― \\
\bottomrule
\end{tabular}}
\caption{Evaluation of five SLAM/VO methods on \projectname. $\downarrow$ means lower is better; $\uparrow$ means higher is better. Coverage roughly means for what percent of the frames the method outputs camera pose. Composite score combines Flow AUC and Coverage.}
\label{tab:evaluation}
\end{table*}

\begin{table}[ht]
\centering
\resizebox{\columnwidth}{!}{
    \begin{tabular}{cccccccccc}
    \toprule
    \# Boards & Filming  & Combination & Pose Graph & Snapping & \multicolumn{3}{c|}{Cam. Init.} & Bundle & Total ATE \\ 
    \cmidrule(lr){6-8}
     & Distance& &Optimization & & 0 & 1 & 2 & PnP& (mm)\\ 
    \midrule
    8  & M & C-M &  & & \checkmark &  &  && 10 \\ 
    8  & M & C-M &  & \checkmark & \checkmark && & & 8.1  \\ 
    8  & M & C-M && \checkmark &  & \checkmark &  && 6.8  \\ 
    8  & M & C-M & &\checkmark &  &  & \checkmark && 5.5 \\ 
    16  & C & C-C & & &  &  & \checkmark & & 22  \\ 
    16  & C & C-C & \checkmark & &  &  & \checkmark && 7.5  \\ 
    16  & C & C-C &  & \checkmark &&  & \checkmark & & 7.1  \\ 
    16  & C & C-C & \checkmark & \checkmark &&  & \checkmark && 3.0  \\ 
    16  & M & M-M & \checkmark & \checkmark &&  & \checkmark & &3.2  \\ 
    16  & M & C-M & \checkmark & \checkmark &&  & \checkmark & &2.6  \\ 
    10  & M & C-M & \checkmark & \checkmark &&  & \checkmark & & 4.4  \\ 
    10  & M & C-M & \checkmark & \checkmark &&  & \checkmark & \checkmark &\textbf{2.1}  \\ 
    \bottomrule
    \end{tabular}
}
\caption{Ablation study on our ground-truth method. Pose graph optimization, snapping, and multi-board PnP all reduce the final error significantly. Furthermore, Bundle PnP more than halves the final error.}
\label{tab:ablation}
\end{table}

\section{Experiments}
\label{sec:experiment}

\subsection{Ground-Truth Validation with MoCap}

We externally validate the accuracy of our ground-truth by comparing it to the output of a Motion Capture (MoCap) system by placing reflective markers on our camera rig. In particular, we use Vicon Vantage V16, which has a 0.201 mm RMSE accuracy \cite{vicon}. 

External validation should ideally test how reliable the ground-truth is in practice and as used in the benchmark. Therefore, we emulate 9 randomly chosen sequences from \projectname. We use the same board setup and try to capture the same trajectory, but in the MoCap room.

We then extract the trajectory from the 360-video by taking into account the relative pose between the markers and the Insta360. We compare this with the one provided by the MoCap system. As shown in \cref{fig:millimeter_all} and \cref{tab:ate_comparison_colmap}, our method achieves an accuracy comparable to MoCap, where the ATE between the trajectory is in millimeters. 

\cref{tab:ate_comparison_colmap} shows that our ground-truth method outperfoms COLMAP on all 9 MoCap sequences. In certain sequences, COLMAP has cm error whereas our method mm error. In fact, the average ATE of our method is 2.88 \textbf{mm}, and the average ATE of COLMAP is 1.14 \textbf{cm} across all sequences.

\subsection{SLAM Benchmark}

\cref{tab:evaluation} shows the evaluation of state-of-the-art methods DPVO \cite{teed_deep_2023}, DPV-SLAM \cite{lipson_deep_2024}, ORB-SLAM3 \cite{campos2021orb}, LEAP-VO \cite{chen2024leap}, and COLMAP \cite{colmap} on \projectname. We categorize each video into easy, medium, and hard. We observe that even scanning videos where the camera movement is simple include quite challenging cases for the current methods with the highest Flow AUC for hard scanning videos being 26.76 out of 100. The methods struggle the most with long indoor and outdoor videos. We combine the Flow AUC and the percentage of frames tracked by the SLAM method into a composite score (see \cref{sec:supp_cov_comp_score} for details).

\subsection{Novel View Synthesis}
\label{sec:NVS}

We train the Nerfacto and Splatfacto models on 6 scenes selected from our benchmark shown in \cref{fig:nvs_benchmark}. Nerfacto and Splatfacto are variants of NeRF~\cite{nerf} and 3DGS~\cite{3DGS} implemented in the NeRFstudio~\cite{nerfstudio} framework. We use the default setting and train both methods for 30K steps, using either the poses reconstructed from COLMAP or provided by our pipeline.

The results are shown in \cref{tab:results_nerf}. Since our dataset is challenging and contains many non-Lambertian objects (\eg, windows and plastics), COLMAP fails on 5 out of the 6 scenes. For the only scene COLMAP succeeded (Window-Night), novel view synthesis models trained with our poses achieve much better PSNR and LPIPS on test views, proving that our poses cause less misalignment and are better for novel view synthesis benchmarks. This can also be seen from the visualizations in \cref{fig:results_nerf}.

\subsection{Ablation Study}
\label{sec:Ablation_Study}

\cref{tab:ablation} shows an ablation study on our design choices. We use three different setups of boards and compare our 6D trajectory with the one obtained from MoCap. The three setups use 8, 10, 16 boards respectively.

We test multiple techniques to obtain the ground-truth pose from the GT-view. For all of these, we assume that the pose graph is already constructed, and the question is how to use multiple boards seen in a single frame. See supplemental for more details.
\begin{itemize}
    \item Method 0: For each board in the frame, we compute the relative pose between the camera and that board using PnP. We return the pose of the camera as the pose computed from the closest board. 
    \item Method 1: We apply a confidence-weighted fusion approach. Since the camera sees several boards at once, we assign a confidence score to each detected board pose based on reprojection error and the number of markers detected. 
     Lower reprojection error signals a more accurate pose. At frame $t$, we weight each board's pose by its re-projection error.
    
    \item  Method 2: We use the idea of multiboard-PnP described in \ref{sec:gt}. We estimate each frame's camera pose by combining all the visible checkerboards, transforming their inner 3D points into a common reference using the pose graph, and then performing a PnP estimation with multiple points.
\end{itemize}

From Table~\ref{tab:ablation}, we can see that the multi-board PnP approach has the best results compared to other PnP approaches. Pose Graph Optimization further improves the results. 
We also conducted two filming setups on the table: Close (C) filming, at an average distance of 50 cm between the camera and boards, and Medium (M) filming, at an average distance of 1.3 meters. Table \ref{tab:ablation} shows that by combining pose graphs that are filmed closely with a video at medium distance, we have millimeter accuracy. This approach forms the basis of our data collection strategy, where we capture one close-up video of the boards for a better pose graph and a trajectory video at a flexible distance for diversity.

We also see that with all settings enabled, Bundle PnP halves the ATE going from 4.4 mm ATE to 2.1 mm ATE. 

See \cref{supp:ablation} for a complete ablation study.

\section{Conclusion}
We introduce \projectname, a large-scale diverse dataset with accurate camera pose. Our novel ground-truth collection method allows us to scale up data collection with annotated camera pose while preserving accuracy and full 6DoF movement. We also hope to provide insights to SLAM researchers with our novel scale-aware Induced Optical Flow metric as well as to provide a new challenging NVS benchmark. 

\section{Acknowledgments}
This work was partially supported by the National Science Foundation, Onassis Foundation and a gift from Meta Reality Labs. 

{
    \small
    \bibliographystyle{ieeenat_fullname}
    \bibliography{main}
}

\clearpage
\setcounter{section}{0}
\renewcommand{\thesection}{\Alph{section}}

\phantomsection
\addcontentsline{toc}{section}{Appendix}
\section*{Appendix}
\label{sec:appendix}

\appendix

\input{supp}

\end{document}

%% file: preamble.tex
\usepackage{float}
\newcommand{\projectname}{Princeton365\xspace}

%% file: supp.tex
\begin{figure}[ht]
  \begin{center}
  \resizebox{\linewidth}{!}{
    \includegraphics[width=\textwidth]{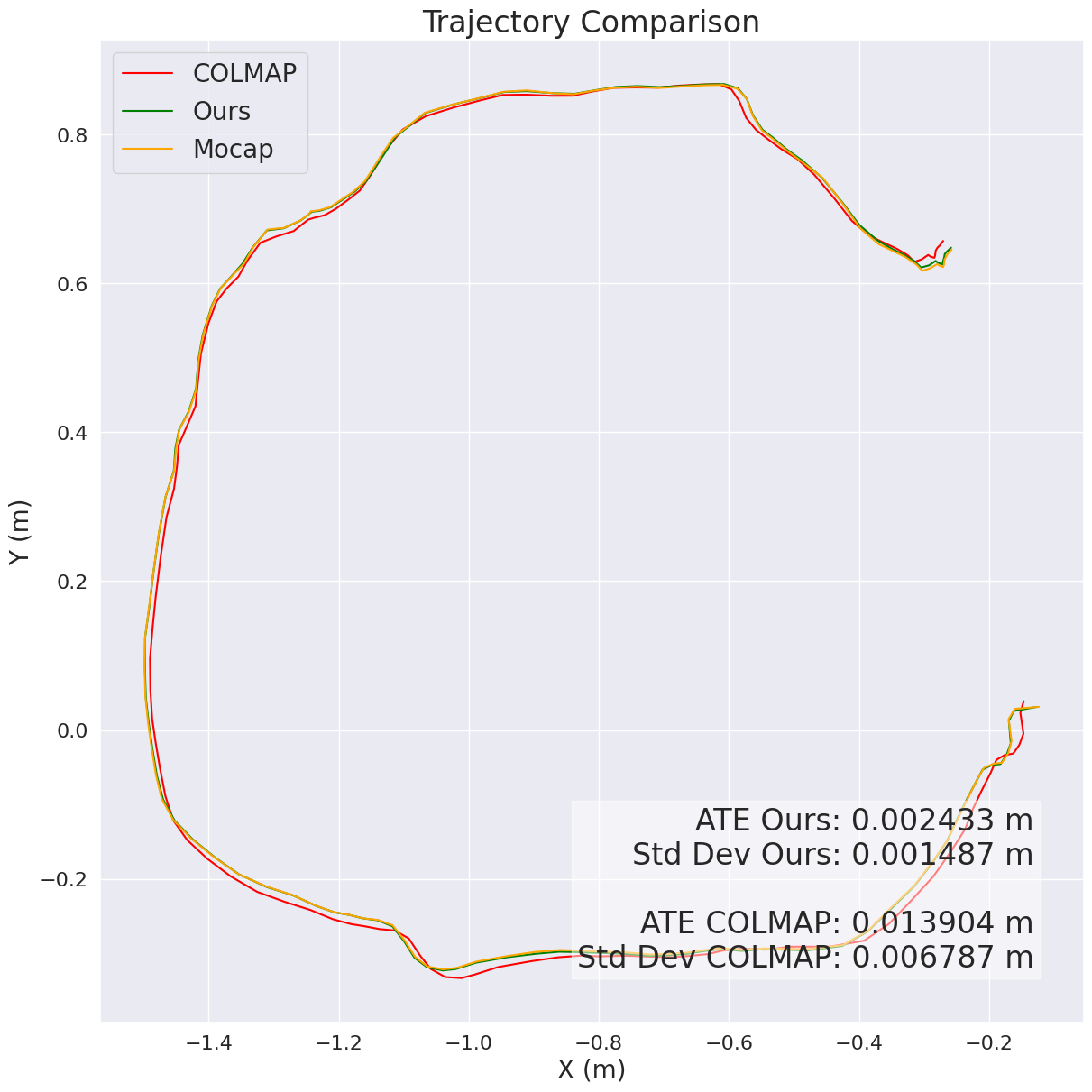}}
  \end{center}
  \vspace{-1.5em}
  \caption{Comparison of our ground truth trajectory with COLMAP and Vicon MoCap system. Our ground-truth has better accuracy than COLMAP.}
  \label{fig:millimeter_all}
  \vspace{-1.5em}
\end{figure}

\section{Ablation Study}
\label{supp:ablation}

\subsection{Complete Ablation Study}
In Tab. \ref{tab:ablation_all}, we report the complete ablation study, which shows the effectiveness of each of the components of our \projectname ground truth method and externally validates it with respect to MoCap ground truth. We observe that the optimal configuration for our ground truth system is when pose graph optimization and snapping are enabled, the final pose is multi-board PnP, Bundle PnP is applied and the board pose graph is constructed from a video filmed at close distance. Thus, we choose this configuration to calculate the ground truth for the \projectname benchmark.

In this experiment, we estimate the camera trajectory using both our \projectname method and a Vicon Motion Capture system in order to measure the accuracy of the \projectname method with different configurations enabled. The MoCap system estimates the pose of an object coordinate system built from the positions of reflective markers in view of the MoCap cameras. Thus, we place reflective MoCap markers on our camera rig in a fixed position. We then calibrate the relative pose between the marker object and the camera coordinates using Perspective-n-Point where we film a Vicon Active Wand with both the MoCap cameras and the 360-camera rig. This allows us to estimate the relative pose between the wand and the 360-camera, as well as the pose between the wand and the MoCap coordinates. Thus, we obtain the pose of the 360-camera in MoCap coordinate system. We measure the accuracy of our ground truth based on the Average Trajectory Error (ATE) with respect to the MoCap trajectory. 

We observe that the filming distance to calibration boards matters considerably in terms of the accuracy of our ground truth. In practice, this is the distance of the camera rig to the ground since we set up the boards on the ground. We observe that the board pose graphs built from videos that are filmed at a close distance (40-70cm) have lower ATE than those that are filmed at a medium distance (70 -150cm). For instance, in the 16 board setting where Snapping, Pose Graph Optimization and Multi-Board PnP are enabled, the close filmed pose graph - medium distance filmed frame has an ATE of 2.6 mm whereas the medium distance filmed pose graph - medium distance filmed frame has an ATE of 3.2 mm. Note that when collecting the \projectname benchmark, we film the calibration boards at a close distance to construct the pose graph, whereas we film the actual video that SLAM and COLMAP methods are evaluated on at varying distances to increase the diversity while preserving accuracy.

Pose graph optimization significantly improves the accuracy of our ground truth. This configuration makes the global board poses consistent with the relative measured poses between them. It almost halves the ATE when compared to the same settings with pose graph optimization disabled. For example, in the 16 board C-C filming the ATE goes from 7.1 mm accuracy to 3 mm accuracy when pose graph optimization is enabled. Thus, we always enable pose graph optimization when obtaining the ground truth for the \projectname benchmark. 

Snapping improves accuracy when the coplanarity assumption holds. This technique snaps the poses of the calibration boards to the same plane under the assumption that the boards are coplanar. For instance, with 16 boards and C-M filming and optimal configurations, the ATE goes from 7.0 mm to 6.5 mm when snapping is enabled. Note that we ensured that the boards are actually coplanar using a bubble level in this experiment. In general, we only enable snapping in \projectname benchmark when calibration boards are coplanar.

Final Pose indicates the method we use to calculate the ground-truth pose of the frame  when constructing the \projectname trajectory, before applying the Bundle PnP. At most frames, the ground truth view of the 360-camera will see multiple boards. 
\begin{itemize}
    \item Method 0 calculates the pose of the frame only with respect to the closest board using PnP. 
    \item Method 1 performs PnP with respect to each board and combines the estimated poses using a weighted average where the weights are determined by the reprojection errors. Specifically, we apply a confidence-weighted fusion approach. Since the camera sees several boards at once, we assign a confidence score to each detected board pose based on reprojection error and the number of markers detected. 
     Lower reprojection error signals a more accurate pose. At frame $t$, we weight each board's pose by its re-projection error:
    \begin{equation}
        W_{reproj}^{(i)} = \frac{1}{reprojection\_error_{(i,t)}}
        \label{eq:reprojection}
    \end{equation}
    Additionally, we set the confidence proportional to the number of markers detected on the board. We compute confidence as:
    \begin{equation}
        W_{num}^{(i)} = \frac{Number\ of\ Markers\ Detected\ on\ Board_{i}}{Total\ Numbers\ of\ Markers}
        \label{eq:confidence}
    \end{equation}

    Combining \ref{eq:reprojection} and \ref{eq:confidence} we define the total confidence as:
        \begin{equation}
            W_{total}^{(i)} = W_{reproj}^{(i)} \ W_{num}^{(i)}
        \end{equation}
     In order to determine the final camera pose, we calculate a weighted average of each detected board pose based on their confidence scores. Since the pose of a camera is represented by a translation vector $t$  and a rotation matrix $R$, we can compute them separately. The final translation $t_{final}$ is given by: 

    \begin{equation}
        t_{final} = \sum_i W_{total}^{(i)} T_i
    \end{equation}
    For the final rotation, we compute the weighted average of quaternions \cite{blow2004understanding, gramkow2001averaging}.

\item Final Pose 2 method identifies the markers on every board, transforms them to a single coordinate system and performs a single multi-board PnP to estimate the pose of the frame. We find this multi-board PnP method to be the most accurate configuration. For instance, the ATE with 8 boards, C-M filming decreases from $8.1 \text{mm} \rightarrow 6.8 \text{mm} \rightarrow  5.5 \text{mm}$ as we change the configuration closest board $\rightarrow$ weighted average $\rightarrow$ multi-board PnP.   
\end{itemize}

\begin{figure}
  \centering
  \subfloat[Baseline]{\includegraphics[width=0.495\linewidth]{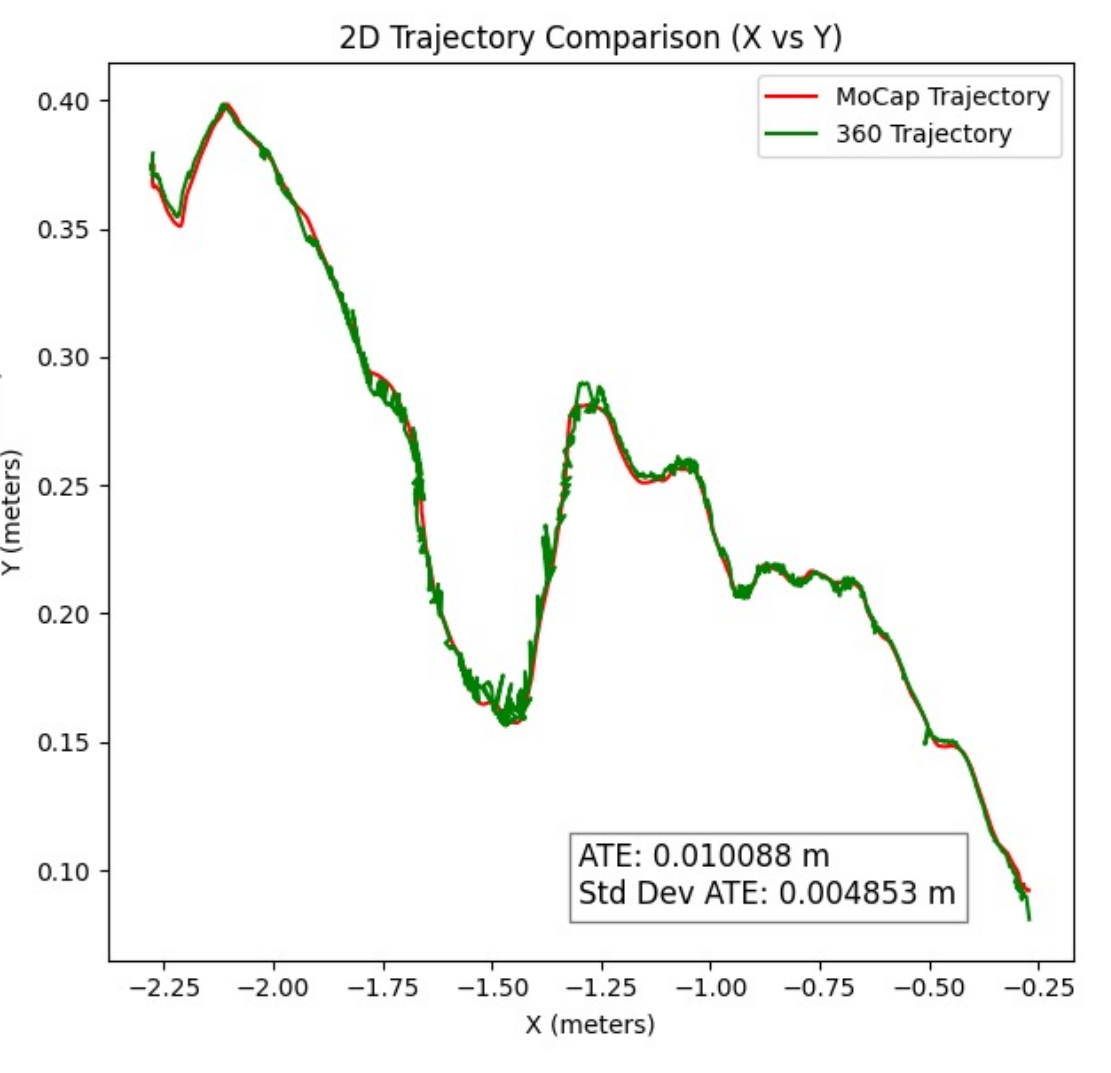}}
  \hfil
  \subfloat[Pose Graph Optimization]{\includegraphics[width=0.495\linewidth]{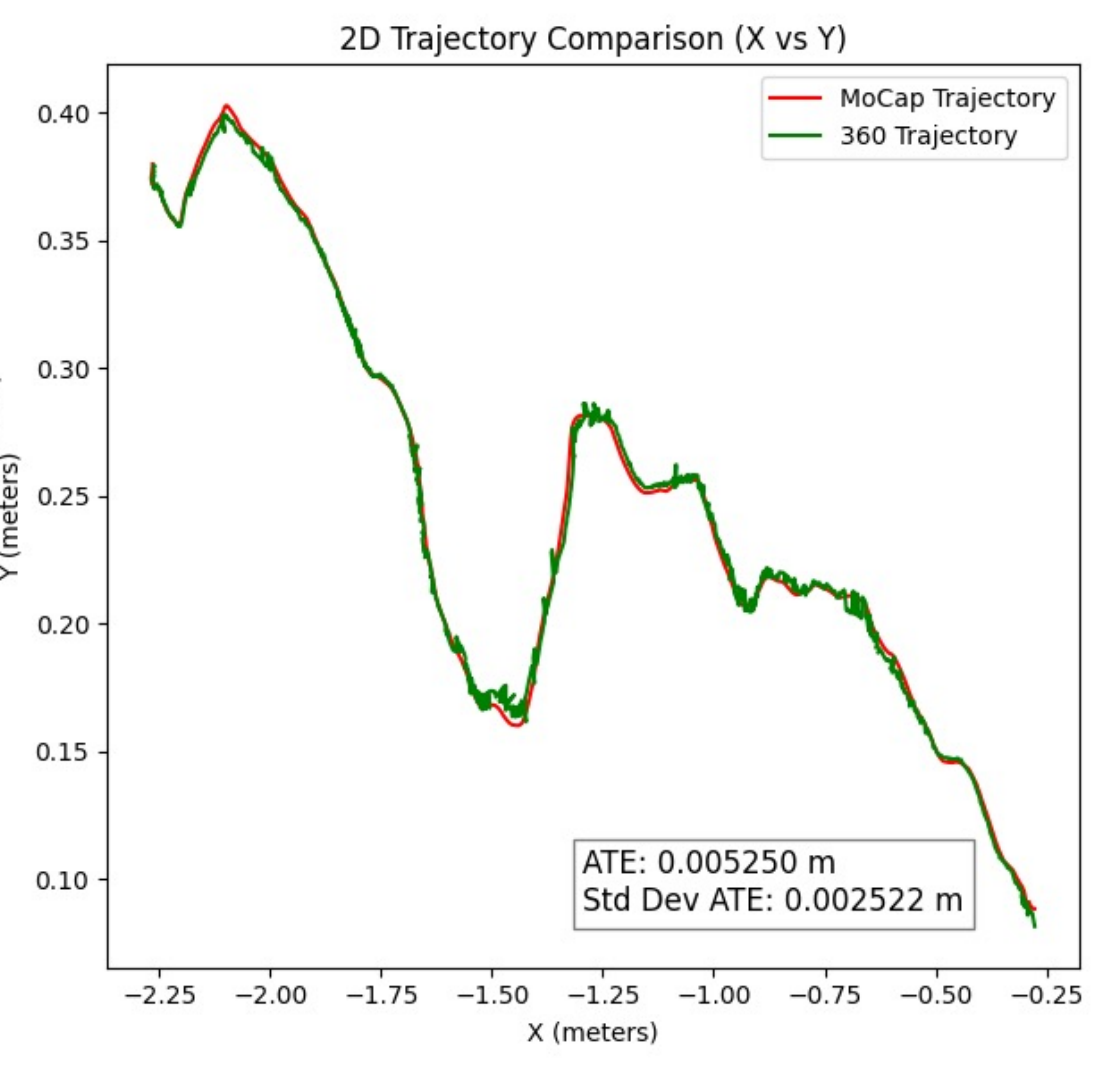}\label{fig:teaser:gt}}
  \\
  \subfloat[Snapping]{\includegraphics[width=0.494\linewidth]{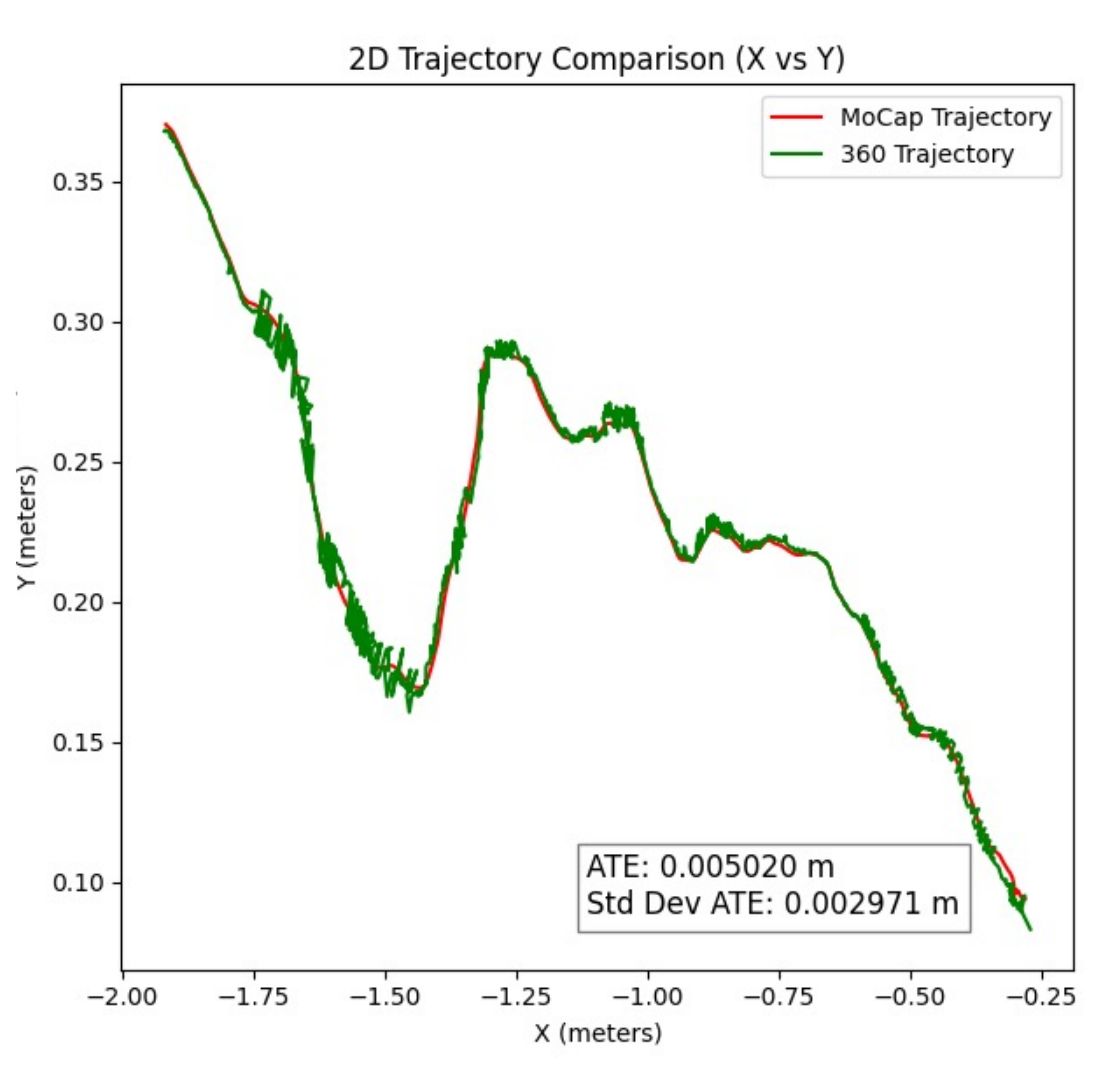}}
  \hfil
  \subfloat[ Pose Graph Optimization + Snapping]{\includegraphics[width=0.495\linewidth]{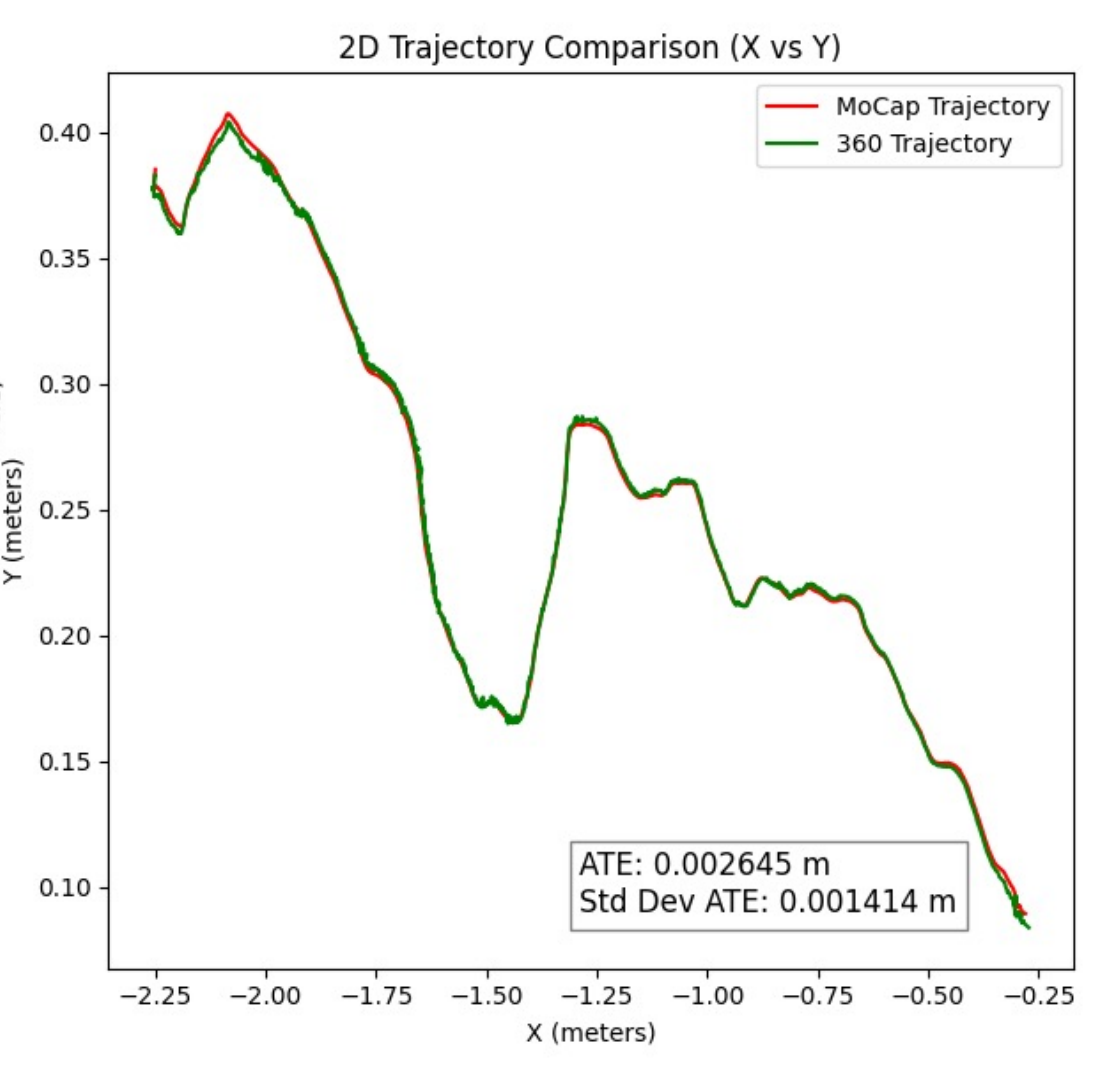}\label{fig:teaser:offset}}
  \caption{Comparison of our ground truth trajectory with MoCap. The figure illustrates the progressive reduction in ATE and standard deviation, while combining Pose Graph Optimization and/or Snapping. The combined approach achieves the best alignment to the ground truth trajectory.}
  \label{fig:optim_snap}
\end{figure}

We can further see the effect of these modifications in \cref{fig:optim_snap}.

\begin{figure*}[ht!]
  \centering

  \subfloat[]{\includegraphics[width=0.38\textwidth]{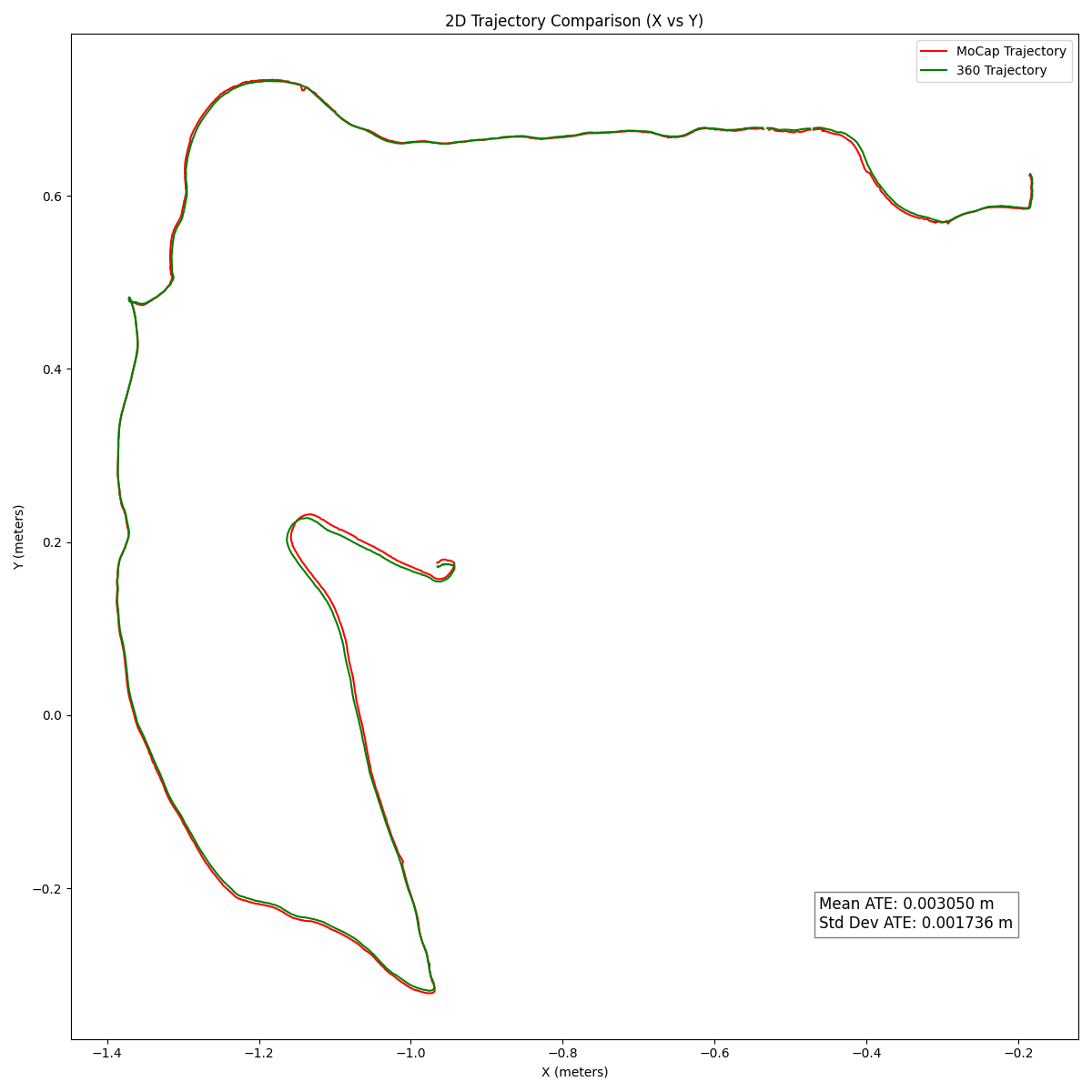}}
  \hfill
  \subfloat[]{\includegraphics[width=0.38\textwidth]{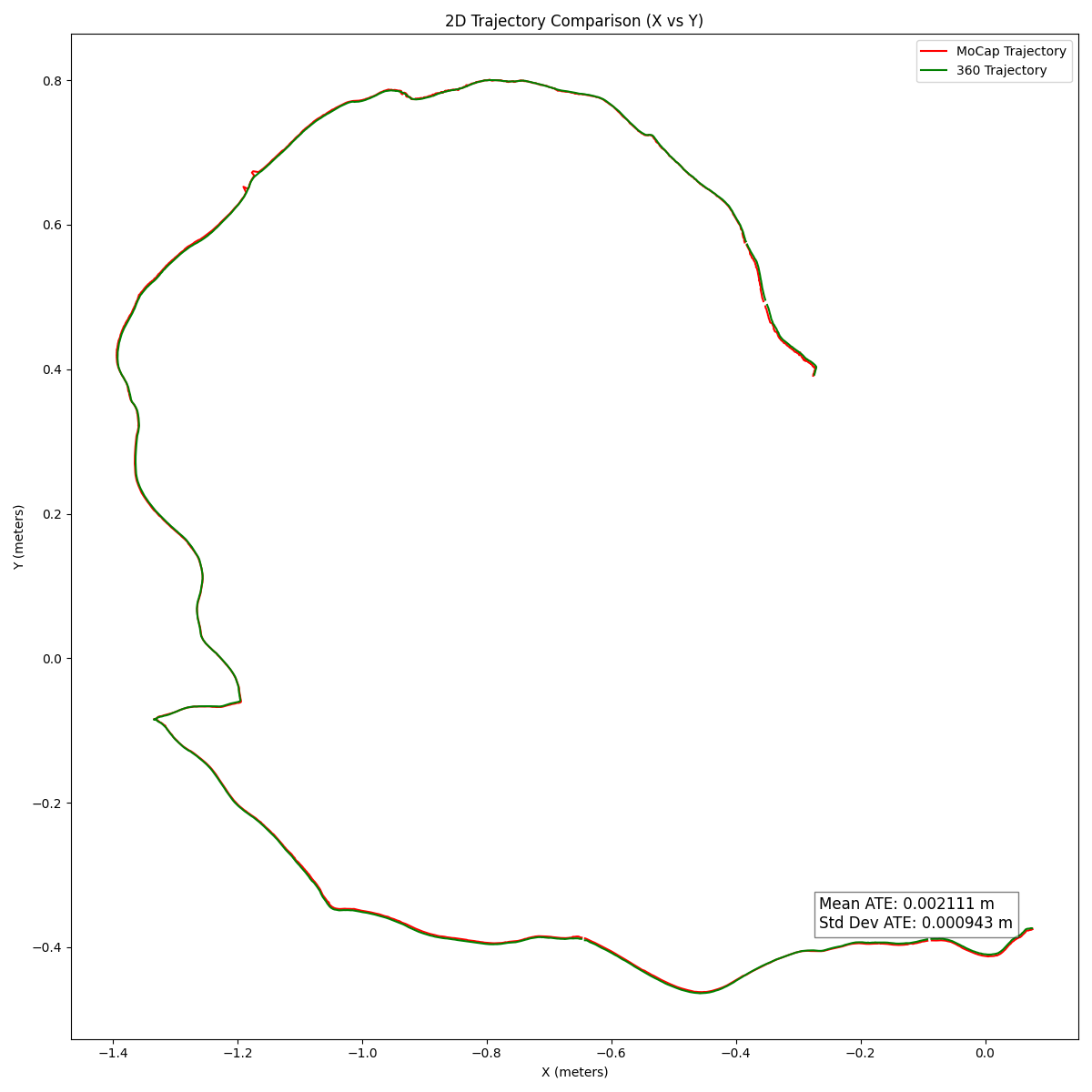}}\\

  \subfloat[]{\includegraphics[width=0.38\textwidth]{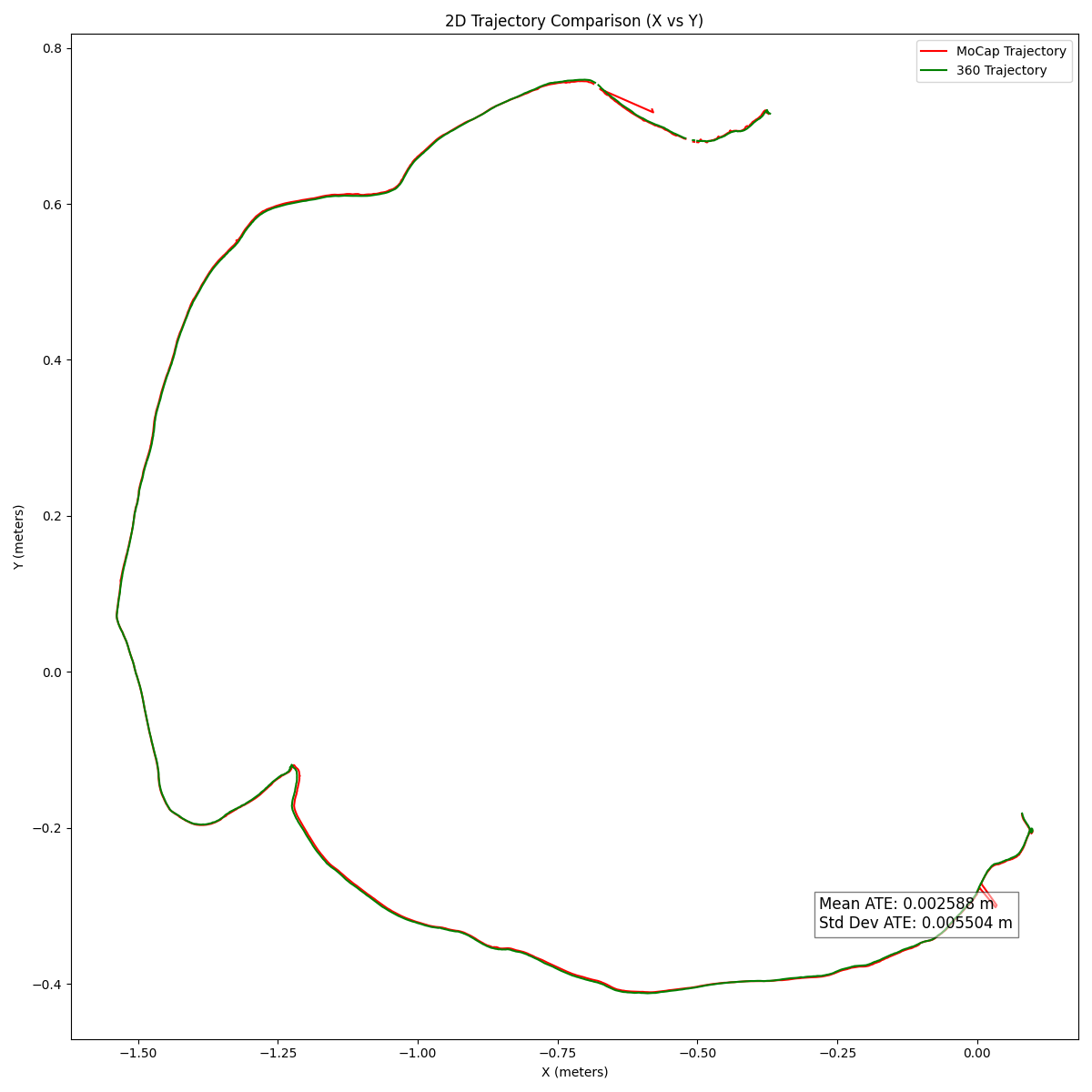}}
  \hfill
  \subfloat[]{\includegraphics[width=0.38\textwidth]{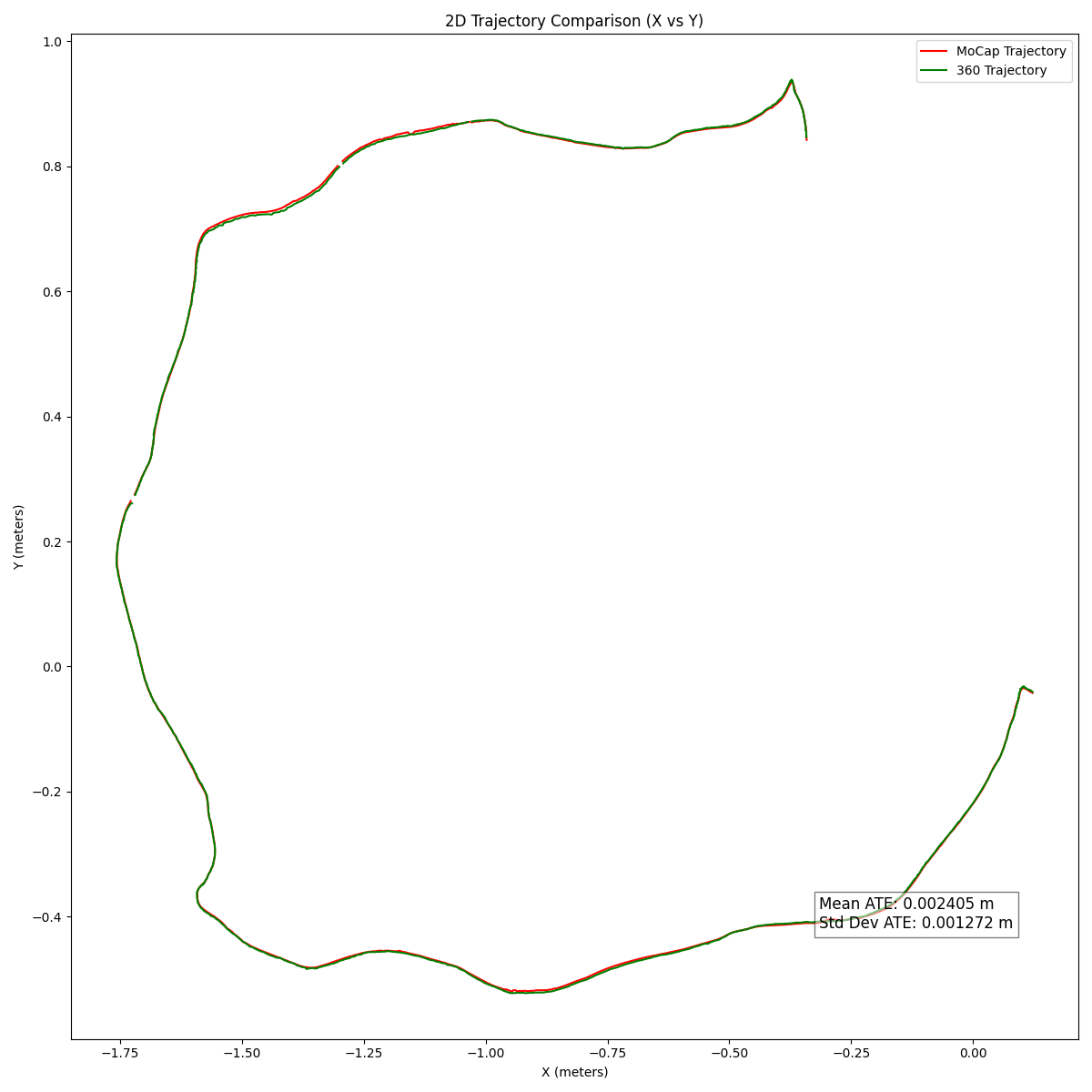}}\\

  \subfloat[]{\includegraphics[width=0.38\textwidth]{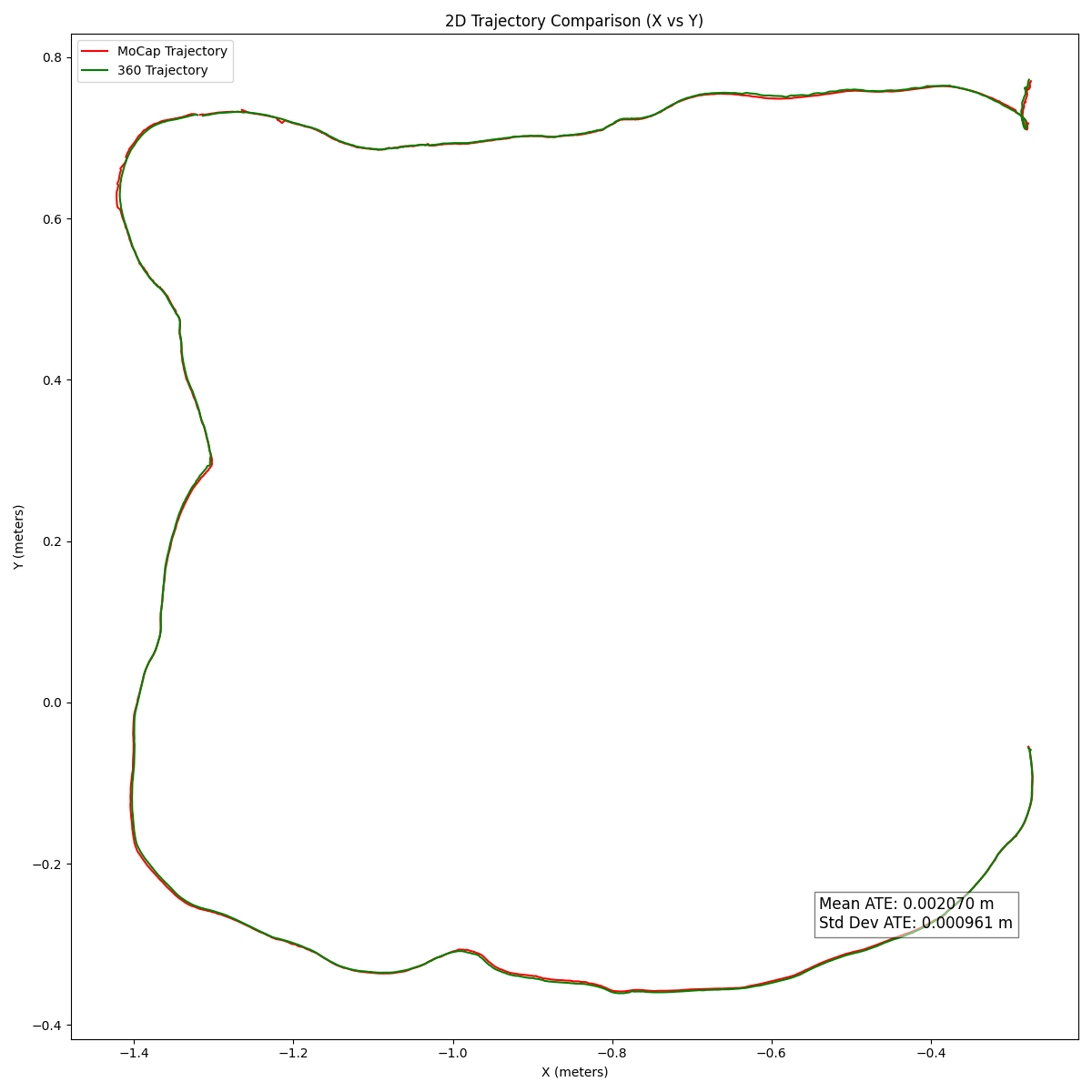}}
  \hfill
  \subfloat[]{\includegraphics[width=0.38\textwidth]{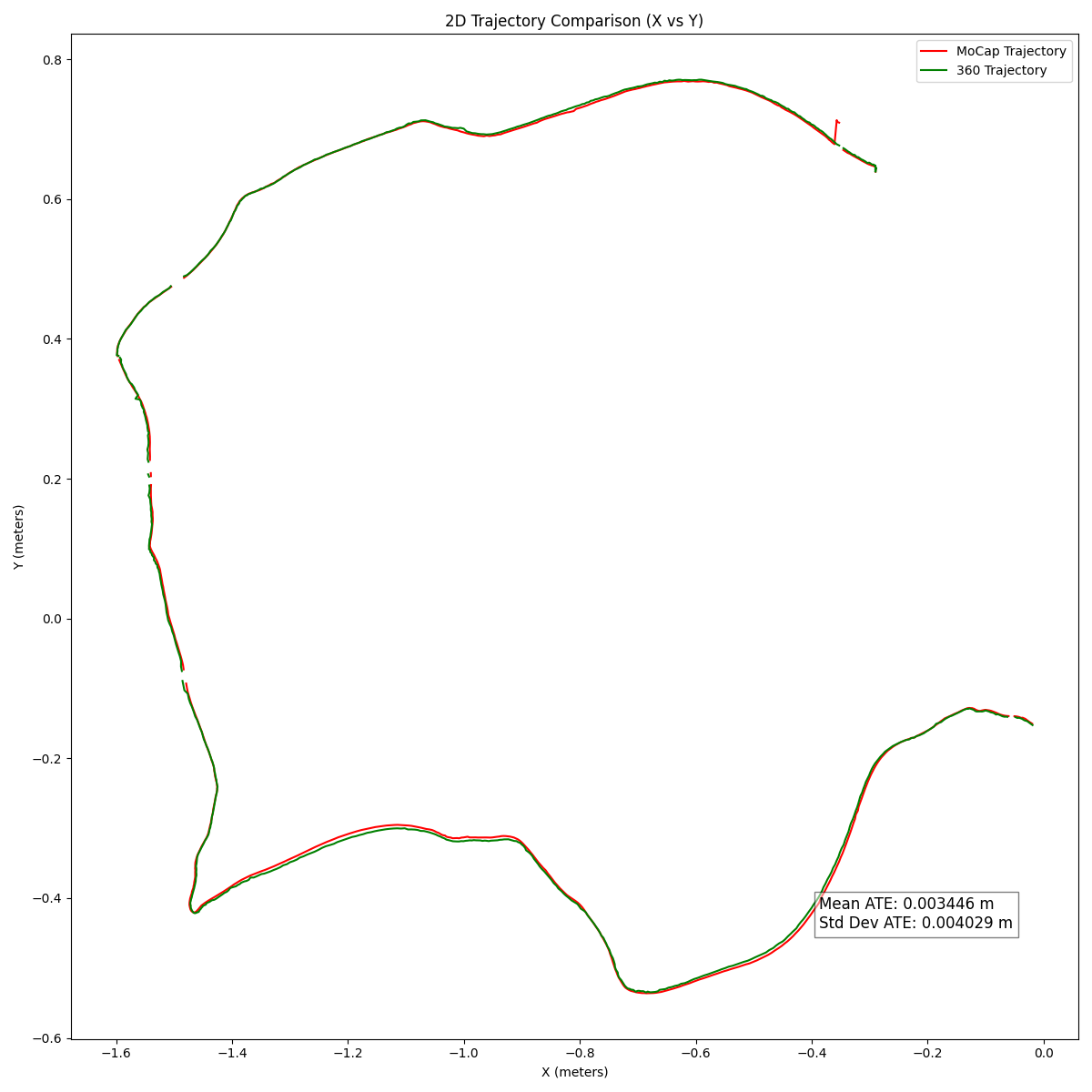}}

  \caption{X and Y comparison of six different trajectories as measured by MoCap and the 360 camera.}
  \label{fig:2dtraj}
\end{figure*}

Finally, we get the final camera poses by using the Bundle PnP. This optimization technique further improves the results. In general, more boards lead to a better pose since there are more points detected per frame. As shown in Table \ref{tab:ablation_all}, even with only 10 boards, applying Bundle PnP leads to  a lower ATE (2.1 mm) than using 16 boards without this optimization (2.6mm). In Fig. \ref{fig:2dtraj} we can see more results of our ground truth trajectories with the optimal settings compared to MoCap.

\begin{table}[ht]
\centering

\resizebox{\columnwidth}{!}{
\begin{tabular}{c|c|c|c|c|ccc|c|c}
\toprule
\# Boards & Filming  & Combination & Pose Graph & Snapping & \multicolumn{3}{c|}{Final Pose} & Bundle &Total ATE \\ 
\cmidrule(lr){6-8}
 & Distance& & Optimization & & 0 & 1 & 2 &PnP &(mm) \\ 
\midrule

 7  & C & C-C & ~ & ~ & \checkmark & ~ & ~  & &5.1  \\ 
7  & C & C-C & ~ & \checkmark & \checkmark & ~ & ~  & &3.1 \\ 
7  & C & C-C & ~ & \checkmark & ~ & \checkmark & ~  & &2.6 \\ 
7  & C & C-C & ~ & \checkmark & ~ & ~ & \checkmark  && \textbf{2.1} \\ 
7  & M & M-M & ~ & ~ & \checkmark & ~ & ~  & &22 \\ 
7  & M & M-M & ~ & \checkmark & \checkmark & ~ & ~  & &9.5 \\ 
7  & M & M-M & ~ & \checkmark & ~ & \checkmark & ~  & &7.2 \\ 
7  & M & M-M & ~ & \checkmark & ~ & ~ & \checkmark & & 9.6 \\ 
7  & M & C-M & ~ & ~ & \checkmark & ~ & ~  && 10 \\ 
7  & M & C-M & ~ & \checkmark & \checkmark & ~ & ~  & & 9.4 \\ 
7  & M & C-M & ~ & \checkmark & ~ & \checkmark & ~  & & 7.4 \\ 
7  & M & C-M & ~ & \checkmark & ~ & ~ & \checkmark  & &\textbf{4.2}  \\ 
\midrule
8  & C & C-C & ~ & ~ & \checkmark & ~ & ~  && 7.1 \\ 
8  & C & C-C & ~ & \checkmark & \checkmark & ~ & ~ & &5.2 \\ 
8  & C & C-C & ~ & \checkmark & ~ & \checkmark & ~ & &4.4 \\ 
8  & C & C-C & ~ & \checkmark & ~ & ~ & \checkmark & &\textbf{4.2} \\ 
8  & M & M-M & ~ & ~ & \checkmark & ~ & ~ & &13 \\ 
8  & M & M-M & ~ & \checkmark & \checkmark & ~ & ~  & &10 \\ 
8  & M & M-M & ~ & \checkmark & ~ & \checkmark & ~  & &7.5 \\ 
8  & M & C-M & ~ & ~ & \checkmark & ~ & ~  & &10 \\ 
8  & M & C-M & ~ & \checkmark & \checkmark & ~ & ~ & &8.1 \\ 
8  & M & C-M & ~ & \checkmark & ~ & \checkmark & ~ & &6.8 \\ 
8  & M & C-M & ~ & \checkmark & ~ & ~ & \checkmark  & &\textbf{5.5} \\ 
\midrule
16  & C & C-C & ~ & ~ & ~ & ~ & \checkmark & &22 \\ 
16  & C & C-C & \checkmark & ~ & ~ & ~ & \checkmark  & &7.5 \\ 
16  & C & C-C & ~ & \checkmark & ~ & ~ & \checkmark & &7.1 \\ 
16  & C & C-C & \checkmark & \checkmark & ~ & ~ & \checkmark & &3.0 \\ 
16  & M & M-M & ~ & ~ & ~ & ~ & \checkmark & &8 \\ 
16  & M & M-M & \checkmark & ~ & ~ & ~ & \checkmark  & &5.5 \\ 
16  & M & M-M & ~ & \checkmark & ~ & ~ & \checkmark & &6.7 \\ 
16  & M & M-M & \checkmark & \checkmark & ~ & ~ & \checkmark  & &3.2 \\ 
16  & M & C-M & ~ & ~ & ~ & ~ & \checkmark & &19 \\ 
16  & M & C-M & \checkmark & ~ & ~ & ~ & \checkmark & &7.0 \\ 
16  & M & C-M & ~ & \checkmark & ~ & ~ & \checkmark & &6.5 \\ 
16  & M & C-M & \checkmark & \checkmark & ~ & ~ & \checkmark & &\textbf{2.6} \\
\midrule

10  & M & C-M & \checkmark & \checkmark & ~ & ~ & \checkmark & & 4.4 \\
10  & M & C-M & \checkmark & \checkmark & ~ & ~ & \checkmark & \checkmark &\textbf{2.1} \\

\bottomrule
\end{tabular}
}
\caption{Ablation Study. This figure summarizes the effects of various configurations on ATE compared to MoCap. The number of boards is the number of ChArUco boards in the ground truth pose graph. The Filming distance indicates how far the camera rig is from the calibration boards where C stands for close (40-70cm) and M stands for medium (70 -150cm). Combination X-Y means that the pose graph is constructed from a video filmed X distance away and the trajectory is constructed from a video filmed Y distance away. Pose graph optimization makes the global board poses consistent with the relative poses between the boards, snapping encodes an assumption that the boards are all coplanar, and final pose indicates the method we use to get the ground truth pose of a frame.}
\label{tab:ablation_all}
\end{table}

\section{Calibration Boards}

\begin{figure}[ht]
  \centering
  \begin{subfigure}[t]{0.49\linewidth}  %
    \includegraphics[width=\linewidth]{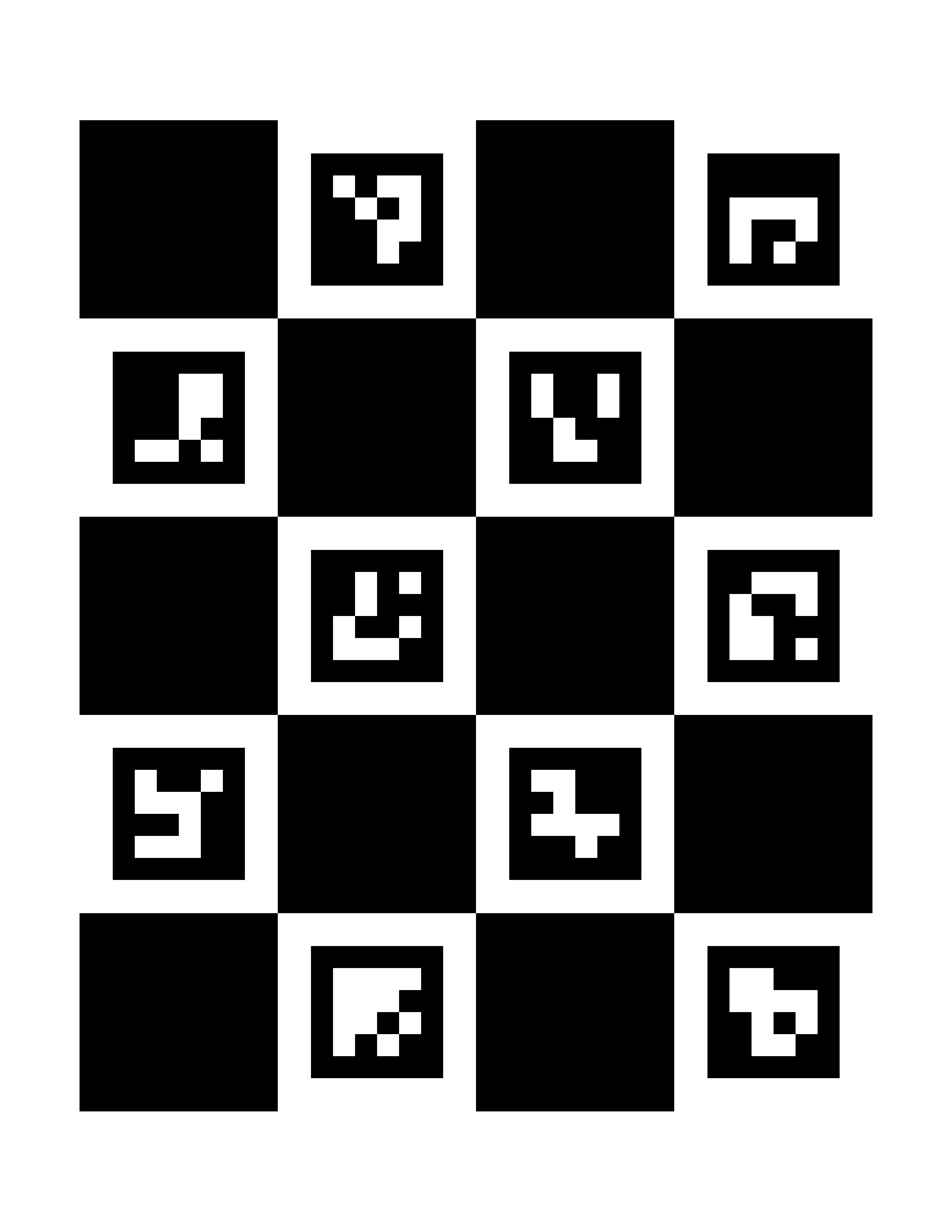}
    \caption{\textbf{ChArUco Board} 
    \label{fig:charuco_board}}
  \end{subfigure}
  \hfill
  \begin{subfigure}[t]{0.43\linewidth}  %
    \includegraphics[width=\linewidth]{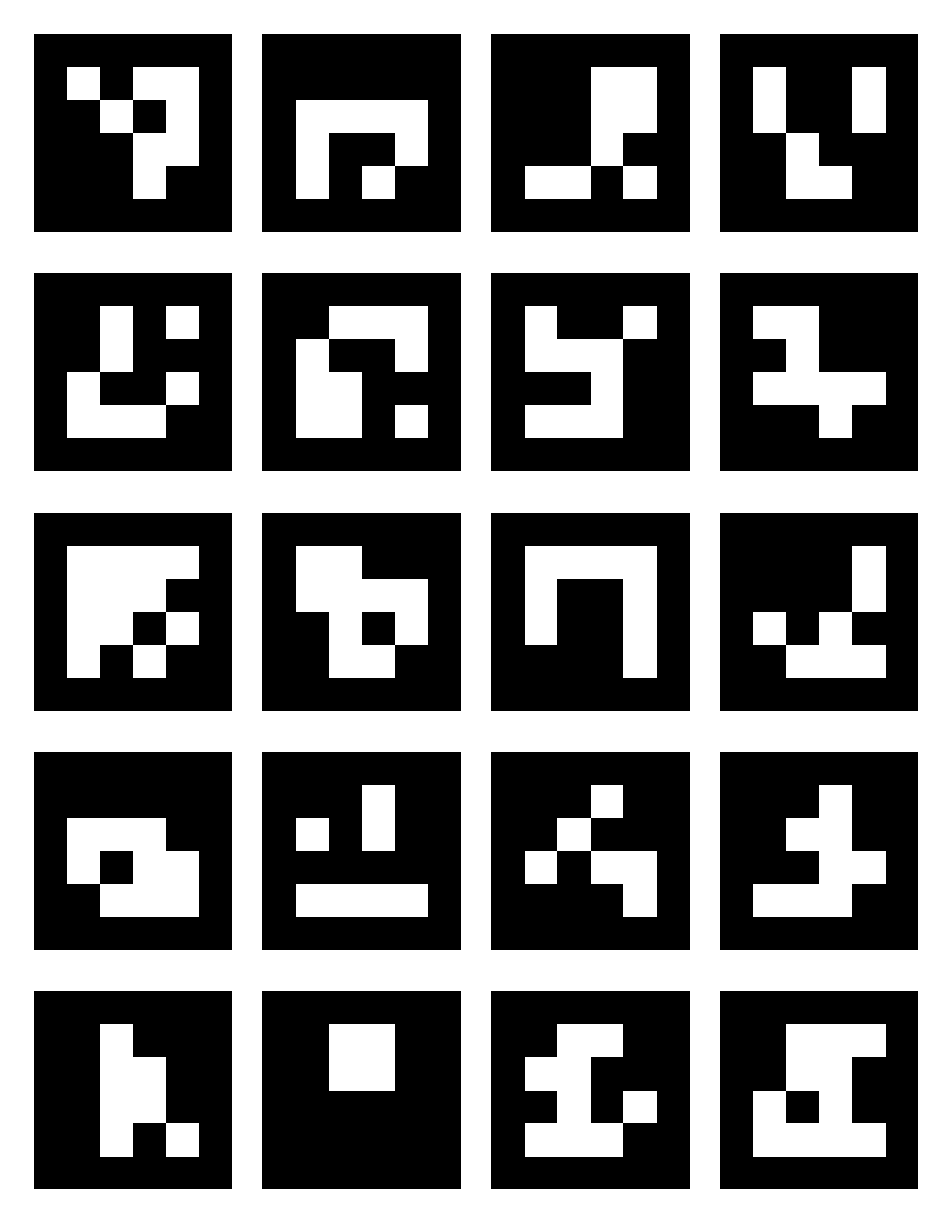}
    \caption{\textbf{GridBoard} 
    \label{fig:gridboard}}
  \end{subfigure}
  \vspace{-1em}
  \caption{Two types of fiducial calibration boards used in our benchmark: ChArUco and GridBoard. Each marker has a unique ID for pose estimation and pose graph construction.}
  \label{fig:boards}
\end{figure}

Fig. \ref{fig:boards} shows the calibration boards used in the \projectname benchmark. We evaluated two types of fiducial-based calibration patterns: ChArUco boards and GridBoards. ChArUco boards are a combination of ArUco markers and  chessboard, enabling subpixel corner refinement while retaining marker-level identification \cite{opencv_charuco}. GridBoards, on the other hand, are a structured grid of ArUco markers without using the chessboard layout. Thus, using the same paper size, GridBoards can contain larger quantity and size of Aruco markers compared to a Charuco board, offering robust detection across a broader range of viewpoints and lighting conditions, though without subpixel refinement \cite{opencv_aruco}.

In order to evaluate the accuracy of both board types, we placed several ChArUco boards and GridBoards side by side and recorded a single trajectory. At the same time we used MoCap system for ground truth. From the same recording, we extracted two separate trajectories—one using only ChArUco board detections and one using only GridBoard detections. Comparing both to the MoCap ground truth, we found that the trajectory estimated with GridBoards resulted in lower ATE as shown in Table \ref{tab:board_ate_comparison}. 

\begin{table}[ht]
  \centering
  \begin{tabular}{l c}
    \toprule
    \textbf{Calibration Board} & \textbf{ATE (mm)} \\
    \midrule
    ChArUco Board & 1.6 \\
    GridBoard     & 1.1 \\
    \bottomrule
  \end{tabular}
  \caption{Comparison of ATE for ChArUco and GridBoard using MoCap. GridBoard achieves lower ATE for the same trajectory, indicating higher trajectory accuracy.}
  \label{tab:board_ate_comparison}
\end{table}

\section{Depth Distribution}
\label{sec:supp_depth_dist}
The ZED X stereo camera returns pixel-wise depth values per frame. Since we want to compute the induced optical flow, one option would be to compute the induced flow directly using these depth samples. We instead fit a parametric distribution to these depth samples. This approach has a few advantages. Firstly, it is more robust to outliers, noise, and low amount of samples. More importantly, a parametric distribution allows us to have an analytical expression for the expectation of the induced optical flow. If we wanted to use depth samples directly, we would have to use a Monte Carlo estimate of the expectation. 

We fit the parametric depth distribution as follows. We take the depth samples for each frame of a sequence and concatenate them. Then, for every number of components from 1 to 8, we fit mixture of Gaussian and mixture of gamma distributions with that number of components. The reason for including the gamma distribution is because depth distributions are positive and the components tend to be skewed, which are properties satisfied by the Gamma distribution. Since higher the number of components, higher the likelihood, we need to balance the complexity of the distribution with the goodness of the fit. Thus, we take the best fitting distribution as measured by the Bayesian Information Criterion (BIC) \cite{bic}. 

\begin{figure*}[ht!]
    \centering
    \includegraphics[width=0.98\linewidth]{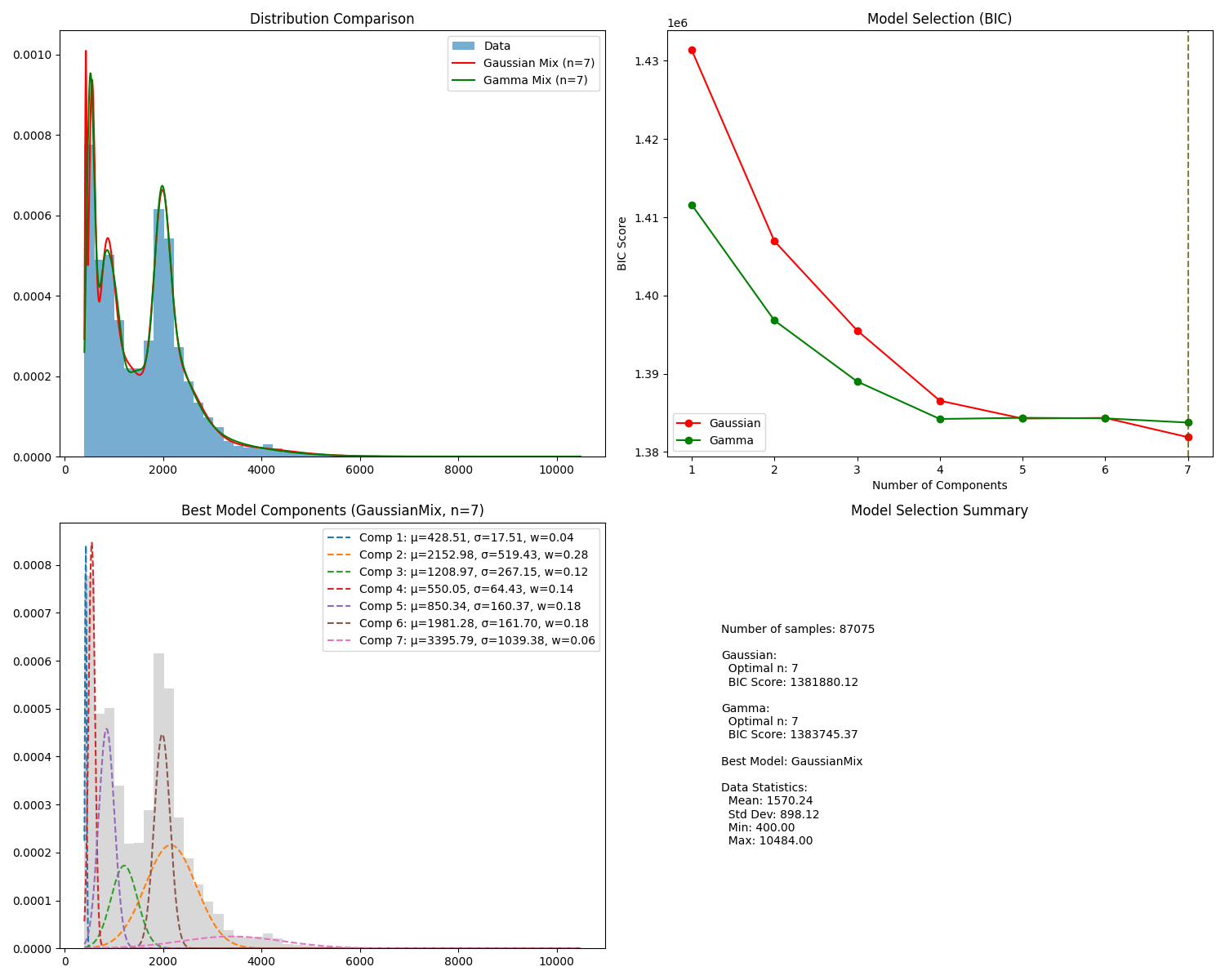}
    \captionsetup{justification=centering}  %
    \caption{An illustration of the parametric model we fit to the depth data. The figure shows fitted distributions, BIC scores for model selection, individual components, and summary statistics of the depth data.}
    \label{fig:depth_experiment}
\end{figure*}

An example of the fitting process for the parametric depth distribution is shown in \cref{fig:depth_experiment}. We show a histogram of a depth distribution, a comparison of the best mixture of Gaussian and mixture of Gamma fits, as well as the BIC selection process.

\section{Pose Graph Optimization}
\label{sec:pose_graph_optimization}
We want construct a pose graph where nodes represent global board poses $\{\mathbf{T}_{o}^{B_i}\}$ and edges represent measured relative poses $\mathbf{T}_{B_i}^{B_j}$ between boards $i,j$. The goal is to estimate a set of global poses $\{\mathbf{T}_{o}^{B_i}\}$ such that the relative pose computed from
them matches the observed local transformations:
\begin{equation}
{\mathbf{T}_{o}^{B_i}}^{-1} \mathbf{T}_{o}^{B_j} \approx \mathbf{T}_{B_i}^{B_j}
\end{equation}
The optimization is the following:
\begin{equation}
\label{eq:g2o_pose_graph}
\mathop{\arg\min}_{\{\mathbf{T}_{o}^{B_i}\}}
\;
\sum_{(i,j)\in\mathcal{E}}
\left\|
  \operatorname{Log}\!\Big(
        \big(\mathbf{T}_{B_i}^{B_j}\big)^{-1}
        \,{\mathbf{T}_{o}^{B_i}}^{-1}
        \,\mathbf{T}_{o}^{B_j}
  \Big)
\right\|_{\boldsymbol{\Omega}_{ij}}^{2}
\end{equation}
where $\operatorname{Log}:\mathrm{SE}(3)\to\mathfrak{se}(3)$ maps a pose to its 6-D Lie-algebra vector, and $\boldsymbol{\Omega}_{ij}$
 is the information matrix for the measurement $(i,j)$. We compute this using g2o\cite{g2o}.

\section{Bundle Rig PnP for two cameras}
\label{sec:bundle_rig_twocam}
As promised, we give the full expression for the Bundle Rig PnP optimization objective for the case of two cameras. We first remind that the general expression minimizes the reprojection error over all cameras while optimizing over the relative pose between the cameras and the rig: 
\[\min_{\mathbf{T}^o_{B_i}, \mathbf{T}^o_{rig, t}, \mathbf{T}_{rig}^c} \sum_{t = 1}^T \sum_c \sum_d \norm{\pi(\mathbf{T}^{c}_{rig}\mathbf{T}^{rig}_{o,t} \mathbf{T}^o_{B_i} p_{B_i}^{(d)}, \boldsymbol{\theta_c}) - u_{c,t}^{(d)}}^2\]

Taking the rig coordinate system to be the ground-truth view (camera 1) coordinate system, we have the optimization objective for the special case

\begin{equation}
\begin{split}
\min_{\mathbf{T}^o_{B_i}, \mathbf{T}^o_{gt, t}, \mathbf{T}_{gt}^u} \sum_{t = 1}^T \bigg[ \sum_d \norm{\pi(\mathbf{T}^{gt}_{o,t} \mathbf{T}^o_{B_i} p_{B_i}^{(d)}, \boldsymbol{\theta_{gt}}) - u_{gt,t}^{(d)}}^2 \\
+ \sum_d \norm{\pi(\mathbf{T}^{u}_{gt}\mathbf{T}^{gt}_{o,t} \mathbf{T}^o_{B_i} p_{B_i}^{(d)}, \boldsymbol{\theta_u}) - u_{u,t}^{(d)}}^2 \bigg],
\end{split}
\end{equation}
which gives us the relative pose between the user view and the ground truth view denoted $\mathbf{T}_{gt}^u$. We use this optimization objective when we estimate the relative pose between the ground truth views and the user views. We run the optimization for each rendered user view. 

\section{Validation of Bundle Rig PnP}
In this section, we show that the estimated relative pose between the user view and the ground truth view is accurate. The idea is to move the camera in two different trajectories, run Bundle Rig PnP to estimate a relative pose for each trajectory, and compare the two relative poses to make sure that they are consistent. If Bundle Rig PnP works well, the two estimated poses should be the same since the views are fixed. 

We set up 8 different board configurations where the boards are placed so that both views can see boards at the same time. For each configuration, we film two sequences with different trajectories. We render the ground-truth view and the user-view with fixed relative pose for both sequences. Note that for each configuration we render different relative poses used in the benchmark to ensure that all of them are accurate. We then run Bundle Rig PnP to estimate two relative poses, one for each sequence. Finally, we compare the poses to each other in terms of translation and rotation difference. 

The results in \cref{tab:relative_pose_diffs_cases} show that the rotation difference between the two estimated poses is $0.070^\circ$ on average, and the translation difference is $0.89$ mm on average. Thus, we conclude that the relative pose between the user-view and the ground-truth view is accurate. 
\begin{table}[t]
\centering
\begin{tabular}{ccc}
    \toprule
    Experiment & Rotation Diff. (deg) & Translation Diff. (mm) \\
    \midrule
    1 & 0.0712 & 1.04 \\
    2 & 0.0447 & 0.51 \\
    3 & 0.0543 & 1.08 \\
    4 & 0.0792 & 1.04 \\
    5 & 0.0576 & 0.39 \\
    6 & 0.0894 & 1.16 \\
    7 & 0.0814 & 1.32 \\
    8 & 0.0974 & 0.59 \\
    \midrule
    \textbf{Mean} & \textbf{0.0705} & \textbf{0.891} \\
    \bottomrule
\end{tabular}
\caption{Relative pose differences for 8 cases using Bundle Rig PnP. It shows that the estimated relative pose between the user view and the ground truth view remains consistent across different trajectories. Thus, we conclude that the relative pose calibration is accurate.}
\label{tab:relative_pose_diffs_cases}
\end{table}

\section{Bundle Rig PnP outperforms regular PnP}

In this section, we show that Bundle Rig PnP calibration outperforms naive PnP when estimating the relative pose between the user-view and the ground-truth view. 

To evaluate naive PnP, we placed reflective markers that can be seen by both MoCap and the user-view of Insta360. We annotate the 2D pixel position of each marker for multiple frames. Running PnP gives us the relative pose between user-view and the markers we have placed. Since the position of the markers can be tracked by MoCap, we obtain the pose of the user-view in MoCap coordinates by chaining poses together. We use this to obtain the fixed relative pose between the Insta360 body (constructed from reflective markers placed on Insta360) and the user-view. We perform a similar process to obtain the relative pose between the ground-truth view and the Insta360 body. Hence, we obtain the relative pose between the user-view and the ground-truth view through just using PnP. 

How does this naive approach compare to Bundle Rig PnP? Since we cannot obtain the relative pose between the two views through an external measurement, we have to rely on internal validation for comparison. 

\begin{figure}
  \centering
  \subfloat[Distance Variation]{\includegraphics[width=0.495\linewidth]{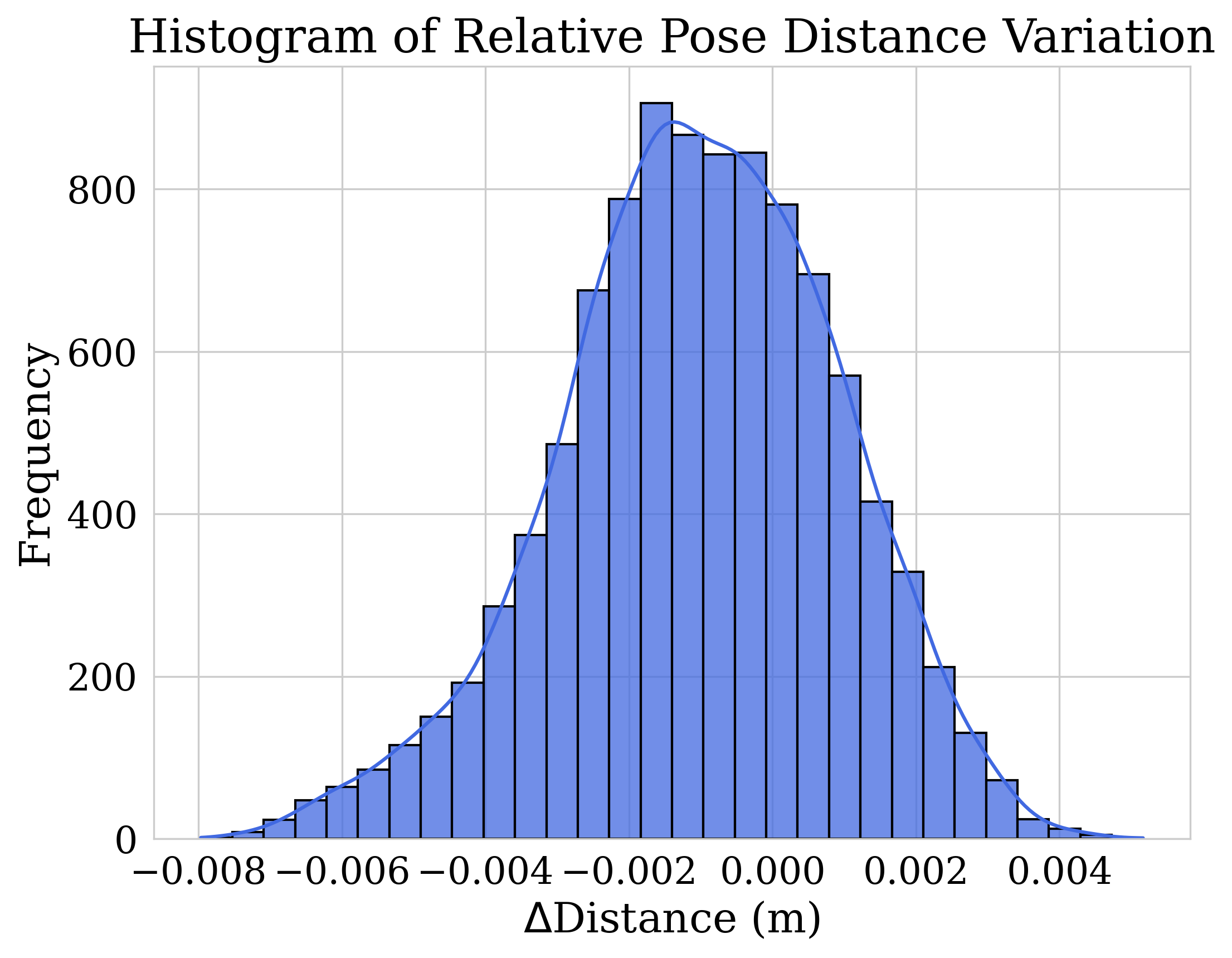}}
  \hfil
  \subfloat[Angle Variation]{\includegraphics[width=0.495\linewidth]{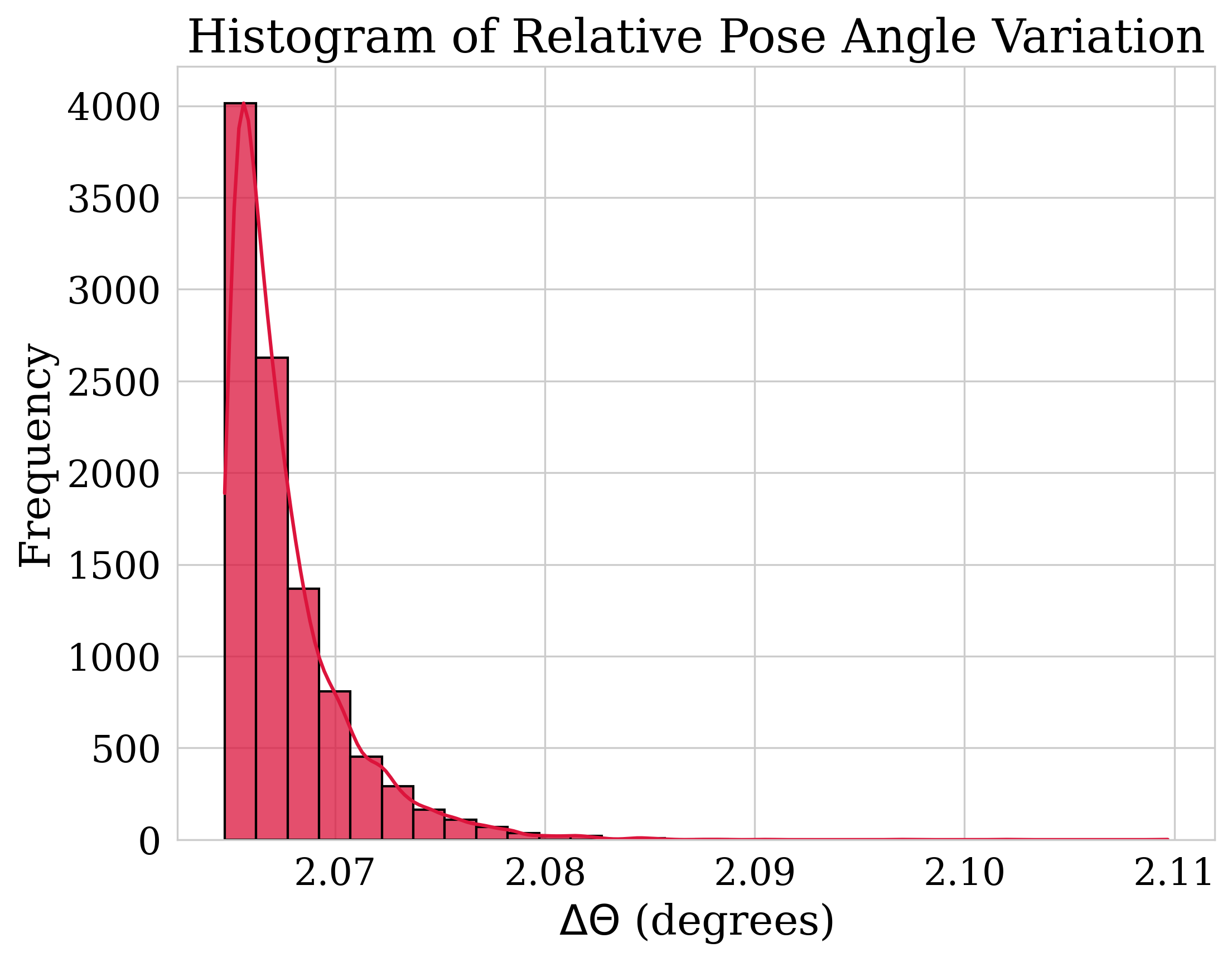}}
  \\
  \caption{Histograms of both distance and angle variation obtained via bootstrapping Naive PnP. We observe that Naive PnP is not as reliable as Bundle Rig PnP when it comes to estimating the relative pose between the user-view and the ground-truth view. The histograms show high variance for the naive PnP estimates. Difference in distance is given in meters and the difference in angle is given in degrees. A total of 10,000 trials were conducted.}
  \label{fig:pose_variance}
\end{figure}

We measure the variance of the estimated relative pose between the ground-truth view and the user view of the 360 camera by using a bootstrapping approach. We take all the 2D-3D correspondences used in the PnP calculation of the relative pose, and we resample with replacement to recalculate the relative pose. We construct a histogram of the translation and the rotational distance from the original relative pose performing 10,000 Monte Carlo trials. \cref{fig:pose_variance} shows the that the variance of the poses obtained by naive PnP is significantly higher than the variance of the poses obtained by Bundle Rig PnP. In particular, note that a large portion of the rotations of Naive PnP in \cref{fig:pose_variance} fall higher than 2 degrees in stark contrast to the average rotation difference of 0.07 deg obtained by Bundle Rig PnP shown in \cref{tab:relative_pose_diffs_cases}.

\section{Further GT validation with MoCap}
As mentioned in \cref{sec:experiment}, we validate our ground-truth pipeline with a Vicon Vantage V16 MoCap system. However, MoCap does not work outdoors. As a result, all validation sequences in \cref{tab:ate_comparison_colmap} have been recorded indoors. The indoor and outdoor sequences are not significantly different from a ground-truth perspective since the ground-truth view films calibration boards on the ground in both cases where the main difference is the indoor vs outdoor lighting. However, we still perform further experiments to make sure that the indoor ground-truth validation generalizes to outdoor sequences as well. 

We measure reprojection errors coming from Bundle PnP for both indoor and outdoor sequences. The ground-truth accuracy, as measured by reprojection error, appears consistent across indoor and outdoor environments, where the average indoor reprojection error is 0.7373 px and the average outdoor reprojection error is 0.7394 px. We use the Welch's t-test to test if there is a statistically significant difference between the two cases. The difference between indoor and outdoor averages is 0.0021 px where the 95\% confidence interval for the difference is [-0.0175, 0.0217] and the p-value is p=0.832, meaning that the difference measured is not statistically significant. As a result, we conclude that the validation experiments showing that our ground-truth is mm accurate generalize to outdoor sequences as well.

\section{Computing Induced Optical Flow}

Recall that the IOF metric can be written as 
\[\mathrm{IOF} = \frac{1}{AT}\sum_t \sum_{u,v} \int_{0}^{\infty} \norm{\mathbf{flow}(t,d,u,v)}_2 \, p(d) \,dd.
\]
We sample $(u,v)$ over a uniform grid of pixels in the image denoted $A$. After fitting the parametric depth distribution $p$, we calculate minimum and maximum depth values for this distribution. For the mixture of Gaussians, the $d_{min}$ is calculated as the value 4 standard deviations smaller than the mean of components. We take the smallest value over all components.  $d_{max}$ is calculated as the value 4 standard deviations higher than the mean of components. Again, we take the largest value. Thus, we write the IOF metric as 
\[\frac{1}{AT}\sum_t \sum_{u,v} \int_{d_{min}}^{d_{max}} \norm{\mathbf{flow}(t,d,u,v)}_2 \, p(d) \,dd.\]

Assuming we have calculated $\norm{\mathbf{flow}(t,d,u,v)}_2$, the inner integral can be numerically integrated. 

Now, we talk about how to calculate $\mathbf{flow}(t,d,u,v)$. We denote the trajectory collected by our ground truth method as $\mathbf{T}_{\text{gt}}^{(t)} \in \mathbf{SE}(3)$ and the trajectory estimated by the SLAM method as $\mathbf{T}_{\text{est}}^{(t)} \in \mathbf{SE}(3)$. Here $t$ denotes the frame or timestamp of the trajectory.

\label{sec:computing_IOF}
We take $\mathbf{K}$ to be the $4 \times 4$ Camera Intrinsic Matrix in Homogeneous Coordinates:
\[
\mathbf{K} = \begin{bmatrix}
f_u & 0 & c_u & 0 \\
0 & f_v & c_v & 0 \\
0 & 0 & 1 & 0 \\
0 & 0 & 0 & 1
\end{bmatrix},
\]
where \( f_u, f_v \) are the focal lengths in the u and v directions (pixels), and \( c_u, c_v \) are the principal point coordinates.

We denote a pixel sampled in the ground truth trajectory at frame $t$ as $\mathbf{p}_{\text{gt}, (u,v)}^{(t)}$. This can be written in homogeneous coordinates with inverse depth $\rho_{(u,v)}^{(t)}$ as 
\[
\mathbf{P}_{\text{gt}, (u,v)}^{(t)}= \begin{bmatrix}
u \\
v \\
1 \\
\rho_{(u,v)}^{(t)}
\end{bmatrix}
,\]
where the depth comes from the numerical integration over the depth distribution. 

The re-projection of $\mathbf{P}_{\text{gt}, (u,v)}^{(t)}$ to frame $t$ of the estimated trajectory can be calculated as:
\[
\mathbf{P}_{\text{est}, (u,v)}^{(t)} = \mathbf{K}\, \mathbf{T}^{(t)}_{\text{est}}\, \mathbf{T}_{\text{gt}}^{(t)-1}\, \mathbf{K}^{-1}\, \mathbf{P}_{\text{gt}, (u,v)}^{(t)},
\]
where we unproject the pixel, transform it using the relative pose between the estimated and ground truth frames, and reproject it to the estimated frame. The induced flow is the difference between inhomogeneous coordinates:

\[
\mathbf{flow}(t,d,u,v) =  \mathbf{p}_{\text{est}, (u,v)}^{(t)} - \mathbf{p}_{\text{gt}, (u,v)}^{(t)}.
\]

\section{Trajectory Alignment Details}
\label{sec:supp_traj_alignment}
In this section, we denote frames with $i$ to avoid confusion. Decomposing the 6 DoF pose $\mathbf{T}^{(i)}$ into the rotation $\mathbf{R}^{(i)} \in \mathbf{SO}(3)$ and translation $\mathbf{t}^{(i)} \in \mathbb{R}^3$, we can write the initial $\mathbf{Sim}(3)$ alignment of the translation part as 

\[
\min_{s \in \mathbf{R}^+,\ \mathbf{R} \in \text{SO}(3),\ \mathbf{t} \in \mathbb{R}^3} \sum_{i=1}^{T} \left\| s\, \mathbf{R}\, \mathbf{t}_{\text{est}}^{(i)} + \mathbf{t} - \mathbf{t}_{\text{gt}}^{(i)} \right\|^2.
\]

We can write the optimization objective for the secondary alignment on the orientation of the estimated trajectory as

\[
\min_{\mathbf{R}_{\text{align}} \in \text{SO}(3)} \sum_{i=1}^{T} \left\| \log\left( \mathbf{R}_{\text{align}}\, \mathbf{R}_{\text{est}}^{(i)}\, (\mathbf{R}_{\text{gt}}^{(i)})^\top \right)^\vee \right\|^2.
\]

We perform both optimizations using SVD. 

\section{Coverage and Composite Score}
\label{sec:supp_cov_comp_score}

A lot of SLAM methods do not output poses for the whole sequence. For example, ORB-SLAM \cite{mur2015orb} will take a while to initialize. This means that first couple of hundred frames will not have a pose. Also, SLAM methods might fail on certain sequences, meaning that they will not output any pose for that sequence.

This creates a precision and recall issue. If we report Flow AUC computed only from frames that have a pose, then models that output only a few poses with high confidence are rewarded, punishing robust methods. 

What if we forced the methods to output the same exact number of frames as our ground-truth, a strategy followed by benchmarks like KITTI \cite{kitti}? In that case, a method would have to pad its gaps with pose guesses. For instance, some methods fill in the gaps with the closest available estimated pose. This strategy will cause disproportionately low Flow AUC when poses for certain frames are missing. Also, the strategy does not handle the case when a method completely fails on certain sequences. 

Our solution is to create a composite score from Induced Optical Flow Metric, and tracking coverage. Tracking coverage is roughly for what percent of frames there is an estimated pose. It can be thought of as a measure of recall. In particular, we compute

\[\text{coverage} = \frac{\# \text{frames predicted}}{\# \text{frames total}}.\]

Note that coverage includes failed sequences as well. Furthermore, coverage has the same range as Flow AUC where both metrics range from $0\%$ to $100\%$. This allows us to easily combine the two scores.  

The composite score, then, is the harmonic mean (f-score) of Average Flow AUC and coverage, which is similar to f-score computed from precision and recall:

\[\text{composite} = \frac{2}{\frac{1}{\text{AUC}_{\text{avg}}} + \frac{1}{\text{coverage}}}\]

Thus, we sort our leaderboard using the composite score by default. This rewards methods that are both accurate and robust.

\section{Experimental Setup}
\label{sec:supp_exp_setup}

In our evaluation Table \ref{tab:evaluation},  we ran DPVO and DPV SLAM with their default settings, processing all sequences at stride 1. As for LEAPVO, we also used the default configuration for the scanning sequences at stride 1. However, we used stride 10 for the indoor and outdoor scenes since increasing the buffer size for longer sequences still resulted in failures.
Lastly, we ran COLMAP at stride 1 with default parameters and sequential match, without intrinsic refinement. We considered a sequence failed if processing required more than 2 days and 16 hours.

\section{Rig Construction}
\label{sec:supp_rig_construction}

Regarding the rig construction, we mounted the Insta 360 camera and the ZED camera using a smallrig Clamp along with a 9.8-inch Adjustable Friction Power Articulating Magic Arm. This setup allowed us to adjust the alignment of both cameras, so as the Insta360's user view closely matches with that of the ZED one.  Once we finalized an acceptable configuration we also applied epoxy glue to the joints, in order to ensure that the relative pose between the ZED and 360 camera remained stable.  Epoxy is a high-strength structural adhesive, which provides extra stability beyond the clamp itself. The complete rig setup can be seen in Fig. \ref{fig:hardware}. 

\section{NVIDIA Jetson Optimizations}

\subsection{Hardware Tuning}
Before achieving maximal data recording frequencies for RGB or IMU data by themselves, we needed to perform additional hardware-related optimizations for the NVIDIA Jetson. These optimizations are crucial, as they introduce about a 3x frequency increase just by themselves.

The first optimization involves \texttt{nvpmodel}, which forces the NVIDIA Jetson to draw maximal power from its power source. This ensures the camera is receiving the most amount of power it is designed to handle, and thus can record at maximum frequency.

The second optimization involves \texttt{jetson\_clocks}, which sets static max frequency to CPU, GPU, and EMC clocks on the NVIDIA Jetson. This ensures that the Jetson's compute resources are being utilized to their maximum potential when recording data.

By combining these different hardware optimizations together, we are able to achieve 60 FPS video recording by itself, or 400 Hz IMU data recording by itself.

\subsection{Software Parallelization}

Although RGB video and IMU data can be collected at their
respective maximum frequencies when recording by them-
selves, new difficulties arise when trying to record both RGB
and IMU together. We looked towards techniques in parallel computing to try and preserve maximal data recording frequency. The primary methods we looked into are multiprocessing and multithreading.

Both multiprocessing and multithreading strive to improve total processor performance and therefore position themselves to decrease the processing time for any application that exposes concurrent software threads for execution. The two technologies however take different approaches in the hardware to address these goals and will subsequently offer different levels of success for any particular example of software code. 
\begin{itemize}
    \item \textbf{Multithreading}  refers to the ability of a processor to execute multiple threads concurrently, where each thread runs a process. 
    \item \textbf{Multiprocessing} refers to the ability of a system to run multiple processors in parallel, where each processor can run one or more threads.
\end{itemize}

Multithreading is beneficial for IO-bound tasks, such as reading files from a network or database, because it allows each thread to concurrently execute these processes. In contrast, multiprocessing is more suitable for CPU-bound tasks that require substantial computational resources, as it can utilize multiple processors, mimicking the efficiency of multicore systems.

There is a clear distinction between concurrency and parallelism: concurrency involves executing multiple tasks in an interleaved manner, one at a time, whereas parallelism involves executing multiple tasks simultaneously.

Python's global interpreter lock (GIL) restricts multithreading to executing only one thread at a time, meaning that it only offers concurrency, not parallelism, particularly for IO-bound processes. However, multiprocessing enables parallel execution.

Using multithreading for CPU-bound tasks can degrade performance, due to the limited execution capabilities under the GIL and the overhead associated with managing multiple threads.

Although multiprocessing can be applied to IO-bound processes, it generally incurs greater overhead than multithreading. Yet, it can lead to increased CPU usage, which is expected given that multiple CPU cores are engaged by the application.

Throughout our experimentation with the ZED X camera to optimize data synchronization and processing, we found out that although multiprocessing allowed parallel data handling, it introduced blocking issues that reduced IMU performance. Multithreading improved responsiveness and IMU capture rate by reducing overhead, offering a more continuous data flow.

\subsection{Estimation of Distance Covered}
\label{sec:dist_est}
In \cref{tab:dataset_stats} we provide estimated total distance covered and distance covered with ground-truth pose. We calculate posed distance by summing up the length of the ground-truth trajectories. To estimate the total distance covered, we need an estimation of the distance covered for frames that do not have a pose. Since the IMU uncertainty compounds quickly it is an unreliable estimate. Thus, we assume an average walking speed of 1.4 m/s and multiply that with the total duration of frames without pose. This assumption roughly holds in practice since for almost all sequence portions without pose, we travel with the camera at walking speed. 

\section{Dynamic Objects}
Examples of dynamic objects in our sequences include people walking and cycling, cars passing by, snow, water fountains, rotating chairs, deformable objects, doors opening, etc. We blur the faces of people as well as the license plates of cars.

\newpage

